\documentclass{style/naturep}

\usepackage{amssymb}
\usepackage{amsmath}
\usepackage{graphicx}
\usepackage{algorithmicx}
\usepackage{algorithm}
\usepackage{adjustbox}
\usepackage{graphicx}

\usepackage[noend]{algpseudocode}
\usepackage{bbm}
\usepackage{lineno}
\usepackage[margin=0.6in]{geometry}
\usepackage{ragged2e}
\usepackage{setspace}
\usepackage{longtable}
\usepackage{hyperref}
\usepackage{anyfontsize}
\usepackage{setspace}
\usepackage{float}
\usepackage{booktabs}
\usepackage{multirow}
\usepackage{blindtext}
\usepackage{tabularx}
\usepackage{caption}
\usepackage{subcaption}
\usepackage{enumitem, kantlipsum}
\usepackage{xspace}
\usepackage{pifont}
\newcommand{\xmark}{\ding{55}}%

\usepackage{graphicx}
\usepackage{amsmath}
\usepackage{amssymb}
\usepackage{booktabs}
\usepackage{float}
\usepackage{placeins}
\usepackage[table,x11names]{xcolor}
\definecolor{maroon}{cmyk}{0,0.87,0.68,0.32}
\definecolor{gray}{rgb}{0.3,0.3,0.3}

\newcommand{\ours}{\textsc{UNI}\xspace}

\newcommand\Heading[1]{
  \noindent\textbf{\Large{#1}}
}

\newcommand\heading[1]{
  \noindent\textbf{\large{#1}}
}

\newcommand\hheading[1]{
  \noindent\textbf{#1}
}

\title{\begin{flushleft}{\begin{spacing}{1}
   A General-Purpose Self-Supervised Model for Computational Pathology
\end{spacing}}\end{flushleft}}

\makeatletter
\let\saved@includegraphics\includegraphics
\AtBeginDocument{\let\includegraphics\saved@includegraphics}

\makeatother

\begin{document}

\maketitle
\vspace{-20mm}
\begin{spacing}{1.4}
\noindent Richard J. Chen$^{1,2,3,4,5\boldsymbol{\ddag}}$, Tong Ding$^{1\boldsymbol{\ddag}}$, Ming Y. Lu$^{1,2,3,4,6\boldsymbol{\ddag}}$, Drew F. K. Williamson$^{1,2,3\boldsymbol{\ddag}}$, Guillaume Jaume$^{1,2,3,4}$, Bowen Chen$^{1,2}$, Andrew Zhang$^{1,2,3,4,7}$, Daniel Shao$^{1,2,3,4,7}$,  Andrew H. Song$^{1,2,3,4}$, Muhammad Shaban$^{1,2,3,4}$, Mane Williams$^{1,2,3,4,5}$, Anurag Vaidya$^{1,2,3,4,7}$, Sharifa Sahai$^{1,2,3,4,9}$, Lukas Oldenburg$^{1}$, Luca L. Weishaupt$^{1,2,3,4,7}$, Judy J. Wang$^{1}$, Walt Williams$^{1,8}$, Long Phi Le$^{2,7}$, Georg Gerber$^{1}$, Faisal Mahmood$^{*1,2,3,4,10}$
\end{spacing}

\vspace{-7mm}
\begin{spacing}{1.4}
\begin{affiliations}
 \item Department of Pathology, Brigham and Women's Hospital, Harvard Medical School, Boston, MA
 \item Department of Pathology, Massachusetts General Hospital, Harvard Medical School, Boston, MA
 \item Cancer Program, Broad Institute of Harvard and MIT, Cambridge, MA 
 \item Cancer Data Science Program, Dana-Farber Cancer Institute, Boston, MA
 \item Department of Biomedical Informatics, Harvard Medical School, Boston, MA
 \item Electrical Engineering and Computer Science, Massachusetts Institute of Technology (MIT), Cambridge, MA
 \item Health Sciences and Technology, Harvard-MIT, Cambridge, MA
 \item Harvard John A. Paulson School of Engineering And Applied Sciences, Harvard University, Cambridge, MA
 \item Department of Systems Biology, Harvard University, Cambridge, MA
 \item Harvard Data Science Initiative, Harvard University, Cambridge, MA
 \\$\boldsymbol{\ddag}$ Contributed Equally
 \\\textbf{*Corresponding author}: Faisal Mahmood (faisalmahmood@bwh.harvard.edu)
\end{affiliations}
\end{spacing}

\vspace{-3mm}
\begin{spacing}{1.2}
\noindent \textbf{Tissue phenotyping is a fundamental computational pathology (CPath) task in learning objective characterizations of histopathologic biomarkers in anatomic pathology. However, whole-slide imaging (WSI) poses a complex computer vision problem in which the large-scale image resolutions of WSIs and the enormous diversity of morphological phenotypes preclude large-scale data annotation. Current efforts have proposed using pretrained image encoders with either transfer learning from natural image datasets or self-supervised pretraining on publicly-available histopathology datasets, but have not been extensively developed and evaluated across diverse tissue types at scale. We introduce \ours, a general-purpose self-supervised model for pathology, pretrained using over 100 million tissue patches from over 100,000 diagnostic haematoxylin and eosin-stained WSIs across 20 major tissue types, and evaluated on 33 representative CPath clinical tasks in CPath of varying diagnostic difficulties. In addition to outperforming previous state-of-the-art models, we demonstrate new modeling capabilities in CPath such as resolution-agnostic tissue classification, slide classification using few-shot class prototypes, and disease subtyping generalization in classifying up to 108 cancer types in the OncoTree code classification system. UNI advances unsupervised representation learning at scale in CPath in terms of both pretraining data and downstream evaluation, enabling data-efficient AI models that can generalize and transfer to a gamut of diagnostically-challenging tasks and clinical workflows in anatomic pathology.
}
\end{spacing}

\clearpage


\begin{spacing}{1.35}
\Heading{Introduction}

\noindent The clinical practice of pathology involves performing a large range of tasks: from tumor detection and subtyping to grading and staging, and with thousands of possible diagnoses, a pathologist must be adept at solving an incredibly diverse group of problems, often simultaneously. Contemporary computational pathology (CPath) has expanded this array even further by enabling `omics' predictions\cite{coudray2018classification,he2020integrating,kather2019deep}, direct prognostication\cite{skrede2020deep,wulczyn2021npj,mobadersany2018predicting,courtiol2019deep}, and therapeutic response prediction\cite{vanguri2022multimodal} from microscopic images, among other applications\cite{bejnordi2017diagnostic, bandi2018detection,tolkach2020high,bulten2020automated,bera2019artificial,echle2021deep,hahn2021expanded,cooper2023machine,graham2023screening}. With a vast array of tasks and the fact that many tasks in pathology are difficult to acquire data for due to the rarity of the underlying diseases or the need for expensive manual annotations by pathologists, training a single deep learning model from scratch for every possible task is impractical. These factors have led to the broad reliance on transfer learning techniques in CPath, which have proven effective in tasks such as metastasis detection\cite{campanella2019clinical}, mutation prediction\cite{kather2020pan, fu2020pan}, prostate cancer grading\cite{bulten2022artificial}, and outcome prediction\cite{chen2022pan, foersch2023multistain}.

The transfer learning, generalization and scaling capabilities of self-supervised (or pretrained) models are intrinsically tied to the size and diversity of the training data\cite{he2022masked,oquab2023dinov2,huang2023leveraging,lu2023towards,balestriero2023cookbook}. In general computer vision, the development and evaluation of many fundamental self-supervised models\cite{chen2021mocov3,caron2021emerging,chen2020simple,grill2020bootstrap,oord2018representation,donahue2019large} are based on the ImageNet Large Scale Visual Recognition Challenge\cite{deng2009imagenet,russakovsky2015imagenet}, starting with ImageNet-1K (IN-1K) encompassing 1.2 million images from 1,000 classes, followed by ImageNet-22K (IN-22K, 14.2 million images, 21,841 classes), and then even larger datasets such as LVD-142M\cite{oquab2023dinov2}, JFT-300M\cite{sun2017revisiting}, and beyond\cite{zhai2022scaling,goyal2019scaling,openai2023gpt4}. Such models have also been described as ``foundation models" due to their ability to adapt to a wide range of downstream tasks when pretrained on massive amounts of data at scale\cite{bommasani2021opportunities,yuan2021florence}. In CPath, The Cancer Genome Atlas (TCGA, $\sim$29,000 FFPE and Frozen H\&E WSIs, 32 cancer types)\cite{weinstein2013cancer} similarly serves as the basis for most self-supervised models\cite{wang2021transpath,chen2022self,wang2022transformer,azizi2023robust,kang2023benchmarking,li2021dual,lazard2023giga,yu2023slpd,schirris2022deepsmile,vu2023handcrafted,quiros2022self,claudio2021adversarial,zhao2020predicting,jiang2023masked,wang2023retccl,filiot2023scaling,wang2023retccl} along with other histology datasets\cite{srinidhi2022self,srinidhi2021improving,saillard2021self,dehaene2020self,koohbanani2021self, boyd2021self,ciga2021self,li2021sslp,yang2021self,lin2023sgcl,dipalma2023histoperm,tellez2019neural, jiang2023hierarchical}, with a number of prior works demonstrating great progress in learning meaningful representations of histology tissue for clinical pathology tasks\cite{wang2022transformer,azizi2023robust,saldanha2023self,niehues2023generalizable,seraphin2023prediction,lu2023visual,mokhtari2023interpretable,jaume2023modeling,horst2023histology,wagner2023fully,horst2023cellvit,zhang2023prompt,kaczmarzyk2023champkit,zhang2022gigapixel,nasrallah2023machine,li2023task}. However, current pretrained models for CPath remain constrained by: 1) limited size and diversity of pretraining data, as the TCGA is comprised of mostly primary cancer histology slides, and 2) limited evaluation of generalization performance across diverse tissue types, as many pan-cancer analyses and popular clinical tasks in CPath are also based on annotated histology region-of-interests (ROIs) and slides from TCGA\cite{fu2020pan,chen2022pan,kather2020pan,saltz2018spatial,komura2022universal,kather2019deep,kaczmarzyk2023champkit,kalra2020yottixel,schmauch2020deep,graham2023one,diao2021human,wulczyn2020deep,riasatian2021fine}. Addressing these two limitations is critical for the broader development of foundation models in CPath that can generalize and transfer to real-world clinical settings with widespread applications.

\begin{figure*}
\centering
\includegraphics[width=1.00\textwidth]{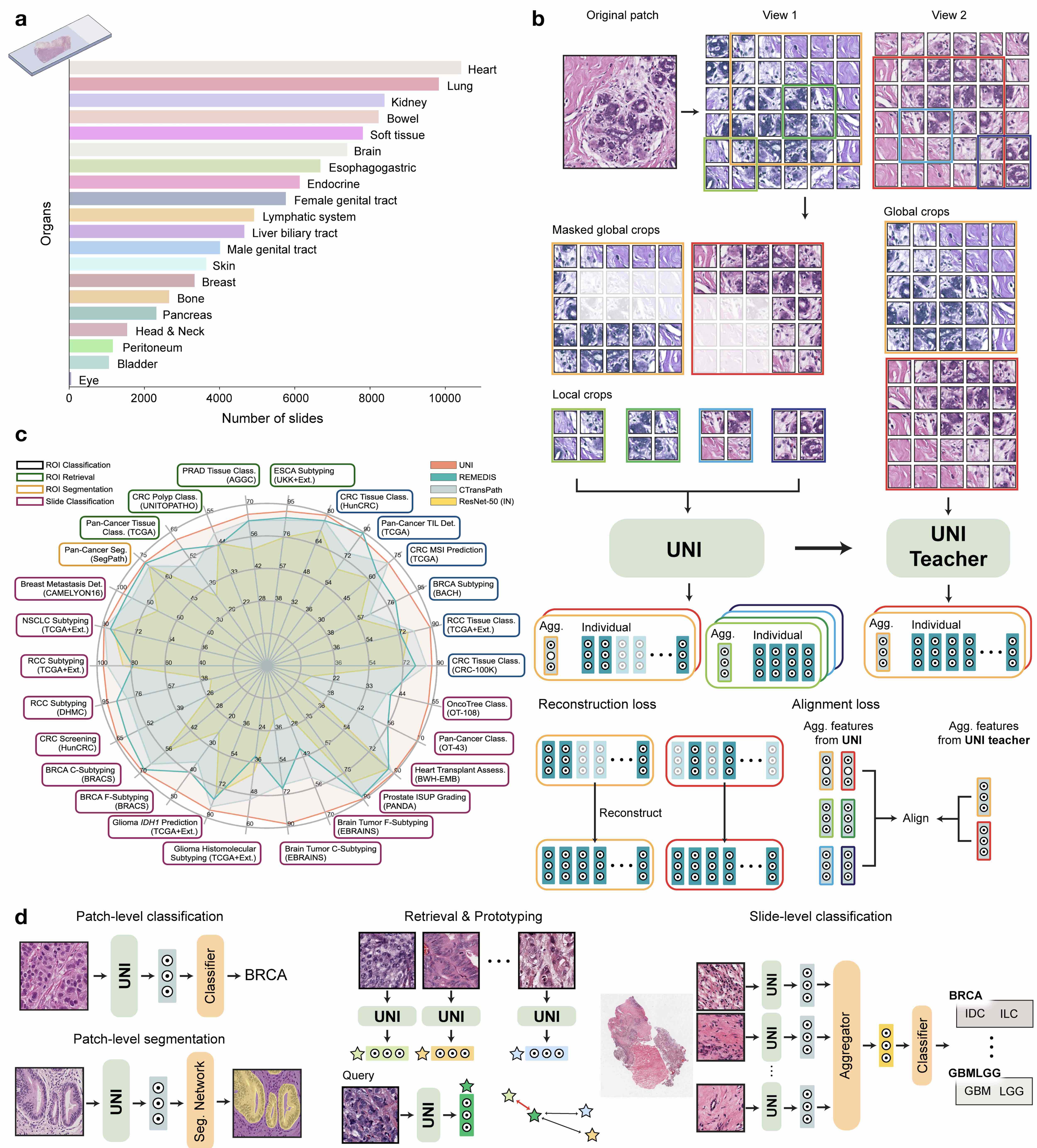}
\vspace{-5mm}
\caption{\textbf{Overview of \ours.} \ours is a general-purpose, self-supervised vision encoder for anatomic pathology based on the Vision Transformer architecture, achieving state-of-the-art performance across 33 clinical tasks in anatomic pathology. \textbf{a.} Slide distribution of Mass-100K, a large-scale and diverse pretraining dataset of 100 million tissue patches sampled from over 100,000 diagnostic whole-slide images across 20 major organ types. \textbf{b.} \ours is pretrained on Mass-100K using the DINOv2 self-supervised training algorithm\cite{oquab2023dinov2}, which consists of: 1) a mask image modeling objective\cite{zhou2021ibot} and a self-distillation objective\cite{caron2021emerging}. \textbf{c.} \ours outperforms other pretrained encoders on 33 clinical tasks in anatomical pathology (average performance of the 8 SegPath tasks reported). \textbf{d.} The evaluation tasks are comprised of ROI-level classification, segmentation, retrieval \& prototyping, and slide-level classification tasks. Further details are described in the \textbf{Online Methods}.}
\label{fig:overall}
\end{figure*}

\begin{figure*}
\centering
\includegraphics[width=\textwidth]{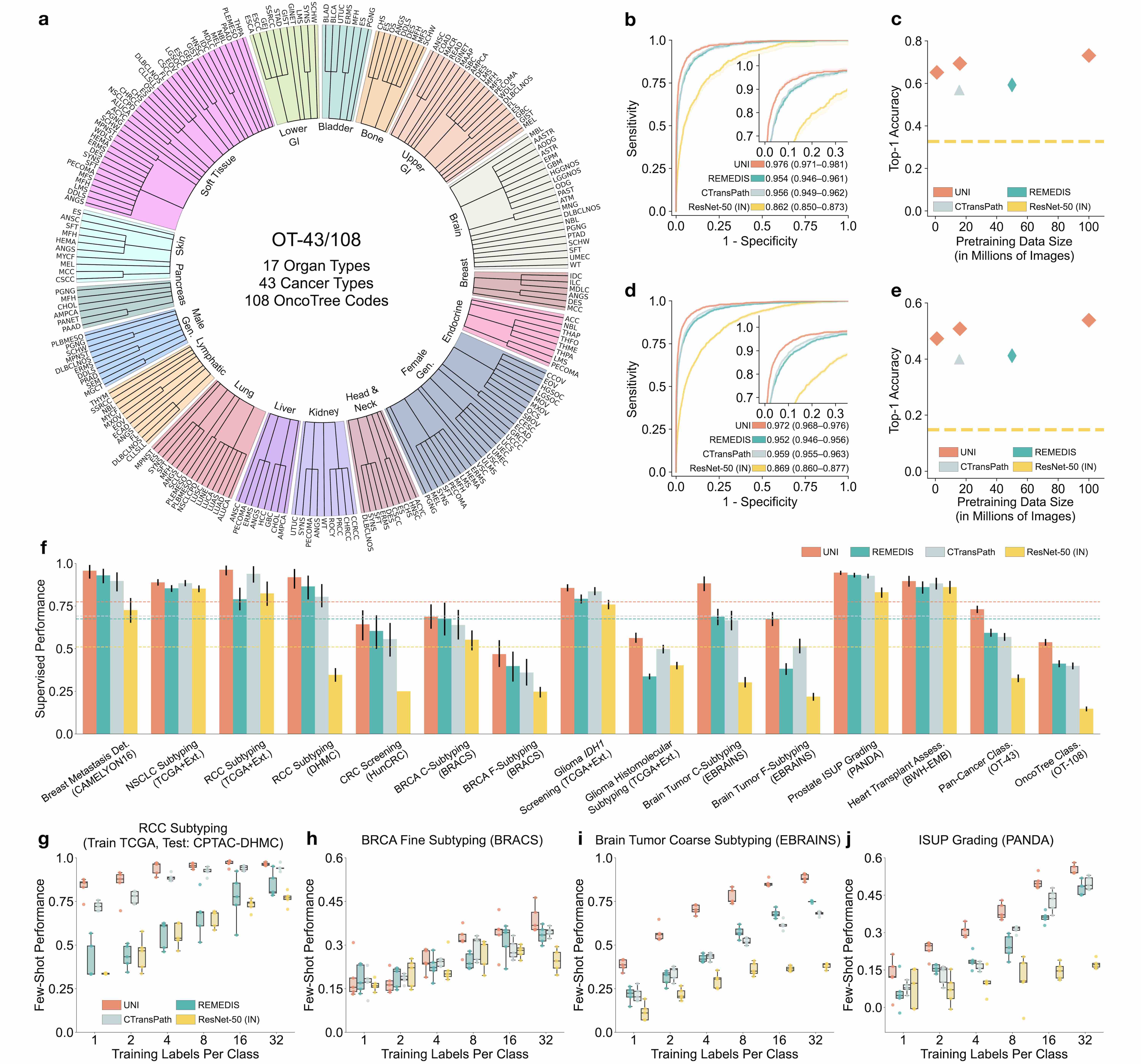}
\caption{\textbf{Slide-level tasks for OncoTree-43, OncoTree-108, and other slide-level tasks.} \textbf{a.} Organ and OncoTree code distribution for the slide-level OncoTree-43 (OT-43) and OncoTree-108 (OT-108) classification tasks. All comparisons with \ours are evaluated on 43-way cancer type classification and 108-way oncotree code classification tasks with OT-43 and OT-108, respectively. Further details regarding data distribution are provided in \textbf{Extended Data Table \ref{tab:op43/108}}. \textbf{b.} AUROC comparisons of \ours with other pretrained encoders on OT-43. \textbf{c.} Top-1 accuracy comparisons of \ours across different pretraining data scales (Mass-1K, Mass-22K, Mass-100K) on OT-43. \textbf{d.} AUROC comparisons of \ours with other pretrained encoders on on OT-108. \textbf{e.} Top-1 accuracy comparisons of \ours across different pretraining data scales (Mass-1K, Mass-22K, Mass-100K) on OT-108. \textbf{f.} Supervised performance of \ours and its comparisons across 15 weakly-supervised slide-level classification tasks. Detailed performance metrics for all slide classification tasks are further provided in \textbf{Extended Data Table \ref{tab:ot-43-compare}-\ref{tab:slide-emb}}. \textbf{g-j.} The few-shot slide-level performance with $K\in\{1,2,4,8,16,32\}$ slides per class reported for four tasks. Boxes indicate quartile values of model performance ($n=5$ runs) and whiskers extend to data points within 1.5$\times$ the interquartile range, with comparisons on all tasks visualized in \textbf{Extended Data Figure \ref{fig:slide-level-fs}}.}
\label{fig:slide-level}
\end{figure*}

In this work, we build upon these prior efforts by introducing a new general-purpose, self-supervised vision encoder for pathology, \textbf{\ours}, a Vision Transformer (ViT-Large)\cite{dosovitskiy2021image} pretrained on the largest histology slide collection used in self-supervised learning to date, termed \textbf{Mass-100K}. Mass-100K is a pretraining dataset that consists of over 100 million tissue patches from 100,426 diagnostic H\&E whole-slide images (WSIs) across 20 major tissue types collected from from Massachusetts General Hospital (MGH) and Brigham \& Women's Hospital (BWH), as well as Genotype-Tissue Expression (GTEx) consortium\cite{gtex2015genotype}, providing a rich source of information for learning objective characterizations of histopathologic biomarkers (\textbf{Figure~\ref{fig:overall}a}, \textbf{Extended Data Table~\ref{tab:op100k}}).
In the pretraining stage, we directly employ a self-supervised learning approach called DINOv2\cite{oquab2023dinov2}, which has been shown to yield strong, off-the-shelf representations for downstream tasks without the need for further finetuning with labeled data (\textbf{Figure~\ref{fig:overall}b}); more details regarding model design and training are available in the \textbf{Online Methods}. We demonstrate the versatility of \ours on diverse machine learning settings within CPath, including ROI-level classification, segmentation and image retrieval, and slide-level weakly-supervised and semi-supervised learning (\textbf{Figure \ref{fig:overall}c}). In total, we assess \ours on 33 clinical tasks across anatomic pathology that range in diagnostic difficulty, such as nuclear segmentation, primary and metastatic cancer detection, cancer grading and subtyping, gene mutation prediction and molecular subtyping, organ transplant assessment, and several pan-cancer classification tasks which includes subtyping to 108 cancer types in the OncoTree cancer classification system\cite{kundra2021oncotree} (\textbf{Figure \ref{fig:overall}d}, \textbf{Figure \ref{fig:slide-level}a}). In addition to outperforming previous state-of-the-art models such as CTransPath\cite{wang2022transformer} and REMEDIS\cite{azizi2023robust}, we also demonstrate new capabilities such as resolution-agnostic tissue classification and few-shot class prototypes for prompt-based slide classification, highlighting the potential of \ours as a foundation model for further developing AI models in anatomic pathology.

\Heading{Results}

\heading{Pretraining scaling laws in CPATH.}

\noindent
A pivotal characteristic of foundation models lies in their capability to deliver improved downstream performance on a variety of tasks when trained on larger datasets. Though datasets such as CAMELYON16\cite{bejnordi2017diagnostic} and TCGA-NSCLC\cite{luadlusc_campbell2016distinct} are commonly used to benchmark pretrained encoders using weakly-supervised multiple instance learning (MIL) algorithms\cite{li2021dual,lu2021data,shao2021transmil,wang2022transformer}, they only source tissue slides from a single organ and are mainly used for predicting binary disease states, which is not reflective of the broader array of disease entities seen in real-world anatomic pathology practice. Instead, we assess the generalization capabilities of \ours across diverse tissue types and disease categories by constructing a large-scale hierarchical classification task for CPath that follows the OncoTree (OT) cancer classification system\cite{kundra2021oncotree}. Using in-house BWH slides, we defined a dataset that comprises 5,564 WSIs from 43 cancer types further subdivided into 108 OncoTree codes, with at least 20 WSIs per OncoTree code. The dataset forms the basis of two tasks that vary in diagnostic difficulty: 1) 43-class cancer type classification (OT-43) and 2) 108-class OncoTree code classification (OT-108), which to our knowledge is the most label-complex CPath task proposed to date (\textbf{Figure \ref{fig:slide-level}a}). To assess scaling trends, we also pretrain UNI across varying data scales, with Mass-100K is subsetted to create: 1) Mass-22K (16 million histology image patches, 21,444 WSIs) and 2) Mass-1K (1 million images, 1,404 WSIs) (\textbf{Extended Data Table \ref{tab:op22k},\ref{tab:op1k}}). We assess \ours on Mass-1K/-22K/-100K on both OT-43 and OT108. For weakly-supervised slide classification, we follow the conventional paradigm of first pre-extracting patch-level features from tissue-containing patches in the WSI using a pretrained encoder, followed by training an Attention-Based MIL (ABMIL) algorithm\cite{ilse2018attention} that localizes and aggregates patch-level features with high diagnostic relevance for predicting the slide-level label. To reflect the label complexity challenges of these tasks, we report top-k accuracy ($k=1,3,5$) as well as weighted F1 and AUROC performance, with top-1 and top-5 accuracy as the primary evaluation metrics. Additional details regarding the OncoTree classification tasks are described in the \textbf{Online Methods}, distribution of cancer types and OncoTree codes provided in \textbf{Extended Data Table \ref{tab:op43/108}}, with detailed reporting of all model performances provided in \textbf{Extended Data Table \ref{tab:ot-43-compare}-\ref{tab:ot-108-length}}.

Overall, we demonstrate data scaling capabilities of self-supervised models in \ours, with the scaling trend for \ours on OT-43 and OT-108 visualized in \textbf{Figure \ref{fig:slide-level}c} and \textbf{Figure \ref{fig:slide-level}e}. On OT-43, we observe a +$4.2\%$ and +$3.6\%$ performance increase (both $p < 0.001$, two-sided paired permutation test) in top-1 and top-5 accuracy when scaling \ours from Mass-1K to Mass-22K, and a +$3.7\%$ and +$0.7\%$ performance increase ($p < 0.001$) when scaling from Mass-22K to Mass-100K. Similarly, on OT-108, we observe a +$6.5\%$ and +$8.6\%$ performance increase ($p < 0.001$) when comparing \ours pretrained on Mass-1K versus Mass-100K. \textbf{Extended Data Table \ref{tab:ot-43-length}} and \textbf{Extended Data Table \ref{tab:ot-108-length}} illustrate the effect of the number of images seen by \ours when evaluating intermediate checkpoints on OT-43 and OT-108 respectively, demonstrating gains with longer pretraining in addition to increasing dataset size. After 50,000 training iterations (115 million samples seen by the network during pretraining), \ours pretrained on Mass-22K outperforms \ours on Mass-100K on OT-43, with comparable performance on OT-108. When comparing models with 50,000 iterations (115 million samples) versus 125,000 iterations (384 million samples), performance on OT-43 and OT-108 monotonically increases for \ours pretrained on Mass-100K, whereas \ours performance pretrained on Mass-22K is less stable and decreases compared to earlier iterations. These scaling trends align with findings observed in many ViT models applied to natural images\cite{dosovitskiy2021image,zhai2022scaling,he2022masked}, in which the performance of larger ViT variants improves as the pretraining dataset grows.

We compare \ours pretrained on Mass-100K to publicly-available pretrained encoders used in CPath on OT-43 and OT-108 tasks: 1) ResNet-50\cite{he2016deep} pretrained on IN-1K, 2) CTransPath\cite{wang2022transformer} pretrained on TCGA and PAIP\cite{kim2019paip}, and 3) and REMEDIS\cite{azizi2023robust} pretrained on TCGA. We observe that \ours outperforms all baselines by a wide margin. On OT-43, \ours achieves a top-5 accuracy of $93.8\%$ and AUROC of $0.976$, outperforming the next best-performing model (REMEDIS) by +$6.3\%$ and +$0.022$ on these respective metrics (both $p < 0.001$) (\textbf{Figure \ref{fig:slide-level}b}, \textbf{Extended Data Table \ref{tab:ot-43-compare}}). On OT-108, we observe a similar margins of performance increase with +$10.8\%$ and +$0.020$ ($p < 0.001$) over REMEDIS (\textbf{Figure \ref{fig:slide-level}c}, \textbf{Extended Data Table \ref{tab:ot-108-compare}}). On more challenging performance metrics such as top-1 accuracy, \ours outperforms the next best-performing models by +$13.8\%$ and +$12.6\%$ on OT-43 and OT-108, respectively. We observe similar performance gains with other \ours configurations over these models. We also emphasize the superior pretraining efficiency of \ours over others. Though \ours is pretrained on a larger histology dataset (number of total patches and WSIs included), CTransPath and REMEDIS are respectively trained $4\times$ and $13\times$ longer than UNI (total number of images seen).

\heading{Weakly-supervised slide classification.} 

\noindent We investigate \ours's capabilities further across a diverse range of 15 slide-level classification tasks, which include: breast cancer metastasis detection (CAMELYON16)\cite{bejnordi2017diagnostic}, ISUP grading in prostate cancer (PANDA)\cite{bulten2022artificial}, and cardiac transplant assessment (in-house BWH slides)\cite{lipkova2022deep} among others. Similar to the OT-43 and OT-108 evaluations, we compare the pre-extracted features from \ours with that of other pretrained encoders in training ABMIL for weakly-supervised slide classification. As CTransPath and REMEDIS were trained using almost all TCGA slides, the reported performance of these models on TCGA tasks may be contaminated with data leakage and thus unfairly inflated. As a result, we exclude TCGA tasks that lack external test cohorts from our comparisons. A detailed description of tasks and evaluations are provided in the \textbf{Online Methods}, with detailed reporting of all slide classification performance presented in \textbf{Extended Data Table \ref{tab:ot-43-compare}-\ref{tab:slide-emb}}.

Across all 15 slide-level classification tasks, \ours consistently outperforms all baselines (ResNet-50, CTransPath, REMEDIS), with more substantial improvements observed on tasks characterized by higher diagnostic complexity (\textbf{Figure \ref{fig:slide-level}f}). On conventional cancer detection and subtyping and detection benchmark tasks such as NSCLC subtyping (trained on TCGA\cite{luadlusc_campbell2016distinct} and tested on CPTAC\cite{edwards2015cptac,luad_cptac_gillette2020proteogenomic,lusc_cptac_satpathy2021proteogenomic}) and RCC subtyping (trained on TCGA\cite{ccrcc_cancer2013comprehensive,chrcc_davis2014somatic,prcc_cancer2016comprehensive} and tested on CPTAC\cite{ccrcc_cptac_li2023histopathologic} and DHMC\cite{zhu2021development}) and breast metastasis detection (official folds in CAMELYON16), \ours achieves balanced accuracy scores of $88.9\%$, $96.3\%$ and $95.7\%$ respectively, outperforming the conventional ResNet-50 baseline (by +$3.7\%$, $p < 0.001$; +$13.9\%$, $p < 0.001$; +$23.1\%$, $p < 0.001$) as well as CTransPath (by +$0.5\%$, $p=0.598$; +$2.4\%$, $p=0.408$; +$6.0\%$, $p=0.102$) and REMEDIS (by +$3.5\%$, $p < 0.001$; +$17.3\%$, $p=0.001$; +$2.7\%$, $p=0.247$) (\textbf{Extended Data Table \ref{tab:slide-c16}-\ref{tab:slide-rcc}}). On more challenging tasks such as the ISUP grading task in PANDA, \ours achieves a balanced accuracy of $75.7\%$ and a quadratic weighted Cohen’s $\kappa$ of $0.946$, outperforming the next best-performing model (REMEDIS) by +$4.6\%$ ($p < 0.001$) and +$0.014$ ($p < 0.05$) on these respective metrics (\textbf{Extended Data Table \ref{tab:slide-panda}}). On hierarchical classification tasks such as BRCA subtyping in BRACS\cite{brancati2021bracs} (\textbf{Extended Data Table \ref{tab:brca-label}}), we observe increases in balanced accuracy and AUROC over the next best-performing model (REMEDIS) when scaling from the coarse-grained subtyping task (+$1.1\%$, $p=0.865$; +$0.023$, $p=0.442$) to the fine-grained subtyping task (+$7.0\%$, $p = 0.285$; +$0.088$, $p < 0.01$) (\textbf{Extended Data Table \ref{tab:slide-bracs-c} and \ref{tab:slide-bracs-f}}). For further assessment of \ours's performance on hierarchical classification tasks, we also developed several challenging slide-level tasks of varying label complexity using the EBRAINS Digital Tumor Atlas\cite{roetzer2022digital} and the TCGA-GBMLGG cohort\cite{gbm_brennan2013somatic,lgg_cancer2015comprehensive}, which include: a 2-class glioma \textit{IDH1} mutation prediction task, a 5-class glioma histomolecular subtyping task (predicting both \textit{IDH1}/1p19q status and histologic subtype), a 12-class brain tumor subtyping task (predicting main cancer type), and a 30-class brain tumor subtyping task (predicting diagnosis) (\textbf{Extended Data Table \ref{tab:ebrains-mut-label},\ref{tab:ebrains-diagnosis-label}}). On these 4 respective tasks, \ours achieves balanced accuracy scores of $85.6\%$, $56.2\%$, $88.3\%$ and $67.5\%$, outperforming the next best-performing model (either CTransPath or REMEDIS), by +$2.0\%$ ($p=0.076$), +$6.4\%$ ($p=0.001$), +$19.6\%$ ($p<0.001$), and +$16.1\%$ ($p<0.001$) respectively (\textbf{Extended Data Table \ref{tab:slide-idh}-\ref{tab:slide-ebrains-f}}). Overall, \ours achieves the highest average supervised performance across all tasks $77.4\%$, with average performance increases of +$26.4\%$, +$8.3\%$, and +$10.0\%$ respective to ResNet-50 ($51.0\%$), CTransPath ($69.1\%$) and REMEDIS ($67.4\%$).

Data contamination is a growing concern in foundation models trained on large collections of public datasets\cite{jacovi2023stop,magar2022data,brown2020language,dodge2021documenting,kapoor2023leakage}. Though labels may not be explicitly leaked into the model during self-supervised training, models evaluated with transductive inference (\textit{e.g.} the test set is made available to the model in either unsupervised or supervised training) may exhibit optimistically-biased performance, which has been empirically observed in CPath\cite{xiang2022exploring}. Though comparisons of all pretrained encoders were evaluated on held-out testing data that were not seen during pretraining, we also assess and compare \ours against CTransPath and REMEDIS on TCGA-evaluated data in the NSCLC subtyping, RCC subtyping, glioma \textit{IDH1} mutation prediction and glioma histomolecular subtyping tasks. We note that though all ABMIL models in these tasks were developed on histology slides of their respective TCGA data sources, all results presented in \textbf{Figure~\ref{fig:slide-level}} are from evaluation on external data sources (CPTAC, DHMC, and EBRAINS). To study optimistic bias of self-supervised models with transductive inference, we additionally hold out an internal test fold in TCGA in our supervised evaluation of these tasks, a common practice in many CPath study designs that use TCGA\cite{lu2021data,li2021dual,shao2021transmil,zhao2020predicting}. We denote all results with transductive inference as ``gray" in \textbf{Extended Data Table \ref{tab:slide-nsclc}, \ref{tab:slide-rcc}, \ref{tab:slide-idh}, \ref{tab:slide-molsub}}.

In comparing \ours against REMEDIS and CTransPath on TCGA-evaluated data, we not only find \ours still outperforms these models on several tasks, but also observe substantial performance decreases when comparing the in-domain versus out-of-domain performance of these models. Foremost, \ours still outperforms the ResNet-50 baseline on these tasks, with an overall improvement of +$12.78\%$. On NSCLC and RCC subtyping, \ours interestingly outperforms CTransPath by a larger margin on the internal TCGA test sets than the external test cohorts (+$2.0\%$ versus +$0.5\%$ on NSCLC subtyping, +$5.5\%$ versus +$2.4\%$ on RCC subtyping). Compared to REMEDIS, though \ours reaches the best performance on NSCLC subtyping, REMEDIS reaches a best balanced accuracy of $97.3\%$ on the TCGA test set in RCC subtyping (compared to $94.7\%$ from \ours). However, we observe substantial performance decreases in REMEDIS ($97.3\%$ to $79.0\%$ from internal to external test set evaluation), whereas \ours maintains its performance (increasing from $94.7\%$ to $96.3\%$). We make a similar observation in glioma \textit{IDH1} mutation prediction, in which both CTransPath and REMEDIS outperform $\ours$ on the TCGA test set ($89.1\%$ and $81.9\%$ respectively compared to $80.8\%$), but then underperform on the EBRAINS test set (decrease to $83.6\%$ and $79.2\%$ in CTransPath and REMEDIS respectively, with increase to $85.6\%$ in \ours). On glioma histomolecular subtyping, \ours achieves the best performance on internal and external test evaluation. Overall, we find evidence of data contamination of self-supervised models in CPath when evaluated on the same data source used in pretraining. We emphasize that data contamination only exists in how the models are utilized, not in the models themselves which have been demonstrated to transfer well in on clinical settings independent of TCGA\cite{niehues2023generalizable,seraphin2023prediction,wagner2023fully,azizi2023robust}. However, as many CPath studies are reliant on TCGA, \ours is more generalizable and adaptable to clinical settings (both pre-clinical research and clinical translation) studying diverse cancer types.




\heading{Label efficiency of few-shot slide classification.} 

\noindent To study the label efficiency of \ours in the MIL paradigm, we additionally evaluate all slide-level tasks using few-shot learning. Few-shot learning is an evaluation scheme that studies the generalization capabilities of pretrained models on new tasks ($C$ classes) given a limited number of examples ($K$ training samples per class, also called supports or shots), and is posed as solving a ``$C$-way, $K$-shot" learning task. For all pretrained encoders, we trained an ABMIL model with $K \in \{ 1, 2, 4, 8, 16, 32 \}$ training examples per class, where $K$ is limited to 32 due to small support sizes in rare disease categories. As the performance can fluctuate depending on which $K$ examples are chosen for each class, we repeat experiments over five runs with $C \cdot K$ training examples randomly sampled each time. A detailed description of our few-shot MIL experimentation is provided in the \textbf{Online Methods}, with few-shot performance for all tasks summarized in \textbf{Extended Data Figure \ref{fig:slide-level-fs}}. 

\ours generally outperforms all other baselines when trained and evaluated with the same number of training examples per class within each task, consistently demonstrating superior label efficiency (\textbf{Figure \ref{fig:slide-level}g-j}, \textbf{Extended Data Figure \ref{fig:slide-level-fs}}). When comparing the 4-shot performance of \ours with that of other models (using the median performance), aside from the coarse-grained BRACS subtyping task, \ours outperforms all other baselines on all tasks, with the next best-performing model needing up of $8\times$ as many training examples per class to reach the same 4-shot performance of \ours. On challenging tasks such as fine-grained brain tumor subtyping in EBRAINS, the 4-shot performance of \ours outperforms other models by a large margin, only matched by the 32-shot performance of REMEDIS (\textbf{Figure \ref{fig:slide-level}i}). On ISUP grading in PANDA, \ours is consistently twice as label efficient across all few-shot settings (\textbf{Figure \ref{fig:slide-level}j}). For several tasks such as BRACS subtyping and glioma \textit{IDH1} mutation prediction, the 1-shot \ours performance is lower than that of other pretrained encoders, with performance gains not observed until 4-shot training and evaluation (\textbf{Figure \ref{fig:slide-level}g}, \textbf{Extended Data Figure \ref{fig:slide-level-fs}h}). Overall, our comprehensive evaluation of slide classification tasks demonstrates \ours's potential as the foundational model to be used routinely for histopathology research tasks, outperforming other baselines in performance and label efficiency.

\heading{Supervised ROI classification in linear classifiers.}

\noindent In addition to slide-level tasks, we also assess the capabilities of \ours on a diverse set of 10 ROI-level tasks of different tissue types, which include: colorectal tissue and polyp classification (CRC-100K-NONORM\cite{kather2019predicting}, HunCRC\cite{pataki2022huncrc}, UniToPatho\cite{barbano2021unitopatho}), CCRCC tissue classification (TCGA)\cite{brummer2022integrative}, CRC microsatellite instability (MSI) prediction (TCGA)\cite{kather2019deep}, PRAD tissue classification (AGGC)\cite{huo2022comprehensive}, BRCA subtyping (BACH)\cite{aresta2019bach}, ESCA subtyping (UKK)\cite{tolkach2023artificial}, and two pan-cancer tasks: tumor-immune lymphocyte (TIL) detection\cite{saltz2018spatial} and 32-class cancer tissue classification (TCGA)\cite{komura2022universal}. We note that 3 out of 10 tasks were trained and evaluated on TCGA, which may unfairly inflate the performance of CTransPath and REMEDIS in comparisons. To mitigate potential biases of site-specific staining variability in TCGA-derived tasks\cite{howard2021impact}, we stain normalize\cite{macenko2009method} all images in CRC MSI prediction, TIL detection, and 32-class pan-cancer tissue classification. For tasks that do not have official train-test folds, we also case- and site-stratify all samples when possible, such as using ROIs from the Helsinki Hospital (HEL) subset in the CCRCC tissue classification task and ROIs from the University Hospital Berlin—Charité (CHA) subset in the ESCA subtyping task as external cohorts respectively. For evaluation and comparisons, we perform logistic regression and K-nearest neighbors (KNN) on top of the pre-extracted features of each encoder, a common practice referred to as linear probing and KNN probing which measure discriminative performance and representation quality of pre-extracted features respectively\cite{balestriero2023cookbook}. We evaluate all tasks using balanced accuracy, with PRAD tissue classification additionally evaluated using weighted F1 score\cite{huo2022comprehensive}. A detailed description of ROI tasks and evaluation settings (linear and KNN probing) are provided in the \textbf{Online Methods}, with detailed reporting of all ROI classification performance presented in \textbf{Extended Data Table \ref{tab:patch-crc100k-lin}-\ref{tab:patch-tcga-tils-knn}}.

\begin{figure*}
\centering
\vspace{-5mm}
\includegraphics[width=\textwidth]{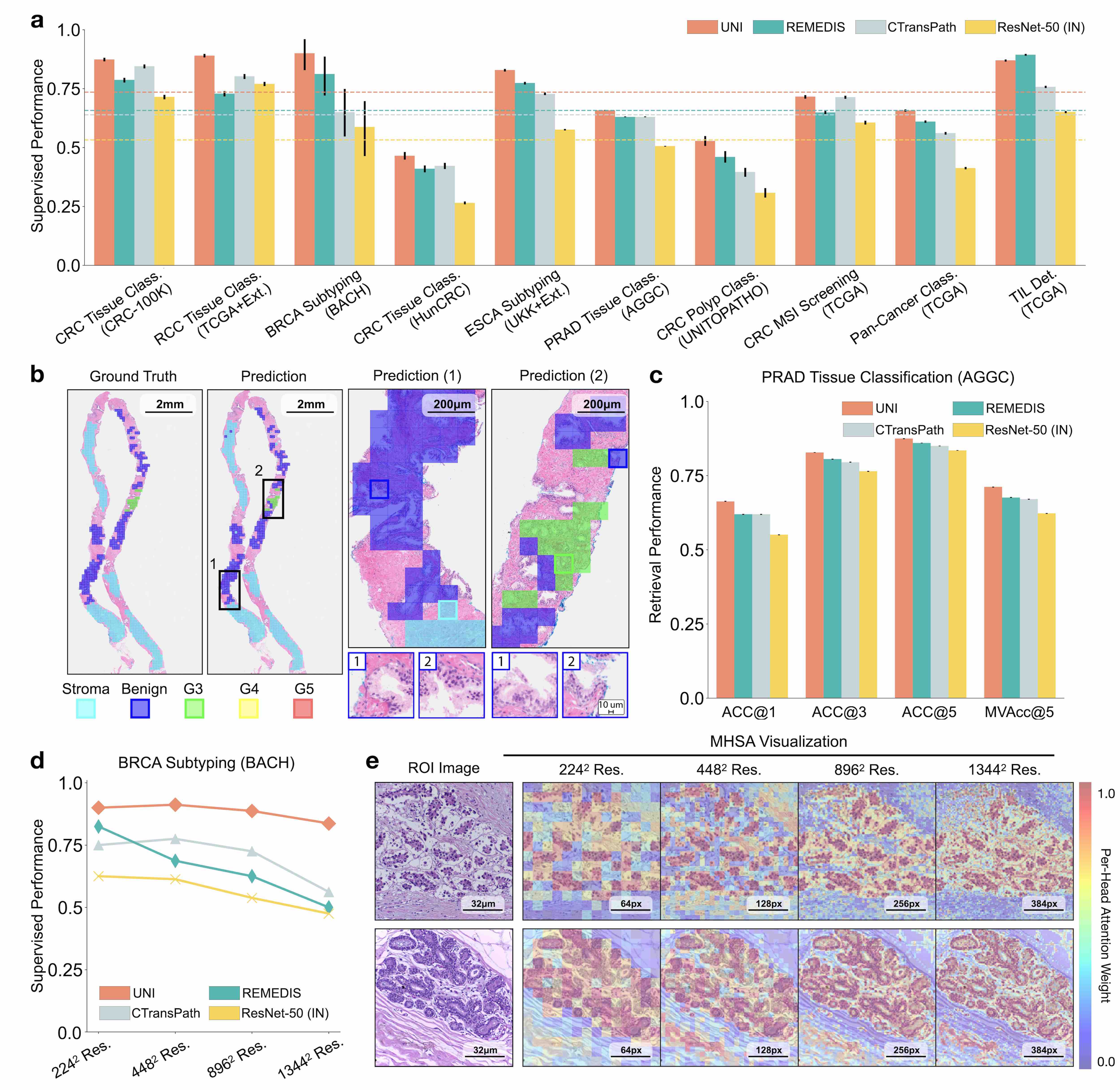}
\caption{
\textbf{ROI-level tasks.} \textbf{a.} Supervised linear probe performance of \ours across 10 ROI-level classification tasks. Dashed lines represent average performance of each model across all tasks, with error bars representing 95\% confidence intervals. \textbf{b.} Illustrative examples of $\ours$ on ROI classification for prostate adenocarcinoma (PRAD) tissue classification in AGGC. \textbf{Left} panel shows ground truth ROI-level labels overlayed on the WSI, and \textbf{right} panel shows predicted patch labels. Regions of interest are enlarged for better visualization, with further comparisons visualized in \textbf{Extended Data Figure \ref{fig:aggc-compare}}. \textbf{c.} ROI retrieval performance of \ours on PRAD tissue classification. We report Recall@$K$ for $K\in\{1,3,5\}$ and the mean recall, with error bars representing 95\% confidence intervals. All ROI retrieval comparisons are provided in \textbf{Extended Data Figure \ref{fig:patch-level-retrieval}}. \textbf{d.} Supervised KNN probe performance of \ours across various image resolutions in BRCA subtyping in BACH. \textbf{e.} Multi-head self-attention (MHSA) heatmap visualization of \ours across different image resolutions in BACH. Each colored square represents a $16 \times 16$ patch token encoded by \ours, with heatmap color corresponding to the attention weight of that patch token to the global $\textsc{[CLS]}$ token of the penultimate layer in \ours. \textbf{Top} and \textbf{bottom} respectively show visualizations for the invasive- and normal-labeled images, with further visualizations and interpretations provided in \textbf{Extended Data Figure \ref{fig:patch-level-res}-\ref{fig:crc2-res}}.}
\label{fig:patch-level-aggc}
\end{figure*}

Across all 10 ROI-level tasks, \ours outperforms nearly all baselines on all tasks with statistical significance, with overall improvements of +$19.9\%$, +$9.5\%$, +$7.7\%$ on linear probing for ResNet-50, CTransPath, and REMEDIS respectively \textbf{Figure \ref{fig:patch-level-aggc}a}. On conventional ROI benchmarks such as CRC tissue classification (CRC-100K) with linear probing, \ours outperforms ResNet-50 (+$15.9\%$, $p < 0.001$), CTransPath (+$2.9\%$, $p < 0.001$) and REMEDIS (+$8.7\%$, $p < 0.001$) by significant margins. Similar performance gains are also seen on more challenging tasks such as PRAD tissue classification (in weighted F1 score, +$0.131$, $p < 0.001$; +$0.020$, $p < 0.001$; +$0.027$, $p < 0.001$) and ESCA subtyping (+$25.3\%$, $p < 0.001$; +$10.1\%$, $p < 0.001$; +$5.5\%$, $p < 0.001$) for all three models respectively. \textbf{Figure \ref{fig:patch-level-aggc}b} visualizes \ours predictions on prostate cancer grading, in which a simple linear classifier trained with pre-extracted \ours features can achieve high agreement with pathologist annotations (\textbf{Extended Data Figure \ref{fig:aggc-compare}}). On CRC MSI prediction and TIL detection, CTransPath and REMEDIS achieve similar or better performance than \ours, which we note are 2 out of the 3 tasks that were pretrained and evaluated on TCGA data only. Interestingly, on a 32-class pan-cancer tissue classification task curated entirely from TCGA, \ours achieves the highest overall balanced accuracy and AUROC of $65.7\%$ and $0.975$, still outperforming the next best-performing model (REMEDIS) by +$4.7\%$ and +$0.017$ (both $p < 0.001$). In comparing performances with and without stain normalization, we observe a uniform decrease in performance across all pretrained encoders (\textbf{Extended Data Figure \ref{tab:patch-msi-lin}-\ref{tab:patch-tcga-tils-knn}}). On KNN probing, \ours similarly outperforms ResNet-50, CTransPath, and REMEDIS with an average of +$16.9\%$, +$10.1\%$, +$9.4\%$ performance increase across all tasks.

\heading{ROI retrieval.}

\noindent In addition to using the semantically-rich representations extracted from \ours for building task-specific classifiers in a supervised learning setting, representations can also be used to perform content-based image retrieval (CBIR). In CBIR, a query image is used to find similar images from a large database - \textit{e.g.} images sharing similar morphologies, diagnosis, or tissue site. Specifically, the query image is embedded into a low-dimensional feature representation using \ours, and then compared to other candidate images in the embedding space via a KNN look-up. For simplicity, we evaluate and compare the effectiveness of \ours with other baselines in retrieving histopathology image ROIs, and acknowledge that more sophisticated preprocessing, indexing, ranking, and filtering techniques may be used to boost performance, scalability and speed\cite{kalra2020yottixel, wang2023retccl, chen2022fast, li2023high} further. We evaluate histology image retrieval on 6 ROI-level tasks (tasks with at least 5 classes). In each task, we consider Acc@K for K $\in \{1,3,5\}$, which represent the standard top-K accuracy scores in retrieving images with the same class label as the query and MVAcc@5, which more strictly enforces that the majority vote of retrieved images must be the same class as the query for retrieval to be considered successful. For each task we use the same test set introduced in the supervised ROI-level evaluation section as queries and treat the supervised training set as the database of keys. Note that no supervised learning occurs in these experiments, and the labels are only used for evaluation. Additional details regarding ROI retrieval are provided in the \textbf{Online Methods}, with detailed reporting of all retrieval tasks provided in \textbf{Extended Data Table \ref{tab:aggc-retrieval}-\ref{tab:unitopatho-retrieval}}.

UNI outperforms other encoders on all tasks, demonstrating superior retrieval performance across diverse settings. On PRAD tissue classification (AGGC), UNI outperforms the next best-performing model (REMEDIS) by +$4.3\%$ and +$3.6\%$ on Acc@1 and MVAcc@5, respectively (both $p < 0.001$). A similar trend is noted for CRC tissue classification (HunCRC) (by +$2.8\%$ and +$2.5\%$ respectively compared to REMEDIS, both $p < 0.001$), ESCA subtyping (by +$3.8\%$ and +$3.0\%$ respectively compared to REMEDIS, both $p < 0.001$) and CRC polyp classification (UniToPatho) (by +$2.4\%$, $p = 0.037$ and +$3.3\%$, $p < 0.001$ respectively compared to CTransPath). On CRC tissue classification (CRC-100K), the gap between the top performing models is relatively small (by +$1.8\%$, $p < 0.001$ and +$1.0\%$, $p = 0.015$ respectively compared to REMEDIS), presumably because the different tissue types have very distinct morphology, as shown by the relatively high classification performance in linear probing. Lastly, on the more challenging 32-class pan-cancer tissue classification task, UNI outperforms the second-best performing model REMEDIS by a large margin of +$4.6\%$ for Acc@1 and +$4.2\%$ for MVAcc@5 (both $p < 0.001$). 

\heading{Representation quality of extracted features from high image resolutions.}


\noindent Though the evaluation of visual recognition models is mostly performed on resized $224 \times 224$ ($224^2$) images, we note that image resizing operations may change the image magnification and thus alter the interpretation of certain morphological features such as cellular atypia. To this end, we additionally study the representation quality of \ours's extracted features at varying resolutions in the BRCA subtyping (BACH) and CRC polyp classification (UniToPatho) tasks. Specifically, we resize and center-crop the original ROI images (BRCA subtyping - $2048\times 1536$ at 0.42 microns per pixel or mpp, \& CRC polyp classification - $1812\times 1812$ at 0.44 mpp) to $\{224^2, 448^2, 896^2, 1344^2\}$ and $\{224^2, 448^2, 896^2, 1792^2\}$ size, respectively, and perform linear and KNN probing. We note that all pretrained encoders use data augmentations that would change the image aspect ratio during self-supervised learning, with \ours additionally pretrained on high-resolution images following DINOv2. Additional details regarding multiple resolution evaluation are provided in the \textbf{Online Methods} and dataset descriptions for BRCA subtyping and CRC polyp classification, with detailed reporting of ROI classification performance for all resolutions reported in \textbf{Extended Data Figure \ref{fig:patch-level-res}, Extended Data Table \ref{tab:patch-bach-lin}, \ref{tab:patch-bach-knn}, \ref{tab:patch-unitopatho-lin}, \ref{tab:patch-unitopatho-knn}}. 

On both tasks, we observe \ours can adapt to high resolutions images, outperforming all comparisons across almost all resolutions and evaluation metrics. When evaluating on $224^2$-resized images (2.88 mpp) in BRCA subtyping, \ours outperforms the next best-performing model (CTransPath) by +$5.0\%$ and +$15.0\%$ on balanced accuracy in linear and KNN probing (\textbf{Figure \ref{fig:patch-level-aggc}d, Extended Data Figure \ref{fig:patch-level-res}a}). On $224^2$-resized images (3.60 mpp) in CRC polyp classification, \ours outperforms CTransPath by +$7.2\%$ on linear probing, with lower performance than CTransPath on KNN probing (\textbf{Extended Data Figure \ref{fig:patch-level-res}b}). When scaling the image resolutions used for evaluation, we observe wider margins of improvement as \ours outperforms CTransPath on $1344^2$-resized images (0.48 mpp) in BRCA subtyping (by +$25.0\%$ linear probe, $p < 0.001$; by +$27.5\%$ KNN probe, $p < 0.001$), and $1792^2$-resized images (0.45 mpp) in CRC polyp classification (by +$13.2\%$ linear probe, $p < 0.001$; by +$6.2\%$ KNN probe, $p < 0.001$). Additionally, we observe that \ours performance is robust across resolutions for both tasks, with the minimum and maximum performance gap ranges -$5.0\%$ (linear) and -$7.5\%$ (KNN) for BRCA subtyping and +$2.6\%$ (linear) and +$5.1\%$ (KNN) for CRC polyp classification. Such robustness is missing in other pretrained encoders, such as the -$22.5\%$ (linear) and -$21.3\%$ (KNN) performance gap for BRCA subtyping using CTransPath. In \textbf{Figure \ref{fig:aggc-compare}e} and \textbf{Extended Data Figure \ref{fig:bach-res}-\ref{fig:crc2-res}}, we can further visualize high-attention $16 \times 16$ patch tokens (represented as a colored square) that contribute to the extracted feature representation of \ours via interpretation of multi-head self-attention (MHSA) weights. In $224^2$-resized images from the BRCA subtyping task in BACH, In comparing MHSA heatmap visualizations across increasing image resolutions in BRCA subtyping, we observe that the high-attention $16 \times 16$ patch tokens in \ours become more fine-grained in localizing individual cells, with more specific delineation of tumor-stroma boundaries observed in $1344^2$-resized images. Interestingly, in $224^2$-resized images, we find that high-attention patch tokens are still able to localize invasive tumor cell nests and the cell-lining of ducts at a low resolution, which demonstrates \ours capabilities in encoding resolution-agnostic features. In contrast to CRC polyp classification, we observe that high-attention patch tokens in \ours for $224^2$-resized images have poor specificity in corresponding to any cell- or tissue-based morphological patterns in comparison with that of $1796^2$-resized images, which alludes to the important effect of image resizing on ROI datasets (\textbf{Extended Data Figure \ref{fig:crc2-res}}). Overall, these observations suggest that \ours can encode semantically-meaningful representations agnostic to most image resolutions, which is especially valuable in CPath tasks known to be optimal at different image magnifications. 


\heading{ROI cell type segmentation.}

\noindent We assess \ours on the largest, public ROI-level segmentation dataset, SegPath\cite{komura2023restaining}, a dataset for segmenting 8 major cell types in tumor tissue: epithelial cells, smooth muscle cells, red blood cells, endothelial cells, leukocytes, lymphocytes, plasma cells, and myeloid cells. Segmentation masks for each cell type were obtained via immunofluorescence and DAPI nuclear staining. All pretrained encoders are finetuned end-to-end using Mask2Former\cite{cheng2021mask2former}, a flexible framework commonly used for evaluating the off-the-shelf performance of pretrained encoders\cite{fang2023eva,fang2023eva2,oquab2023dinov2}. As the SegPath dataset divides the cell types into separate dense prediction tasks (8 tasks total), each model is individually finetuned per cell type, with the dice score used as the primary evaluation metric. As plain (non-hierarchical) ViT architectures lack vision-specific inductive biases for dense prediction tasks\cite{li2022exploring}, we additionally used the ViT-Adapter module alongside the Mask2Former head for finetuning \ours\cite{chen2022vitadapter,oquab2023dinov2}. Additional details regarding segmentation tasks are provided in the \textbf{Online Methods}, with detailed reporting of all segmentation tasks provided in \textbf{Extended Data Table \ref{tab:patch-level-seg}}.

Though hierarchical vision backbones such as Swin Transformers (CTransPath) and Convolutional Neural Networks (ResNet-50 and REMEDIS) have well-known advantages over plain backbones (ViT-Large in \ours) on dense prediction tasks, we observe \ours still outperforms all comparisons on a majority of cell types in SegPath. On individual segmentation tasks for the epithelial, smooth muscle, and red blood cell types, \ours achieves dice scores of $0.827$, $0.690$, and $0.803$, outperforming the next best-performing model (REMEDIS) by +$0.003$ ($p=0.164$), +$0.016$ ($p<0.001$), and +$0.008$ ($p=0.001$). On other segmentation tasks, \ours is the best-performing model on most cell types, with comparable results on lymphocyte ($0.651$ vs. $0.653$, $p=0.419$), and plasma cell segmentation ($0.737$ vs. $0.742$, $p=0.378$) against the best-performing model (REMEDIS). Across all 8 cell types in SegPath, \ours achieves the overall performance with an average dice score of $0.721$, outperforming ResNet-50 ($0.696$), CTransPath ($0.695$), and REMEDIS ($0.716$). Despite architectural constraints in using plain ViTs, we demonstrate that \ours is still competitive with state-of-the-art CNN and hierarchical vision models on cell segmentation.

\heading{Few-shot ROI classification with class prototypes.}

\begin{figure*}
\centering
\vspace{-3mm}
\includegraphics[width=0.9\textwidth]{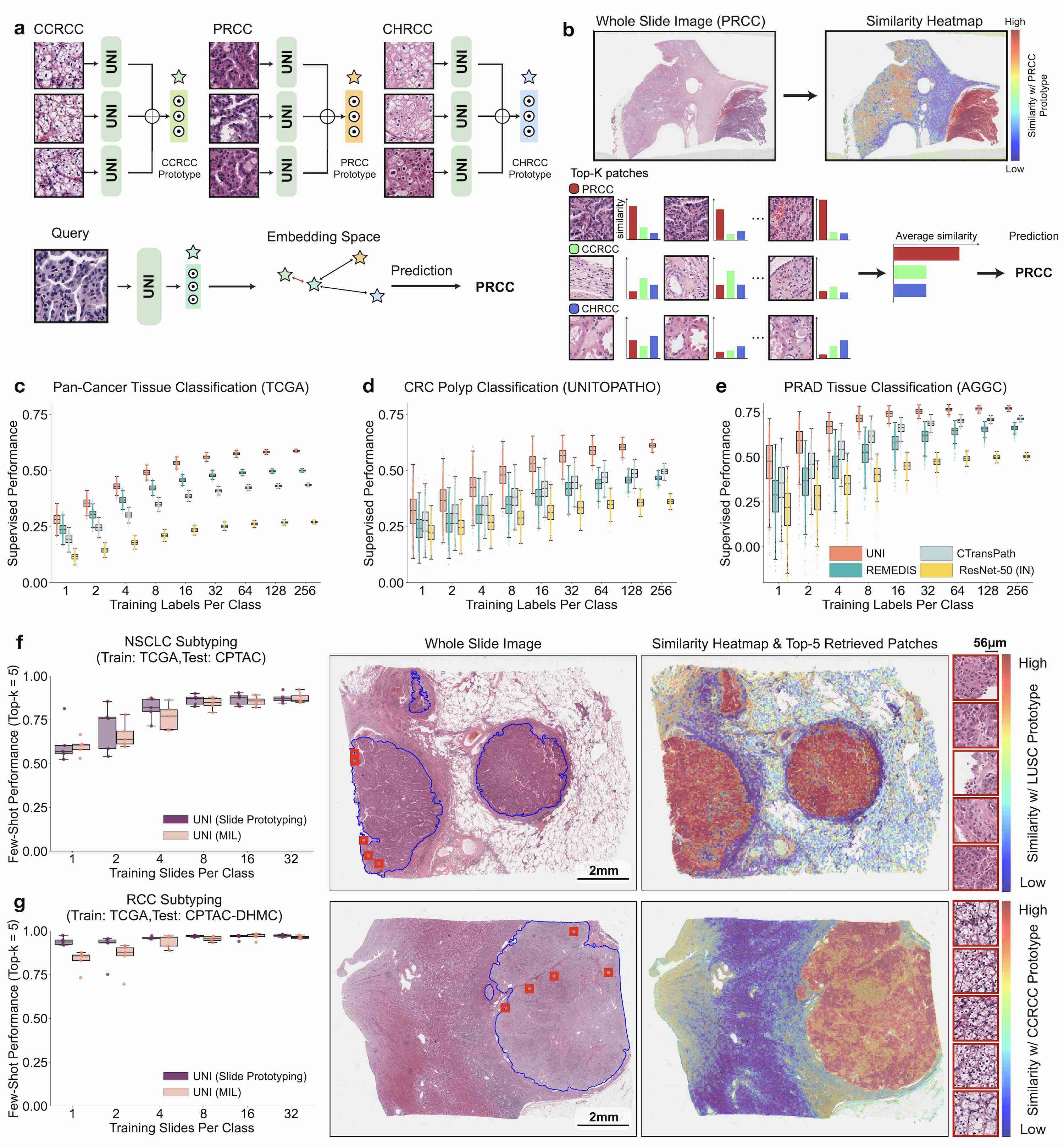}
\caption{\textbf{Few-shot ROI- and slide-level prototyping.} \textbf{a.} Prototypical few-shot ROI classification via SimpleShot. A class prototype is constructed by averaging the extracted features from ROIs of the same class. For a test ROI, SimpleShot assigns the class of the most similar class prototype (smallest Euclidean distance) as the predicted ROI label. \textbf{b.} Prototypical few-shot slide classification via MI-SimpleShot. Using a precomputed set of ROI-level class prototypes (sharing the same class labels as the slide), MI-SimpleShot predicts the slide label using the class prototype with the highest average similarity of top $K$ patches queried from the WSI. The similarity heatmap visualizes the the similarity between the ground-truth class prototype and each patch in WSI. \textbf{c-e.} Few-shot ROI classification performance via SimpleShot on three tasks, with boxes indicating quartile values of model performance ($n=1000$ runs) and whiskers extend to data points within 1.5$\times$ the interquartile range. \textbf{f-g.} Few-shot slide classification performance and similarity heatmaps via MI-SimpleShot on NSCLC subtyping (\textbf{f}) and RCC subtyping (\textbf{g}). In both tasks, using pre-extracted features from \ours, we compare MI-SimpleShot in the same few-shot settings as ABMIL ($K\in\{1,2,4,8,16,32\}$ slides per class, 5 runs), and visualize similarity heatmaps and the top-5 similar patches (indicated in red bounding boxes) for a LUSC (\textbf{f}) and CCRCC (\textbf{g}) slide, with further methodological details, comparisons, visualizations, interpretations provided in the \textbf{Online Methods} and \textbf{Extended Data Figure \ref{fig:patch-level-fs}, \ref{fig:slide-level-proto-fs}, \ref{fig:slide-level-proto-heatmaps}}.}
\label{fig:slide_proto}
\end{figure*}

\noindent Similar to slide-level classification, we also assess the label-efficiency of \ours on ROI-level tasks. We evaluate all pretrained encoders using the non-parametric SimpleShot framework\cite{wang2019simpleshot}, a strong baseline in the few-shot classification literature that proposes averaging extracted feature vectors of each class as the support examples in $K=1$ nearest neighbors (or nearest centroid) classification\cite{snell2017prototypical}. These averaged feature vectors can also be viewed as ``class prototypes", a set of one-shot exemplars that are unique in representing semantic information such as class labels (\textit{e.g.,} LUAD versus LUSC morphologies). At test time, unseen test examples are assigned the label of the nearest class prototype via Euclidean distance (\textbf{Figure \ref{fig:slide_proto}a}). For all pretrained encoders, we evaluate their pre-extracted features using SimpleShot with $K \in \{ 1, 2, 4, 8, \dots, 256 \}$ training examples per class for a majority of tasks (BRCA subtyping, CRC tissue classification in HunCRC, and ESCA subtyping limited to $K=64,128,128$ respectively), with experiments repeated over 1000 runs where $C\cdot K$ training examples are sampled for each run. A detailed description of the SimpleShot framework is provided in the \textbf{Online Methods}, with few-shot performance for all tasks summarized in \textbf{Extended Data Figure \ref{fig:patch-level-fs}}.

We observe similar label efficiency trends as few-shot slide classification in few-shot ROI classification tasks, as \ours outperforms all pretrained encoders across almost all tasks and few-shot evaluation settings. In the 1-shot and 2-shot evaluation of most tasks, though the median value for \ours is generally higher than that of the next best-performing model, the variance in balanced accuracy performance across runs is very high, which can be attributed to poor selection of support examples. However, as the number of support examples increases in forming the class prototypes, we observe a monotonic decrease in variance of few-shot performance runs ($0.32-1.59\%$ standard deviation across tasks in \ours's 256-shot performance), which demonstrates performance stability in permuting training examples to average as class prototypes in SimpleShot. When comparing the median 8-shot performance of \ours with that of other models, \ours consistently exceeds the 128-shot and 256-shot performance of the next best-performing model on many tasks ($16-32\times$ label efficiency), which include challenging tasks such as PRAD tissue classification (AGGC), CRC polyp classification (UniToPatho), pan-cancer tissue classification (TCGA), and other tasks (\textbf{Figure \ref{fig:slide_proto}c-e, Extended Data Figure \ref{fig:patch-level-fs}}). On these tasks, we also observe several instances in which the lowest few-shot performance of \ours exceeds the median and even the maximum few-shot performance reported across 1000 runs of other models. On PRAD tissue classification, the lowest-performing runs for \ours in 32-shot and 128-shot evaluation outperforms the best-performing run possible for CTransPath and REMEDIS, respectively. We observe a similar finding in CRC polyp classification, in which the lowest-performing run for \ours in 64-shot evaluation outperforms the median performance of all few-shot evaluation settings for CTransPath and REMEDIS. In pan-cancer tissue classification, the lowest-performing run for \ours in 2-shot, 8-shot, and 32-shot evaluation outperforms the best possible run possible for ResNet-50, CTransPath, and REMEDIS, respectively. To assess SimpleShot-like evaluation in a fully-supervised setting, we also report results in using all training examples as the support set for forming class prototypes, in which SimpleShot evaluation reaches competitive performance with linear and KNN probing. Overall, these findings demonstrate not only the label efficiency of \ours in discriminative tasks, but also its superior representation quality, such that averaging the extracted features for a few ROIs can create effective class prototypes.


\heading{Prompt-based slide classification using class prototypes.}

\noindent Though weakly-supervised learning via MIL has shifted slide-level classification to no longer requiring patch-level annotations\cite{campanella2019clinical}, accessing and curating histology slide collections may still exist as barriers for clinical tasks that address rare and underrepresented diseases. Moreover, ROI-level analysis is still an important component of many study designs in CPath in further post-hoc interpretability of MIL\cite{chen2022pan,lu2021ai,liang2023deep} and other fine-grained assessment of the tissue microenvironment\cite{saltz2018spatial,abduljabbar2020geospatial,wang2022spatial,leo2021computationally,diao2021human,graham2023screening,jaume2021quantifying,pati2022hierarchical}. From observing the strong retrieval performance and few-shot capabilities in \ours, we re-visit the problem of few-shot slide classification using a semi-supervised, prompt-inspired approach based on class prototypes. Previous works in CPath have demonstrated that textual prompts encoding class labels (known as a class prompt) can be used for ``zero-shot" slide classification\cite{lu2023visual}, in which the class prompt with the best average retrieval score computed using their top-$K$ retrieved patches (top-$K$ average pooling) assigns the slide label. We note that SimpleShot and textual prompting, though using different modalities, perform retrieval-based classification using an embedding space. Though the requirement of annotations for class prototype construction prevents zero-shot learning in the SimpleShot setting, we demonstrate \ours only needs a few examples per class. Similar to textual prompting, we used the class prototypes from SimpleShot as ``prompts" for top-$K$ average pooling of retrieved patches (with class prototypes replacing textual prompts), which we term Multiple Instance SimpleShot (MI-SimpleShot) (\textbf{Figure \ref{fig:slide_proto}b}). We evaluate this approach on two slide-level tasks which have matching ROI training examples from datasets that can be used as the support set - NSCLC subtyping (\textbf{Figure \ref{fig:slide_proto}f}) and RCC subtyping (\textbf{Figure \ref{fig:slide_proto}g}), which can be used in conjunction with the annotated LUAD, LUSC, CCRCC, PRCC, and CHRCC ROIs from the pan-cancer tissue classification task based on TCGA\cite{komura2022universal}. Similar to our evaluation in few-shot slide classification, we evaluate MI-SimpleShot on $\{ 1, 2, 4, 8, 16, 32 \}$ training slides per class using the same 5 folds as the trained ABMIL models, with prototypes created from ROIs annotated within the same training slides. We compare against pre-extracted features of other encoders using MI-SimpleShot, as well as the MIL baseline for \ours. We also develop similarity heatmaps that visualize the normalized Euclidean distances of all patches within a slide with respect to the class prototype of the ground-truth label, with pathologist annotations of tissue regions that match the slide label outlined in blue. A detailed description of the MI-SimpleShot framework is provided in the \textbf{Online Methods}, with further reporting of results provided in \textbf{Extended Data Figure \ref{fig:slide-level-proto-fs},\ref{fig:slide-level-proto-heatmaps}}, and \textbf{Extended Data Table \ref{tab:proto-nsclc},\ref{tab:proto-rcc}}

Using only a few annotated ROI examples per class as prototypes, we demonstrate the potential of applying \ours with MI-SimpleShot as a simple but highly-efficient system for slide-level disease subtyping and detection. On NSCLC and RCC subtyping (trained on TCGA and tested on external cohorts), MI-SimpleShot with top-5 pooling achieves better performance than ABMIL when using 1, 2, and 4 training slides per class for creating prototypes, and achieves similar performance to ABMIL when using more slides (\textbf{Figure \ref{fig:slide_proto}f,g}). From visualization of similarity heatmaps, we also observe that retrieved patches of \ours (corresponding to the slide label) has strong agreement with pathologist annotations, as observed in the right-hand side of \textbf{Figure \ref{fig:slide_proto}f,g} for LUSC and CCRCC slides. We believe that the effectiveness of MI-SimpleShot is due to a combination of: 1) no trainable parameters needed in MI-SimpleShot, whereas ABMIL models may still over- and underfit in few-shot settings, and 2) the strong representation quality of features extracted by \ours. Compared to other pretrained encoders, \ours demonstrates $8\times$ label efficiency on both NSCLC and RCC subtyping (\textbf{Extended Data Figure \ref{fig:slide-level-proto-fs}}). When using all training slides, \ours achieves balanced accuracy scores of $90.2\%$ on NSCLC subtyping, outperforming the next best performing model (CTransPath) by +$5.7\%$ ($p<0.001$). On RCC subtyping, \ours achieves comparable results to the best performing model, REMEDIS ($95.2\%$ vs. $95.7\%$, $p=0.511$). Interestingly, we note that REMEDIS performance via MI-SimpleShot on RCC subtyping outperforms that of ABMIL ($79\%$, $p < 0.001$), which demonstrates the general versatility of MI-SimpleShot when coupled with pretrained encoders. We observe similar trends when using top-50 average pooling, and performance for all MI-SimpleShot experiments (\textbf{Extended Data Figure \ref{fig:slide-level-proto-fs}}, \textbf{Extended Data Table \ref{tab:proto-nsclc}, \ref{tab:proto-rcc}}). Lastly, while REMEDIS, CTransPath, and even ResNet-50$_{\text{IN}}$ can also be used to create similarity heatmaps (\textbf{Extended Data Figure \ref{fig:slide-level-proto-heatmaps}}), we observe instances of label mismatch in the slide prediction and retrieved patches, in which either: 1) retrieved patches of the class prototype (corresponding to the slide label) had poor agreement with the pathologist's annotations, or 2) retrieved patches had strong agreement with the pathologist's annotation but the slide was misclassified. Overall, our comprehensive evaluation of MI-SimpleShot reinforces the strong representation quality of \ours and its potential as a foundational model that can be used in routine clinical tasks.

\Heading{Discussion}

\noindent Given the large range of tasks performed in the practice of anatomic pathology and those enabled by computational pathology (CPath) techniques, transfer learning has been a cornerstone of rapid progress in the field. Many works in CPath have leveraged image encoders trained on a large database of natural images (such as ImageNet) or, more recently and showing better performance, those trained using large public repositories of histopathology data such as the TCGA\cite{ciga2021self, wang2022transformer, azizi2023robust, kang2023benchmarking}. However, there is still room for improvement in the performance of these approaches and further, these studies have generally focused on relatively narrow tasks and disease morphologies, ignoring the full diversity seen in histopathology slides in practice. 

In this study, we demonstrate the versatility of \ours, a general-purpose, self-supervised model pretrained on the largest histology slide collection (for self-supervised learning) to date in CPath. We curated Mass-100K, a large and diverse pretraining dataset containing over 100 million tissue patches from 100,426 whole-slide images (WSIs) across 20 major organ types including normal tissue, cancerous tissue, and other pathologies. The size of our pretraining dataset, coupled with the DINOv2 self-supervised learning approach (demonstrated to scale well to large dataset sizes)\cite{oquab2023dinov2}, allow \ours to significantly outperform other histopathology image encoders across a range of 33 clinical tasks that include a variety of formulations of ROI-level classification, retrieval, segmentation, and slide classification. Additionally, we also demonstrate new capabilities of self-supervised models in CPath, which include resolution-agnostic feature extraction, few-shot slide classification using class prototypes, and disease subtyping generalization of up to 108 labels in the OncoTree classification task. Lastly, as \ours is also trained on mostly in-house histology slides, \ours can be used freely for further development and evaluation of AI models on many public clinical tasks in CPath such as those derived from TCGA.

Our study has several limitations. Based on the plain ViT-Large architecture, \ours lacks vision-specific inductive biases for solving dense prediction tasks in CPath such as cell segmentation, and note that observed performance increases in SegPath are not as drastic as in other tasks. Though \ours still outperforms the next best-performing model (REMEDIS), we envision further improvement as better recipes emerge for adapting plain ViT architectures\cite{chen2022vitadapter}. In addition, our study also does not evaluate the best-performing ViT-Giant architecture in DINOv2, an even larger model that would likely translate well in CPath but remains out of scope due to the enormous computational resources needed for pretraining. Though our study organizes the largest collection of clinical tasks for evaluating pretrained models in CPath (to our knowledge), other clinical tasks such as those in hematopathology are not represented in our analyses. Moreover, \ours is a unimodal model for CPath, with multimodal capabilities such as image captioning, cross-modal retrieval, and zero-shot classification remaining out-of-scope, which we explore in concurrent work\cite{lu2023towards}. Due to these limitations and also following conventional nomenclature of self-supervised models in computer vision\cite{dosovitskiy2021image,oquab2023dinov2}, though \ours demonstrates state-of-the-art results across many clinical applications, further development is needed before a ``visual foundation model" is achieved that would serve all use cases in anatomic pathology.

\end{spacing}

\begin{spacing}{1.35}
\Heading{Online Methods}

\heading{Large-scale visual pretraining}

\noindent In developing and evaluating self-supervised models in CPath, an important and relatively under-discussed challenge is the difficulty in developing large-scale models that can also be used for evaluation on public histology datasets. For natural images, IN-1K is an integral dataset for the model development and evaluation lifecycle of self-supervised learning methods. Specifically, models are first pretrained on the training set of IN-1K and then evaluated with finetuning and linear probe performance on the validation set (treated as the test set) reported as a community-accepted ``goodness-of-fit"\cite{kolesnikov2019revisiting,zhai2019s4l}, with further evaluation of generalization performance via other downstream tasks such as fine-grained classification and activity video recognition. Though such off-the-shelf self-supervised learning methods can readily be adapted to CPath, we note that there is considerably less public data for pretraining in CPath than natural images, and that pretraining on large, public collections of histology slides also restricts their adaptability for public CPath benchmarks. Specifically, development of many self-supervised pathology models have been limited to pretraining on TCGA~\cite{weinstein2013cancer}, one of the largest and most diverse public histology datasets for CPath, with many models opting using the entire TCGA collection in order to realize data scaling benefits in self-supervised learning~\cite{azizi2023robust,wang2022transformer,chen2022scaling}. However, their usability in evaluating on public CPath benchmarks may be restricted to transductive inference\cite{wang2021transpath,wang2022transformer,chen2022scaling,li2021dual,jaume2023modeling,zhao2020predicting,lazard2023giga}, as many popular clinical tasks in CPath are also derived from TCGA (\textit{e.g.} - pan-cancer analyses\cite{fu2020pan,chen2022pan,kather2020pan,saltz2018spatial,komura2022universal,kather2019deep,kaczmarzyk2023champkit,kalra2020yottixel,schmauch2020deep,graham2023one,diao2021human,wulczyn2020deep,riasatian2021fine}) and thus limits extensive evaluation of out-of-domain, generalization performance. Though datasets such as CAMELYON\cite{bejnordi2017diagnostic,bandi2018detection} and PANDA\cite{bulten2022artificial} can be used in evaluating TCGA-pretrained models, we note that these datasets are limited to single tissue types with limited disease categories.


\hheading{Dataset curation for Mass-100K.} To overcome this limitation, we present \textit{Mass-100K}, a large-scale and diverse pretraining dataset comprised of in-house histology slides from Massachusetts General Hospital (MGH) and Brigham \& Women's Hospital (BWH), and external histology slides from the Genotype-Tissue Expression (GTEx) consortium. Following natural image datasets, we also created three partitions of Mass-100K that vary in size in order to evaluate the data scaling laws, an empirical observation found in natural language and image foundation models that scaling dataset size would also increase model performance~\cite{oquab2023dinov2,dosovitskiy2021image,he2022masked,balestriero2023cookbook}. Analogous to IN-22K and IN-1K, we developed the Mass-22K dataset, which contains 16,059,454 histology image patches sampled from 21,444 diagnostic formalin-fixed paraffin-embedded (FFPE) haematoxylin and eosin (H\&E) WSIs across 20 major tissue types comprised of mostly cancer tissue, as well as its subset Mass-1K (1,064,615 images, 1,404 WSIs). All histology slides in Mass-22K and Mass-1K were collected from BWH, and scanned using an Aperio GT450 scanner or a Hamamatsu S210 scanner. To make the image dataset sizes roughly equivalent to that of IN-22K and IN-1K, we sample approximately 800 image patches from histology tissue regions of each WSI, with image resolutions of $256 \times 256$ pixels at $20 \times$ magnification. For slide preprocessing, we adapted the WSI preprocessing in the CLAM toolbox\cite{lu2021data}, which performs: 1) tissue segmentation at a low resolution via binary thresholding of the saturation channel in RGB$\rightarrow$HSV color space, 2) median blurring, morphological closing, and filtering contours below a minimum area to smooth tissue contours and remove artifacts, 3) patch coordinate extraction of non-overlapping 256 $\times$ 256 tissue patches in the segmented tissue regions of each WSI at 20$\times$ magnification. The distribution of Mass-22K and Mass-1K are respectively reported in \textbf{Extended Data Table~\ref{tab:op22k}} and \textbf{Extended Data Table~\ref{tab:op1k}}.

Inspired by even larger natural image datasets such as LVD-142M\cite{oquab2023dinov2} and JFT-300M\cite{sun2017revisiting}, we developed Mass-100K, which combines Mass-22K with further in-house FFPE H\&E histology slide collections (including renal and cardiac transplant tissue) and GTEx\cite{gtex2015genotype} which is comprised of 24,782 non-cancerous, human autopsy WSIs. Additional in-house slides were collected from both BWH and MGH, and scanned using an Aperio GT450 scanner or a Hamamatsu S210 scanner. We purposefully excluded using other public histology slide collections such as TCGA, CPTAC, and PAIP for external evaluation of \ours. Altogether, Mass-100K includes 100,426 histology slides, with it's distribution reported in \textbf{Extended Data Table~\ref{tab:op100k}}. Following the slide preprocessing protocol reported above, sampling approximately 800 histology tissue patches per WSI in Mass-100K yielded 75,832,905 images at 256 $\times$ 256 pixels at $20\times$. For high-resolution finetuning in DINOv2, we sampled an additional 24,297,995 images at 512 $\times$ 512 pixels at $20\times$, which altogether yielded 100,130,900 images for pretraining in Mass-100K.

\hheading{Network architecture and pretraining protocol.} For large-scale visual pretraining on Mass-100K, we used DINOv2\cite{oquab2023dinov2}, a state-of-the-art self-supervised learning method based on student-teacher knowledge distillation for pretraining large ViT architectures. DINOv2 is an extension of two previous methods (DINO\cite{caron2021emerging} and iBOT\cite{zhou2021ibot}) and uses two main loss objectives: self-distillation loss (\textit{i.e.,} alignment loss in \textbf{Figure~\ref{fig:overall}b}) and masked image modeling loss (\textit{i.e.,} reconstruction loss in \textbf{Figure~\ref{fig:overall}b}), which achieves state-of-the-art results in linear probe accuracy. DINOv2 also demonstrates capabilities in understanding the semantic layout of histopathology images when pretrained using knowledge distillation\cite{chen2022scaling}. Self-distillation, introduced in BYOL\cite{grill2020bootstrap} for CNN pretraining and DINO\cite{caron2021emerging} for ViT pretraining, minimizes the predictive categorical distributions from the teacher (\textbf{\ours Teacher} in \textbf{Figure~\ref{fig:overall}b})  and student network (\textbf{\ours} in \textbf{Figure~\ref{fig:overall}b}) obtained from two augmented views of the same image by minimizing their cross-entropy loss. The teacher is updated as an exponential moving average of previous iterations of the student. Masked image modeling using an online tokenizer, introduced in iBOT\cite{zhou2021ibot}, involves strategically masking specific regions within an input image and training the model to predict the masked regions based on the remaining contextual information. This approach captures high-level visual features and context, inspired by masked language modeling in BERT\cite{devlin2018bert}. Specifically, we denote two augmented views of an input image $x$ as $u$ and $v$, which are subsequently randomly masked. The masked images of $u$ and $v$ are represented as $\hat{u}$ and $\hat{v}$, respectively. While $u$ and $v$ are propagated through the teacher network, the student network receives $\hat{u}$ and $\hat{v}$ as inputs. For the self-distillation objective, we compute cross-entropy loss between the \textsc{[CLS]} token from the teacher network and the \textsc{[CLS]} token from the student network. For the masked image modeling objective, DINOv2 utilizes the output of the masked tokens from the student network to predict the patch tokens from the teacher network, where the teacher network can be regarded as an online tokenizer. We used DINOv2 as an important property for pretrained vision models in histopathology is linear probe performance, as these models are often used as frozen feature extractors for pre-extracting patch features in weakly-supervised slide-level tasks. Though other ViT-based self-supervised methods have demonstrated superior finetuning performance\cite{he2022masked,tian2022designing}, their linear probe performance are not comparable and note that full-finetuning in ROI-level and slide-level tasks is not always feasible due to cost in collecting annotations.

Additionally, we note that DINOv2 is an extension of iBOT\cite{zhou2021ibot} which also uses the two loss objectives described above. Both methods are originally based on the original DINO framework\cite{caron2021emerging} (which introduced student-teacher knowledge distillation for ViTs). Specifically, iBOT extends DINO by introducing an online tokenizer component for masked image modeling, and DINOv2 extends iBOT by introducing additional modifications to improve training stability and efficiency for larger ViT architectures. The DINOv2 modifications can be summarized as the following: 1) untying the head weights between the above loss objectives\cite{zhou2021ibot}, 2) Sinkhorn-Knopp centering instead of teacher softmax-centering\cite{zhou2021ibot}, 3) KoLeo regularizer to improve token diversity\cite{sablayrolles2018spreading}, 4) high-resolution finetuning toward the end of pretraining\cite{NEURIPS2019_d03a857a}, 5) an improved implementation of \textsc{FlashAttention}\cite{dao2022flashattention}, stochastic depth, and fully-sharded data parallel mechanisms, and 6) an improved pretraining recipe of the ViT-Large architecture on large-scale datasets. Of these modifications that DINOv2 proposes, we used the default configuration~\footnote{\href{https://github.com/facebookresearch/dinov2/blob/main/dinov2/configs/ssl_default_config.yaml}{https://github.com/facebookresearch/dinov2/blob/main/dinov2/configs/ssl\_default\_config.yaml}} which omits the modifications (1) and (2), as outlined in \textbf{Extended Data Table \ref{tab:hparams_dinov2}}. High-resolution finetuning was conducted on the last 12,500 iterations of pretraining (out of 125,000 iterations total).

\heading{Evaluation Setting}

\hheading{Comparisons \& baselines.} For slide- and ROI-level evaluation, we compare \ours against three pretrained encoders commonly used in the CPath community. As a comparison to models with ImageNet Transfer, we compare against a ResNet-50\cite{he2016deep} pretrained on ImageNet\cite{deng2009imagenet} (truncated after the third residual block, 8,543,296 parameters), which is a commonly-used baseline in many slide-level tasks\cite{lu2021data,chen2021multimodal}. As comparisons to the current state-of-the-art encoders, we compare against CTransPath\cite{wang2022transformer}, which is a Swin Transformer\cite{liu2021swin} using the "tiny" configuration with a window size of 14 (Swin-T/14, 28,289,038 parameters) pretrained mostly on the TCGA via MoCoV3\cite{chen2021mocov3}, and REMEDIS\cite{azizi2023robust}, a ResNet-152$\times$2 (232,230,016 parameters) initialized with the "Big Transfer"-medium protocol\cite{kolesnikov2020big} on ImageNet-22K and then pretrained with SimCLR\cite{chen2020simple}. Regarding data distributions, CTransPath was pretrained using 29,753 WSIs across 25 anatomic sites in TCGA (including both FFPE and frozen tissue slides) and 2,457 WSIs from the Pathology AI Platform (PAIP)\cite{kim2019paip} across 6 anatomic sites, with 15,580,262 tissue patches and 32,120 WSIs used for pretraining altogether. REMEDIS was pretrained with a random sample of approximately $\sim 50$ million patches from 29,018 WSIs also across 25 anatomic sites in TCGA. For self-supervised learning, CTransPath was trained using the MoCoV3\cite{chen2021mocov3} algorithm for 100 epochs with approximately $\sim 1.56 \times 10^9$ (or 1.56 billion) images seen during pretraining, with REMEDIS trained using the SimCLR algorithm for a maximum of 1,000 epochs with upwards of $\sim 50 \times 10^9$ (or 50 billion) images seen during pretraining. In our implementation of these pretrained encoders, we use the truncated ResNet-50 implementation provided by CLAM\cite{lu2021data}, and used the official model checkpoints for CTransPath and REMEDIS. The image embeddings outputted by these models are 1024, 768, and 4096 respectively. Similar to ResNet-50 and other ResNet models in which the penultimate feature layer before the classification head is a grid-like feature map of $[1 \times 7 \times 7 \times 4096]$-dimension, we apply a two-dimensional adaptive average pooling layer to output a single $[1 \times 4096]$-dimensional image embedding. For all images used in ROI tasks and extracted patches for multiple instance learning (MIL) in slide tasks, across all models, all feature extraction operations are performed on resized $224 \times 224$ images, due to constraints in the Swin-T/14 architecture used by CTransPath which can only take image dimensions in which the length is divisible by 224. All pretrained encoders use ImageNet mean and standard deviation parameters for image normalization (including \ours).

Lastly, we note that while many slide and ROI tasks are created using annotated data using the TCGA, CTransPath and REMEDIS were also trained using almost all slides in the TCGA, which can result in information leakage that inflates the performance of these models on TCGA benchmarks. For all tasks, when possible, we report evaluation on external cohorts outside of TCGA. This may not be possible for all tasks, as the official train-validation-test folds may all be developed using TCGA.

\hheading{Weakly-supervised slide classification.} Training and evaluation for weakly-supervised slide classification tasks follows the conventional two-stage multiple instance learning (MIL) paradigm of: 1) pre-extracting ROI-level features as instances from non-overlapping tissue patches of segmented tissue regions of the WSI, 2) learning a trainable permutation-invariant pooling operator that aggregates patch-level (or instance) features into a single slide-level (or bag) feature. For slide preprocessing, we use the same WSI preprocessing pipeline as described in the dataset curation section which uses the CLAM toolbox\cite{lu2021data}, with additional patch feature extraction using a pretrained encoder performed on the patched coordinates. Images are resized down to $224 \times 224$ and normalized using ImageNet mean and standard deviation parameters. As quality control, we performed the additional following steps: 1) for slides with under- or over-segmented tissue masks, we adjust the segmentation parameters in CLAM (threshold value and downsample level) to segment only tissue regions, 2) removal of slides that were non-H\&E and non-FFPE, and 3) for slides that did not have a downsample level equivalent to $20\times$ magnification in their WSI pyramidal format, we patched the tissue into non-overlapping $512 \times 512$ tissue patches at $40\times$ magnification and then later resized these images to $224 \times 224$ during feature extraction. Pre-extracted features for all pretrained encoders used the same set of patch coordinates for feature extraction of each WSI.
 
For comparison of pre-extracted features of pretrained encoders in weakly-supervised learning, we used the Attention-Based Multiple Instance Learning (ABMIL) algorithm\cite{ilse2018attention} across all tasks, which is a canonical weakly-supervised baseline in slide classification tasks. We use the two-layer gated variant of the ABMIL architecture with all input embeddings mapped to an embedding dimension of 512 in the first fully-connected (FC) layer, followed by hidden dimensions of 384 in the following intermediate layers. For regularization, we use dropout with $P=0.10$ applied to the input embeddings and $P = 0.25$ after each intermediate layer in the network. Aside from the first FC layer which is dependent on the embedding dimension of the pre-extracted features, all comparisons used the same ABMIL model configuration. We trained all ABMIL models using the AdamW optimizer\cite{loshchilov2018decoupled} with a cosine learning rate scheduler, learning rate of 1e-4, cross-entropy loss, and maximum 20 epochs. We additionally performed early stopping on the validation loss if a validation fold was available. For all slide classification tasks, we case-stratified and label-stratified the slide dataset into train-validation-test folds, or used official folds if available. As CTransPath and REMEDIS were pretrained using all slides in TCGA, we considered TCGA slide tasks in which additional external evaluation was possible (\textit{e.g.} - NSCLC subtyping was included due to availability of LUAD and LUSC slides in CPTAC, whereas BRCA subtyping was excluded). For glioma \textit{IDH1} mutation prediction and histomolecular subtyping, train-validation-test folds were additionally site-stratified to mitigate potential batch effects.

\hheading{Linear and K-nearest neighbors probe evaluation in ROI classification.} For ROI-level classification tasks, we follow previous works that use logistic regression (linear) probing and K-nearest neighbors (KNN\cite{bentley1975multidimensional}) probing for respectively evaluating discriminative transfer performance and representation quality of pre-extracted feature embeddings on downstream tasks\cite{balestriero2023cookbook}. For linear probing, following the practice recommended by the self-supervised learning community, we fix the $\ell_2$ regularization coefficient $\lambda$ to $\frac{100}{MC}$, where $M$ is the embedding dimension and $C$ is the number of classes, and use the L-BFGS solver\cite{zhu1997algorithm} with a maximum of 1000 iterations\cite{kolesnikov2019revisiting}. KNN probing is an additional evaluation technique advocated by the self-supervised community for measuring representation quality of pre-extracted features\cite{caron2021emerging, mu2022slip, li2022understanding, lee2022unsupervised, sar2023no, fang2020seed}. In comparison with linear probing, KNN probing is non-parametric (aside from the choice of $K$), as it classifies unseen test examples based on only their feature similarity with labeled training examples (\textit{e.g.} similar examples in representation space should also be visually similar and share the same class label). We use the KNN implementation from Scikit-Learn\cite{pedregosa2011scikit} trained fitted using $K=20$ and Euclidean distance as the distance metric, following observed stability of this evaluation setup of other self-supervised works\cite{caron2021emerging}. For all ROI tasks, we approximately case-stratified and label-stratified datasets into train-test folds, or used official folds if available.

For all tasks, we resize images to $224 \times 224$ (or $448 \times 448$ if available) and normalize using ImageNet mean and standard deviation parameters. Additionally, we note that many ROI datasets consist of images with high image resolutions, with image resizing to a fixed $224 \times 224$ or $448 \times 448$ resolution also changing the image magnification and microns per pixel (mpp). For example, resizing ROIs in the CRC polyp classification task in UniToPatho (ROIs having an original image resolution of $1812 \times 1812$ at 0.45 mpp) to $224 \times 224$ would change the magnification to 3.6 mpp. For CRC polyp classification as well as BRCA subtyping (BACH), we evaluate on resized image resolutions of $\{224^2, 448^2, 896^2, 1792^2\}$ and $\{224^2, 448^2, 896^2, 1344^2\}$, with multiples of 224 chosen due to due to constraints with CTransPath. To pre-extract features from high-resolution images, for ViTs such as the plain ViT-large architecture in \ours and the hierarchical Swin Transformer-T architecture in CTransPath, the forward passes of these architectures are not modified, interpolation of positional embeddings is performed to have the same sequence length as patch tokens in the ROI. To illustrate, in the patch embedding layer of our ViT-Large architecture in \ours that has a patch token size of $16 \times 16$, a $224 \times 224$ image would be converted into a $[ 14 \times 14 \times D ]$-dimension 2D grid of patch embeddings using a 2D convolutional layer (kernel and stride size of 16, 3 incoming channels from RGB-input image inputs and $D$-dim outgoing channels set as a hyper-parameter for feature embedding length), followed by flattening and transposing (now a [$196 \times D$]-dimension sequence of patch embeddings), which can now be used in Transformer attention (called ``patchifying"). For a $1792 \times 1792$ image in CRC polyp classification, patchifying this image using the same patch embedding layer would result in a $[ 112 \times 112 \times D ] \rightarrow [ 12544 \times D ]$-dimension sequence of patch embeddings. Feeding this sequence into the forward pass of Transformer attention, though computationally expensive, is still tractable via memory-efficient implementations such as \textsc{FlashAttention} or \textsc{MemEffAttention}~\footnote{\href{https://pytorch.org/docs/master/generated/torch.nn.functional.scaled_dot_product_attention.html}{https://pytorch.org/docs/master/generated/torch.nn.functional.scaled\_dot\_product\_attention.html}}. For positional embedding interpolation, we used the implementation provided in the DINO codebase~\footnote{\href{https://github.com/facebookresearch/dino/blob/main/vision_transformer.py\#L174}{https://github.com/facebookresearch/dino/blob/main/vision\_transformer.py\#L174}}. For multi-head self-attention (MHSA) visualization, we visualize the weights from the last attention layer using the notebook implementation provided by the HIPT codebase\cite{chen2022scaling}~\footnote{\href{https://github.com/mahmoodlab/HIPT/tree/master/HIPT\_4K}{https://github.com/mahmoodlab/HIPT/tree/master/HIPT\_4K}}, which we note is only applicable for plain VIT architectures.

\hheading{ROI retrieval.} To assess the quality of embeddings produced by different encoders for CBIR of histopathology images, we use ROI-level classification datasets, in which the goal is to retrieve similar images (\textit{i.e.,} images with the same class label) to a given query image. For each benchmark, we first embed all images into a low-dimensional feature representation using the pretrained encoders. We treat each image in the test set as a query. Each query image is compared to each image from the ROI-level classification training set, which serves as a database of candidates (keys). Note that no supervised learning takes place in these experiments and the class labels are only used for evaluation purposes (i.e., to assess whether retrieved images share the same class label as the query). We first center the database of keys by subtracting their Euclidean centroid from each embedding followed by $\ell_2$ normalization of each key to unit length. For each new query, we apply the same shift and normalization steps, and then measure it against each key in the database via the $\ell_2$ distance metric, where lower distance is interpreted as higher similarity. The retrieved images are sorted by their similarity scores and their corresponding class labels are used to evaluate the success of a given retrieval using Acc@K for K $\in {1,3,5}$ and MVAcc@5, which are described in \textbf{Evaluation metrics}.

\hheading{ROI-level cell type segmentation.} For training and evaluation of ROI-level cell type segmentation tasks, we follow previous works in using Mask2Former, which is a flexible framework commonly used for evaluating off-the-shelf performance of pretrained vision encoders\cite{cheng2021mask2former}. In the case of the ViT architecture which is non-hierarchical, we additionally use the ViT-Adapter framework alongside the Mask2Former head\cite{chen2022vitadapter}. For both ViT-Adapter and Mask2Former, we use the same hyper-parameters used for ADE20k semantic segmentation. More specifically, we use the AdamW\cite{loshchilov2018decoupled} optimizer along with a step learning rate schedule. The initial learning rate was set to 0.0001, and a weight decay of 0.05 is applied. To adjust the learning rate specifically for the backbone, we apply a learning rate multiplier of 0.1. Additionally, we decay the learning rate by a factor of 10 at 0.9 and 0.95 fractions of the total number of training steps. For all backbones, we finetune the full model for 50 epochs with a batch size of 16. The model's performance on the validation set is evaluated every 5 epochs, and the optimal model based on validation performance is saved for testing. To augment the data, we use the large-scale jittering (LSJ) augmentation\cite{Du2021SimpleTS, ghiasi2021simple}, with a random scale sampled from a range of 0.5 to 2.0, followed by a fixed size crop to 896 $\times$ 896 to accommodate the size constraints of CTransPath. At inference time, we resize the image dimensions to their nearest multiples of 224.

\hheading{Few-shot ROI classification and prototype learning.} For few-shot classification, we follow previous works that use the SimpleShot framework for evaluating the few-shot learning performance of prototypical representations of self-supervised models\cite{wang2019simpleshot,el2023learning}. Prototypical (or prototype) learning is a longstanding task in the few-shot learning community,\cite{snell2017prototypical,allen2019infinite,koch2015siamese,vinyals2016matching,edwards2016towards}, and has been also posed (in many related forms) in CPath as well\cite{yu2023prototypical,vu2023handcrafted,yu2023slpd,quiros2022self,yang2021towards}. In contrast with traditional few-shot learners based on meta learning, SimpleShot and related works demonstrate that strong feature representations combined with specific transformations and simple classifiers can reach state-of-the-art performance on few-shot tasks\cite{wang2019simpleshot,tian2020rethinking,el2023learning}. SimpleShot is similar to nearest neighbors classification, in which the training set (called "supports" in few-shot learning literature) is drawn from $C$ classes (``ways") with $K$ examples per class (``shots") for predicting unseen images in the test set (``queries"). Instead of nearest neighbors, SimpleShot uses a nearest-centroid approach based on ProtoNet\cite{snell2017prototypical}, in which the average feature vector (centroid) for each class is used as a prototypical "one-shot" example for labeling the query set via distance similarity. As mentioned, these averaged feature vectors can also be viewed as ``class prototypes", a set of one-shot exemplar examples that are unique in representing semantic information such as class labels (\textit{e.g.,} LUAD versus LUSC morphologies). As SimpleShot is a simple and surprisingly strong baseline in the few-shot learning community and popularized in evaluating self-supervised models\cite{wang2022transductive,xu2022alleviating,zhu2023transductive,fei2021z,stojanov2021using,el2023learning}, we adopt this baseline in evaluating \ours and its comparisons in few-shot ROI classification tasks. We follow the recommendations in SimpleShot that suggest centering (subtracting the mean computed on the support set) and $\ell_2$ normalizing the support set before computing the class prototypes, with the query set also transformed (also centered using the mean of the the support set) before nearest centroids classification.

Conventional few-shot learners on natural image classification tasks are evaluated by drawing 10,000 $C$-way, $K$-shot episodes from the training set with 15 query images per class as the test set. For equivalent comparison with metrics in linear and KNN probing, we instead draw 1000 $C$-way, $K$-shot episodes but use all images in the test set per episode. Due to the relatively larger number of training examples available in ROI tasks than that of slide tasks, we vary the number of labeled examples per class from $K \in \{1,2,4,8,16,32,\dots 256\}$ or the maximum number of labeled examples available for a given class. To compare with linear and KNN probing that use all training examples, we also evaluate SimpleShot by averaging all training examples per class, which we denote as ``1-NN" in \textbf{Extended Data Tables \ref{tab:patch-crc100k-knn}-\ref{tab:patch-tcga-tils-knn}}.

\hheading{Prompt-based slide classification using Multiple Instance (MI) SimpleShot.} To evaluate the quality of extracted representations in serving as class prototype for slide classification tasks, we adapt class prototypes from SimpleShot (described above) as ``prompts" (similar to usage of textual prompts in zero-shot classification\cite{lu2023visual}, which we describe as Multiple Instance SimpleShot (MI-SimpleShot). As described in the main text, we use two slide-level datasets (NSCLC and RCC subtyping datasets) which have matching ROI training examples from datasets that can be used as the support set. Concretely, we use the annotated LUAD and LUSC ROIs from the TCGA Uniform Tumor dataset for NSCLC subtyping, and annotated CCRCC, PRCC, and CHRCC ROIs from the TCGA Uniform Tumor dataset for RCC subtyping. The TCGA Uniform Tumor dataset (described further in the \textbf{Online Methods}) consists of 271,170 $256 \times 256$ ROIs at around 0.5 mpp of 32 cancer types annotated and extracted from 8,736 H\&E FFPE diagnostic histopathology WSIs. We note that the number annotated ROIs per slide range from 10 to 70 examples in the TCGA-LUAD, -LUSC, -CCRCC, -PRCC, and -CHRCC cohorts. For each class, we first embed ROIs in the support set into a low-dimensional feature representation using the pretrained encoders, followed by average-pooling of all ROI features within the class. The average-pooled feature representations are considered as the class prototypes, which are used as ``prompts" for labeling the top $K$ ROIs for each slide in the query set via normalized Euclidean distance similarity. The slide-level prediction is then made by majority voting of the top $K$ ROI predictions. For each benchmark, we evaluate MI-SimpleShot with both top-5 average pooling and top-50 average pooling and on $\{ 1, 2, 4, 8, 16, 32 \}$ training slides per class similar to our evaluation in few-shot slide classification using the same 5 folds as the trained ABMIL models, with prototypes created from the annotated ROIs within the same training slides. We note little performance change in considering the average scores of the top-5 and top-50 patches per class prototype. To compare with the performance that uses all training slides with ROI annotations, we also evaluate MI-SimpleShot by averaging all training ROI feature representations per class, with results detailed in \textbf{Extended Data Table \ref{tab:proto-nsclc}, \ref{tab:proto-rcc}}. To create similarity heatmaps, we visualize the normalized Euclidean distances of all patches within a slide with respect to the ground-truth class prototype.



\hheading{Evaluation metrics.} We report balanced accuracy, weighted F1 score, and AUROC for classification tasks. \textbf{Balanced accuracy} is computed by taking the unweighted average of the recall of each class, which takes into account class imbalance in the evaluation set. \textbf{Weighted F1 score} is computed by averaging the F1 score (the harmonic mean of precision and recall) of each class, weighted by the size of its respective support set. \textbf{AUROC} is the area under the receiver operating curve plotting true positive rate against the false positive rate as the classification threshold is varied. Additionally, we compute \textbf{quadratic weighted Cohen's $\kappa$} (inter-annotator agreement between two sets of labels, \textit{e.g.} - ground-truth and predictions) which we perform for ISUP grading (PANDA), and \textbf{top-K accuracy} for K $\in \{1,3,5\}$ (for a given test sample, a sample is scored correctly if the ground-truth label is among the top-K labels predicted) for OT-43 and OT-108. For retrieval, we consider \textbf{Acc@K} for K $\in \{1,3,5\}$, which represent the standard top-K accuracy scores in retrieving images with the same class label as the query. Specifically, a retrieval is considered successful if at least one image among the top-K retrieved images have the same class label as the query. We also report \textbf{MVAcc@5}, which compared to Acc@5 more strictly enforces that the majority vote of the top 5 retrieved images must be the same class as the query for retrieval to be considered successful. For segmentation, we report the \textbf{Dice score} (same definition as the F1 score), the precision and recall, macro averaged across all images and classes. 

\noindent\textbf{Statistical analysis}\\
For all semi- and fully-supervised experiments, we estimate 95\% confidence intervals for the model performance with non-parametric bootstrapping using 1,000 bootstrap replicates. For statistical significance, we use a two-sided paired permutation test with 1,000 permutations to assess observed differences between the performance of two models. For all few-shot settings, we report results using box plots that indicate quartile values of model performance ($n=5$ runs) with whiskers extending to data points within 1.5$\times$ the interquartile range. For ROI-level few-shot classification, for each $C$-way, $K$-shot setting, we randomly sample $K$ training examples per $C$ classes with 1,000 repeated experiments (called ``episodes" or ``runs") evaluated on the entire test set. For slide-level few-shot classification, we follow the same setting as above but with the number of runs limited to 5 due to small support sizes in rare disease categories.

\heading{Tasks and datasets}

\hheading{OncoTree Cancer Classification based on in-house BWH data (43 cancer types, 108 OncoTree codes)}: As described in the main text, OncoTree cancer classification is a large-scale hierarchical classification task for CPath that follows the OncoTree (OT) cancer classification system\cite{kundra2021oncotree}. This task was devised to assess generalization capabilities of pretrained models on classifying diverse disease categories and tissue types. Using in-house BWH slides, we defined a dataset that comprises 5,564 WSIs from 43 cancer types further subdivided into 108 OncoTree codes, with at least 20 WSIs per OncoTree code. The dataset forms the basis of two tasks that vary in diagnostic difficulty: 1) 43-class cancer type classification (OT-43) and 2) 108-class OncoTree code classification (OT-108). Due to small support sizes for several OncoTree codes in OT-108, all ABMIL models were trained using train-test folds and without early stopping. For training and evaluation, we approximately label-stratified the dataset into 71:29 train-test folds (3944:1620 slides) using the same folds for OT-43 and OT-108, with a 15 slides used per OncoTree code in the test set and a minimum of 5 slides used per OncoTree code in the training set. The hierarchical classification of the coarse- and fine-grained task is reported in \textbf{Extended Data Table \ref{tab:op43/108}}. We note that slides in the training fold of OT-43 and OT-108 were included in OP-1K and OP-22K pretraining, with the test set held out from these pretraining sources (following practices in ImageNet).

Due to storage limitations in repeatedly extracting features for all non-overlapping tissue patches per WSI for all pretrained models (including intermediate checkpoints), we sampled 200 representative patches per WSI for feature extraction. To select these patches, we first extracted ResNet-50$_{\text{IN}}$ features followed by clustering\cite{lloyd1982least}, employed previously in other works such as WSISA\cite{zhu2017wsisa}, DeepAttnMISL\cite{yao2019deep,yao2020whole}, and others\cite{li2018graph}. We note that these works are inspired by visual bag-of-words (vBOW)\cite{sivic2003video,fei2005bayesian}, which has been adapted to pathology for formulating high-resolution ROIs and WSIs as smaller but representative collections of tissue patches via clustering applied to deep features\cite{cruz2011visual,xu2014weakly}, with downstream applications such as MIL\cite{zhu2017wsisa,li2018graph,yao2019deep,yao2020whole} and retrieval\cite{kalra2020yottixel,chen2022fast}. For all pretrained encoders, we extract features from the same sampled collection of patches. Though additional computational steps were taken to derive these sampled patches, we note that this does not fall under transductive inference, as the entire test set (all WSI samples) is never made visible to any learning component (clustering is fitted per WSI, with ``samples" defined at the slide-level instead of the patch-level). To validate that this approach has comparable performance using features for all tissue patches per WSI, we compare the performance of sampled versus full features of \ours, CTransPath, REMEDIS, and ResNet-50$_{\text{IN}}$, which we also report in \textbf{Extended Data Table \ref{tab:ot-43-compare} and \ref{tab:ot-108-compare}}. We observe not only marginal performance decrease when using sampled features (maximum decrease of -$0.9$\% in top-1 accuracy, -$0.007$ in AUROC), but also performance increases for many models. For REMEDIS, we observe that the performance of ABMIL models collapses when using full features, with top-1 accuracy performances of 4.0\% and 11.8\% respectively on OT-43 and OT-108 (compared to 59.3\% and 41.2\% respectively with sampled features). We hypothesize that these performance increases are due to the difficult nature of OT-43 and OT-108, with patch sampling reducing the input data complexity for ABMIL (\textit{e.g.} - instead of finding diagnostically-relevant features in a bag of $10,000$+ patches, only $200$ representative patches are considered).

\hheading{Breast Metastasis Detection based on CAMELYON16 (2 classes)}\cite{bejnordi2017diagnostic}: The breast metastasis detection task from the Cancer Metastases in Lymph Nodes Challenge 2016 (CAMELYON16) consists of 400 H\&E FFPE histopathology WSIs of sentinel lymph node from Radboud University Medical Center and the University Medical Center Utrecht for metastasis detection. We removed 1 slide from the test set that was mislabeled, resulting in 399 slides (239 normal, 160 metastasis). For training and evaluation, we used the official train-test folds, and label-stratified the training set into 90:10 train-validation, resulting in 61:7:32 train-validation-test folds (243:27:129 slides).

\hheading{NSCLC Subtyping based on TCGA and CPTAC (LUAD vs. LUSC, 2 classes)}\cite{luadlusc_campbell2016distinct,luad_cptac_gillette2020proteogenomic,lusc_cptac_satpathy2021proteogenomic}: The NSCLC subtyping task consists of non-small cell lung cancer (NSCLC) H\&E FFPE diagnostic histopathology WSIs sourced from TCGA and CPTAC for classifying two subtypes: primary lung adenocarcinoma (LUAD) and lung squamous cell carcinoma (LUSC) cases. For quality control, in TCGA, we excluded slides with missing or incorrect metadata, which resulted in 1,041 slides (529 LUAD and 512 LUSC). In CPTAC, we excluded slides that were frozen tissue, non-tumor tissue or were not labeled as having acceptable tumor segments, which resulted in 1,091 slides (578 LUAD and 513 LUSC). For training and evaluation, we label-stratified the TCGA-NSCLC cohort into 80:10:10 train-validation-test folds (848:97:98 slides), with external evaluation using the held-out CPTAC cohort.

\hheading{RCC Subtyping based on DHMC (CCRCC vs. PRCC vs. CHRCC vs. ROCY vs. Benign, 5 classes)}\cite{zhu2021development}: The RCC subtyping task consists of 563 renal cell carcinoma (RCC) H\&E FFPE diagnostic histopathology WSIs (485 resections and 78 biopsies) from the Dartmouth-Hitchcock Medical Center (DHMC) for classifying five subtypes: primary clear cell renal cell carcinoma (CCRCC, 344 slides), papillary renal cell carcinoma (PRCC, 101 slides) and chromophobe renal cell carcinoma (CHRCC, 23 slides), renal oncocytomas (ROCY, 66 slides), and benign cases (29 slides). For training and evaluation of both tasks, we used a modified configuration of the train-validation-test folds with a 70:4:26 ratio (393:23:147 slides), with 8 CHRCC cases moved from the test to train fold due to CHRCC being absent in the train fold.

\hheading{RCC Subtyping based on TCGA, DHMC, and CPTAC (CCRCC vs. PRCC vs. CHRCC, 3 classes)}\cite{ccrcc_cancer2013comprehensive,prcc_cancer2016comprehensive,chrcc_davis2014somatic,ccrcc_cptac_li2023histopathologic,zhu2021development}: The RCC subtyping task consists of 1,794 renal cell carcinoma (RCC) H\&E FFPE diagnostic histopathology WSIs from TCGA, DHMC, CPTAC for classifying three subtypes: primary clear cell renal cell carcinoma (CCRCC), papillary renal cell carcinoma (PRCC) and chromophobe renal cell carcinoma (CHRCC). For quality control, in TCGA, we excluded slides with missing low-resolution downsamples, which resulted in 922 slides (519 CCRCC, 294 PRCC and 109 CHRCC). In DHMC, we filtered out oncocytomas in the previously-described DHMC-Kidney cohort, which resulted in 468 slides (344 CCRCC, 101 PRCC and 23 CHRCC). In CPTAC, we excluded slides that were frozen tissue, non-tumor tissue or were not labeled as having acceptable tumor segments, which resulted in 404 slides (404 CCRCC). For training and evaluation, we label-stratified the TCGA-NSCLC cohort into 80:10:10 train-validation-test folds (736:89:97 slides), with external evaluation on the held-out DHMC and CPTAC cohorts. As CPTAC only includes CCRCC cases, we combined DHMC and CPTAC into a single evaluation cohort.

\hheading{CRC Screening based on HunCRC (4 classes)}\cite{pataki2022huncrc}: The CRC screening task consists of 200 H\&E FFPE diagnostic histopathology WSIs of colorectal biopsies from the Hungarian Colorectal Cancer Screening (HunCRC) dataset from Semmelweis University. Within this dataset, we defined a 4-way coarse-grained subtyping task into the categories of negative (10 slides), non-neoplastic lesion (38 slides), CRC (46 slides), and adenoma (106 slides), in which the ground-truth label was set by the study's pathologist. For training and evaluation, we label-stratified the HunCRC slide dataset into 50:25:25 train-validation-test folds (158:21:21 slides).

\hheading{BRCA Coarse- and Fine-Grained Subtyping based on BRACS (3 and 7 classes)}\cite{brancati2021bracs}: The BRCA coarse- and fine-grained subtyping tasks consists of 547 breast carcinoma H\&E slides from 187 patients sourced from the Breast Carcinoma Subtyping (BRCA) task sourced from IRCCS Fondazione Pascale, The Institute for High Performance Computing and Networking (ICAR) of National Research Council (CNR), and IBM Research-Zurich. Within this dataset, we defined a 3-way coarse-grained subtyping task using the "benign tumor", "atypical tumor", and "malignant tumor" labels. Furthermore, we define a 7-way fine-grained subtyping task that subtypes benign tumors further as "normal", "pathological benign", "usual ductal hyperplasia", atypical tumors as "flat epithelial atypia" and "atypical ductal hyperplasia", and malignant tumors as "ductal carcinoma in situ" and "invasive carcinoma". The hierarchical classification of the coarse- and fine-grained tasks is reported in \textbf{Extended Data Table \ref{tab:brca-label}}. For training and evaluation of both tasks, we used the official train-validation-test folds with a 72:12:16 ratio (395:65:87 slides), using the same folds for both coarse- and fine-grained tasks.

\hheading{Glioma \textit{IDH1} Mutation Prediction and Histomolecular Subtyping based on TCGA and EBRAINS (2 and 5 classes}\cite{gbm_brennan2013somatic,lgg_cancer2015comprehensive,roetzer2022digital,ebrains_doi}: The glioma \textit{IDH1} mutation prediction and histomolecular subtyping tasks consists of 1,996 H\&E FFPE diagnostic histopathology WSIs from glioblastoma, astrocytoma, and oligodendroglioma cases with molecular status from the TCGA and the EBRAINS Digital Tumor Atlas. We first defined a 5-way glioma histomolecular subtyping task with the following labels: \textit{IDH1}-mutant Astrocytomas (257 slides), \textit{IDH1}-mutant Glioblastomas (93 slides), \textit{IDH1}-mutant and 1p/19q -codeleted Oligodendrogliomas (408 slides), \textit{IDH1}-wildtype Glioblastomas (1094 slides), and \textit{IDH1}-wildtype Astrocytomas (144 slides). Additionally, we defined a simpler 2-way task that only predicts \textit{IDH1} status: \textit{IDH1}-wildtype (1,238 slides) and \textit{IDH1}-mutant (756 slides). The hierarchical classification of the coarse- and fine-grained tasks is reported in \textbf{Extended Data Table \ref{tab:ebrains-mut-label}}. For training and evaluation of both tasks, we approximately label-stratified the TCGA-GBMLGG dataset into a train-validation-test fold with a 47:22:31 ratio (525:243:355 slides), with external evaluation using the held-out EBrains cohort (873 slides), using the same folds for both coarse- and fine-grained tasks.

\hheading{Brain Tumor Coarse- and Fine-Grained Subtyping based on EBRAINS (12 and 30 classes)}\cite{roetzer2022digital,ebrains_doi}: The brain tumor coarse- and fine-grained subtyping tasks consists of 2,319 H\&E FFPE diagnostic histopathology WSIs from the EBRAINS Digital Tumor Atlas sourced from the University of Vienna. With an original dataset size of 3,114 slides, we defined a 30-way fine-grained brain tumor subtyping task limited to diagnostic labels that have at least 30 slides: 1) \textit{IDH1}-wildtype Glioblastoma (474 slides), 2) Pilocytic Astrocytoma (173 slides), 3) Meningothelial Meningioma (104 slides), 4) Pituitary Adenoma (99 slides), 5) \textit{IDH1}-mutant and 1p/19q codeleted Anaplastic Oligodendroglioma (91 slides), 6) Ganglioglioma (88 slides), 7) Haemangioblastoma (88 slides), 8) Adamantinomatous Craniopharyngioma (85 slides), 9) \textit{IDH1}-mutant and 1p/19q codeleted Oligodendroglioma (85 slides), 10) Atypical Meningioma (83 slides), 11) Schwannoma (81 slides), 12) \textit{IDH1}-mutant Diffuse Astrocytoma (70 slides), 13) Transitional Meningioma (68 slides), 14) Diffuse Large B-Cell Lymphoma of the CNS (59 slides), 15) Gliosarcoma (59 slides), 16) Fibrous Meningioma (57 slides), 17) Anaplastic Ependymoma (50 slides), 18) \textit{IDH1}-wildtype Anaplastic Astrocytoma (47 slides), 19) Metastatic Tumours (47 slides), 20) \textit{IDH1}-mutant Anaplastic Astrocytoma (47 slides), 21) Ependymoma (46 slides), 22) Anaplastic Meningioma (46 slides), 23) Secretory Meningioma (41 slides), 24) Lipoma (38 slides), 25) Haemangiopericytoma (34 slides), 26) \textit{IDH1}-mutant Glioblastoma (34 slides), 27) Non-WNT/Non-SHH Medulloblastoma (32 slides), 28) Langerhans Cell Histiocytosis (32 slides), 29) Angiomatous Meningioma (31 slides), and 30) Haemangioma (30 slides). From the same 2,319 slide dataset in the fine-grained task, we also defined a 12-way coarse-grained brain tumor subtyping task that groups the above labels into the following categories: 1) Adult-Type Diffuse Gliomas (837 slides), 2) Meningiomas (430 slides), 3) Mesenchymal, Non-Meningothelial Tumours Involving The CNS (190 slides), 4) Tumours Of The Sellar Region (184 slides), 5) Circumscribed Astrocytic Gliomas (173 slides), 6) Ependymal Tumours (96 slides), 7) Haematolymphoid Tumours Involving The CNS (91 slides), 8) Glioneuronal And Neuronal Tumours (88 slides), 9) Cranial And Paraspinal Nerve Tumours (81 slides), 10) Paediatric-Type Diffuse Low-Grade Gliomas (70 slides), 11) Metastatic Tumours (47 slides) and 12) Embryonal Tumors (32 slides). The hierarchical classification of the coarse- and fine-grained tasks is reported in \textbf{Extended Data Table \ref{tab:ebrains-diagnosis-label}}. For training and evaluation of both tasks, we approximately label-stratified the dataset into a train-validation-test fold with a 50:25:25 ratio (1,151:595:573 slides), using the same folds for both coarse- and fine-grained tasks.

\hheading{ISUP Grading based on PANDA (6 classes)}\cite{bulten2020automated,bulten2022artificial}: The ISUP grading task is derived from the Prostate Cancer Grade Assessment (PANDA) challenge, which comprises 10,616 prostate cancer core needle biopsies of prostate cancer sourced from the Radboud University Medical Center and the Karolinska Institute. Each slide is assigned an ISUP score that defines prostate cancer grade (6-class grading task). For quality control, we follow prior work\cite{pati2023weakly} in excluding slides which were erroneously annotated~\footnote{\href{https://www.kaggle.com/competitions/prostate-cancer-grade-assessment/discussion/169230}{www.kaggle.com/competitions/prostate-cancer-grade-assessment/discussion/169230}} or had noisy labels~\footnote{\href{https://www.kaggle.com/competitions/prostate-cancer-grade-assessment/discussion/169230}{www.kaggle.com/competitions/prostate-cancer-grade-assessment/discussion/169230}}, which resulted in 9,555 slides (2,603 G0, 2,399 G1, 1,209 G2, 1,118 G3, 1,124 G4, 1,102 G5). For training and evaluation, we label-stratified PANDA into 80:10:10 train-validation-test folds (7,647:954:954 slides).

\hheading{Endomyocardial Assessment based on in-house BWH data (2 classes)}\cite{lipkova2022deep}: BWH-EMB dataset consists of 5,021 H\&E FFPE histopathology WSIs from 1,688 in-house endomyocardial biopsies (EMB) collected from Brigham \& Women's Hospital for cellular-mediated allograft rejection (ACR) (2,444 ACR, 2,577 others). For training and evaluation, we case- and label-stratified the dataset into 70:10:20 train-validation-test folds (3,547:484:900 slides). 

\hheading{CRC Tissue Classification based on CRC-100K (9 classes)}\cite{kather2019predicting, kather_jakob_nikolas_2018_1214456}: The CRC tissue classification task is based on the CRC-100K dataset, which consists of 107,180 $224 \times 224$ ROIs at 0.5 mpp annotated and extracted from H\&E FFPE diagnostic histopathology WSIs of 136 colorectal adenocarcinoma samples from the National Center for Tumor Diseases (NCT) biobank and the University Medical Center Mannheim (UMM) pathology archive. ROIs were labeled with the following 9 classes: adipose (11,745 ROIs), background (11,413 ROIs), debris (11,851 ROIs), lymphocyte (12,191 ROIs), mucus (9,931 ROIs), smooth muscle (14,128 ROIs), normal colon mucosa (9,504 ROIs), cancer-associated stroma (10,867 ROIs) and colorectal adenocarcinoma epithelium (15,550 ROIs). For training and evaluation, we used the official case-stratified train-test folds (100,000:7,180 ROIs), with the training fold constructed from 100,000 ROIs (86 WSIs) from the NCT biobank and UMM pathology archive (referred to as ``NCT-CRC-HE-100K"), and test fold constructed from 7,180 ROIs (50 WSIs) from the NCT biobank (referred to as``CRC-VAL-HE-7K"). Additionally, we use the version of NCT-CRC-HE-100K without stain normalization. We use the same folds for linear probe, KNN, and SimpleShot evaluation. We evaluate this dataset on ROIs of $224 \times 224$ image resolution at 0.5 mpp.

\hheading{CRC Tissue Classification based on HunCRC (9 classes)}\cite{pataki2022huncrc}: The CRC tissue classification task is based on the HunCRC dataset, which consists of 101,398 $512 \times 512$ ROIs at 0.48 mpp, annotated and extracted from the same 200 H\&E FFPE diagnostic histopathology WSIs of colorectal biopsies also described in the slide-level task. ROIs were labeled with the following 9 classes: adenocarcinoma (4,315 ROIs), high-grade dysplasia (2,281 ROIs), low-grade dysplasia (55,787 ROIs), inflammation (763 ROIs), tumor necrosis (365 ROIs), suspicious For invasion (570 ROIs), resection edge (534 ROIs), technical artifacts (3,470 ROIs) and normal (31,323 ROIs). For training and evaluation, we case-stratified and approximately label-stratified the dataset into train-test folds (151:49 cases, 76,753:22,655 ROIs), which was used in linear probe, KNN, and SimpleShot evaluation. We evaluate this dataset on resized ROIs of $448 \times 448$ image resolution at 0.55 mpp.

\hheading{BRCA Subtyping based on BACH (4 classes)}\cite{aresta2019bach}: The BRCA subtyping task is based on the Breast Carcinoma Subtyping (BACH) dataset, which consists of 400 $2048 \times 1536$ image ROIs at 0.42 mpp, annotated and extracted from H\&E FFPE diagnostic histopathology WSIs of breast carcinoma samples from the ICIAR 2018 grand challenge on breast cancer histology images. ROIs were labeled with the following 4 classes: normal (100 ROIs), benign (100 ROIs), \textit{in situ} carcinoma (100 ROIs) and invasive carcinoma (100 ROIs). For training and evaluation, we label-stratified the dataset into train-test folds (320:80 ROIs), which was used in linear probe, KNN, and SimpleShot evaluation. Additionally, we evaluate this dataset across the following center-cropped and resized image resolutions: $224 \times 224$ at 2.88 mpp, $448 \times 448$ at 1.44 mpp, $896 \times 896$ at 0.72 mpp, and $1344 \times 1344$ at 0.48 mpp.

\hheading{CCRCC Tissue Classification based on TCGA and HEL (3 classes)}\cite{brummer2022integrative, oscar_bruck_2022_7898308}: The CCRCC tissue classification task consists of 52,713 $256 \times 256$ and $300 \times 300$ ROIs at approximately 0.25 mpp, annotated and extracted from H\&E FFPE diagnostic histopathology WSIs of clear cell renal cell carcinoma (CCRCC) samples from TCGA (502 samples) and Helsinki University Hospital (HEL) (64 samples). ROIs were labeled with the following 6 classes: cancer (13,057 ROIs), normal (8,652 ROIs), stroma (5,460 ROIs), red blood cells (996 ROIs), empty background (16,026 ROIs), and other textures (8,522 ROIs). For this task, we considered only the cancer, normal, and stroma labels due to label imbalance when stratifying by data source and ambiguities of the ``other" category. We used ROIs from TCGA (21,095 ROIs) and Helsinki (6,074 ROIs) as train and test cohorts respectively (train-test fold with 21,095:6,074 ratio), which we used for linear probe, KNN, and SimpleShot evaluation. We evaluate this dataset on resized ROIs of $224 \times 224$ image resolution at approximately 0.29 mpp.

\hheading{PRAD Tissue Classification based on AGGC (5 classes)}\cite{huo2022comprehensive}: The PRAD tissue classification task is based on the Automated Gleason Grading Challenge 2022 (AGGC) from the National University Hospital and Agency of Science, Technology and Research (A$\ast$STAR) in Singapore. It comprises 221 WSIs obtained from prostatectomies (105 training, 45 testing) and biopsies (37 training, 16 testing) digitized using an Akoya Biosciences scanner at 20$\times$ magnification at 0.5 mpp. Each slide includes partial pixel-level annotations delineating different Gleason patterns and stromal regions. From the original WSIs and annotations, we built a ROI dataset consisting of 1,125,640 non-overlapping $256 \times 256$ pixel ROIs (train-test fold with 780,619:345,021 ratio), which we used for linear probe, KNN, and SimpleShot evaluation. ROIs with more than one Gleason pattern were assigned the most aggressive grade. We evaluate this dataset on resized ROIs of $224 \times 224$ image resolution at approximately 0.57 mpp.

\hheading{ESCA Tissue Classification based on UKK, WNS, TCGA, and CHA (11 classes)}\cite{tolkach2023artificial,tolkach_yuri_2023_7548828}: The ESCA tissue classification task consists of 367,229 $256 \times 256$ ROIs at 0.78 mpp, annotated and extracted from 320 H\&E FFPE diagnostic histopathology WSIs of oesophageal adenocarcinoma and adenocarcinoma of the oesophagogastric junction from four sources: University Hospital Cologne (UKK, 22 slides), Landesklinikum Wiener Neustadt (WNS, 62 slides), TCGA (22 slides), and the University Hospital Berlin—Charité (CHA, 214 slides). ROIs were labeled with the following 11 classes: adventitia (71,131 ROIs), lamina propria mucosae (2,173 ROIs), muscularis mucosae (2,951 ROIs), muscularis propria (83,358 ROIs), regression tissue (56490 ROIs), mucosa gastric (44,416 ROIs), muscosa oseophagus (18,561 ROIs), submucosa (22,117 ROIs), submucosal glands (1,516 ROIs), tumor (63,863 ROIs) and ulceration (753 ROIs). For training and evaluation, we combined UKK, WNS and TCGA into a combined training cohort (189,142 ROIs) and used CHA as a test cohort (178,187 ROIs) with a train-test fold ratio of 51:49, which we then used for linear probe, KNN, and SimpleShot evaluation. We evaluate this dataset on resized ROIs of $224 \times 224$ image resolution at approximately 0.89 mpp.

\hheading{CRC Polyp Classification based on UniToPatho (6 classes)}\cite{barbano2021unitopatho,9fsv-tm25-21}: The CRC polyp classification task is based on the UniToPatho dataset, which consists of 9,536 $1812 \times 1812$ image ROIs at 0.44 mpp, annotated and extracted from 292 H\&E FFPE diagnostic histopathology WSIs of colorectal polyp samples from University of Turin. ROIs were labeled with the following 6 classes: normal (950 ROIs), hyperplastic polyp (545 ROIs), tubular adenoma with high-grade dysplasia (454 ROIs), tubular adenoma with low-grade dysplasia (3,618 ROIs), tubulo-villous adenoma with high-grade dysplasia (916 ROIs), and tubulo-villous adenoma with low-grade dysplasia (2,186 ROIs). For training and evaluation, we used the official train-test folds (6,270:2,399 ROIs). We evaluate this dataset across the following resized image resolutions: $224 \times 224$ at 3.60 mpp, $448 \times 448$ at 1.80 mpp, $896 \times 896$ at 0.90 mpp, and $1792 \times 1792$ at 0.45 mpp.

\hheading{CRC MSI prediction based on TCGA CRC-MSI}\cite{kather2019deep,zenodo_crcmsi}: The CRC MSI prediction task is based on the ``TCGA CRC-MSI" dataset, which consists of 51,918 $512 \times 512$ ROIs at approximately 0.5 mpp, extracted from H\&E FFPE diagnostic histopathology WSIs of colorectal adenocarcinoma samples annotated and extracted from TCGA. ROIs were labeled with the following 2 classes according to the patient-level label of the sample: microsatellite instable (15,002 ROIs) and microsatellite stable (36,916 ROIs). For training and evaluation, we used the official train-test folds (19,557:32,361 ROIs), which was used in linear probe, KNN, and SimpleShot evaluation. We evaluate this dataset on resized ROIs of $448 \times 448$ image resolution at 0.57 mpp. To mitigate potential biases from site-specific H\&E staining variability in TCGA, we used Macenko normalization\cite{macenko2009method} to normalize all ROIs.

\hheading{Pan-Cancer Tissue Classification based on TCGA Uniform Tumor (32 classes)}\cite{daisuke_komura_2021_5889558,komura2022universal}: The pan-cancer tissue classification task is based on the ``TCGA Uniform Tumor" dataset, which consists of 271,170 $256 \times 256$ ROIs at around 0.5 mpp of 32 cancer types annotated and extracted from 8,736 H\&E FFPE diagnostic histopathology WSIs in TCGA. Images were labeled with the following 32 classes: 1) Adrenocortical Carcinoma (4,980 ROIs), 2) Bladder Urothelial Carcinoma (9,990 ROIs), 3) Brain Lower Grade Glioma (23,530 ROIs), 4) Breast Invasive Carcinoma (23,690 ROIs), 5) Cervical Squamous Cell Carcinoma And Endocervical Adenocarcinoma (6,270 ROIs), 6) Cholangiocarcinoma (900 ROIs), 7) Colon Adenocarcinoma (8,150 ROIs), 8) Esophageal Carcinoma (3,380 ROIs), 9) Glioblastoma Multiforme (23,740 ROIs), 10) Head And Neck Squamous Cell Carcinoma (11,790 ROIs), 11) Kidney Chromophobe (2,460 ROIs), 12) Kidney Renal Clear Cell Carcinoma (11650 ROIs), 13) Kidney Renal Papillary Cell Carcinoma (6,790 ROIs), 14) Liver Hepatocellular Carcinoma (8,370 ROIs), 15) Lung Adenocarcinoma (16,460 ROIs), 16) Lung Squamous Cell Carcinoma (16,560 ROIs), 17) Lymphoid Neoplasm Diffuse Large B-Cell Lymphoma (840 ROIs), 18) Mesothelioma (2,090 ROIs), 19) Ovarian Serous Cystadenocarcinoma (2,520 ROIs), 20) Pancreatic Adenocarcinoma (4090 ROIs), 21) Pheochromocytoma and Paraganglioma (1,350 ROIs), 22) Prostate Adenocarcinoma (9810 ROIs), 23) Rectum Adenocarcinoma (1,880 ROIs), 24) Sarcoma (13,480 ROIs), 25) Skin Cutaneous Melanoma (10,060 ROIs), 26) Stomach Adenocarcinoma (9,670 ROIs), 27) Testicular Germ Cell Tumors (6,010 ROIs), 28) Thymoma (3,600 ROIs), 29) Thyroid Carcinoma (11,360 ROIs), 30) Uterine Carcinosarcoma (2,120 ROIs), 31) Uterine Corpus Endometrial Carcinoma (12,480 ROIs), and 32) Uveal Melanoma (1,640 ROIs). For training and evaluation, we case-stratified and approximately label-stratified the dataset into train-test folds (216,350:55,360 ROIs), which was used in linear probe, KNN, and SimpleShot evaluation. We evaluate this dataset on resized ROIs of $224 \times 224$ image resolution at approximately 0.57 mpp. To mitigate potential biases from site-specific H\&E staining variability in TCGA\cite{howard2021impact}, we used Macenko normalization\cite{macenko2009method} to normalize all ROIs.

\hheading{Pan-Cancer Tumor-Immune Lymphocyte (TIL) Detection based on TCGA-TILS (2 classes)}\cite{abousamra2022deep, saltz2018spatial, kaczmarzyk_jakub_r_2022_6604094,kaczmarzyk2023champkit}: The TIL detection task is based on the ``TCGA-TILs" dataset, which consists of 304,097 100 $\times$ 100 pixels histopathology ROIs at approximately 0.5 mpp, annotated and extracted from H\&E FFPE diagnostic histopathology WSIs in TCGA. ROIs were labeled with the following 2 classes: TIL-positive (if there are at least two TILs present in the image, 54,910 ROIs) and TIL-negative (249,187 ROIs). For training and evaluation, we used the official train-validation-test folds (209,221:38,601:56,275 ROIs) and combine the train and validation folds into a single training fold. We bilinearly upsampled all images to $224 \times 224$ at 0.20 mpp for equivalent comparisons with CTransPath. To mitigate potential biases from site-specific H\&E staining variability in TCGA, we used Macenko normalization\cite{macenko2009method} to normalize all ROIs.

\hheading{Pan-Cancer Cell Type Segmentation based on SegPath (8 cell types treated as individual tasks}\cite{komura2023restaining}: The cell type segmentation tasks are derived from the SegPath dataset, which consists of 158,687 $984 \times 984$ ROIs at 0.22 mpp, annotated and extracted from H\&E FFPE diagnostic histopathology WSIs of eight major cell types in cancer tissue from University of Tokyo Hospital. Immunofluoresence and DAPI nuclear staining was performed on ROIs and used as image masks for the following classes: endothelium (10,647 ROIs), epithelium (26,509 ROIs), leukocyte (24,805 ROIs), lymphocyte (12,273 ROIs), myeloid cell (14,135 ROIs), plasma cell (13,231 ROIs), red blood cell (25,909 ROIs), and smooth muscle (31,178 ROIs). Each cell type in the dataset forms an independent tissue segmentation task with two classes, tissue/cell region and non-tissue/cell region. For training and evaluation, we used the official train-validation-test split with an approximate 80:10:10 ratio.

\heading{Computing hardware and software}\\
We used Python (version 3.8.13) and PyTorch\cite{paszke2019pytorch} (version 2.0.0, CUDA 11.7) (\href{https://pytorch.org/}{pytorch.org}) for all experiments and analyses in the study (unless specified), which can be replicated using open-source libraries as outlined below. To train \ours via DINOv2, we modify the vision transformer implementation maintained by the open-source Timm library (version 0.9.2) from Hugging Face (\href{https://huggingface.co/}{huggingface.co}) for the encoder backbone and use the original DINOv2 self-supervised learning algorithm (\href{https://github.com/facebookresearch/dinov2}{github.com/facebookresearch/dinov2}) for pretraining, which utilized 4$\times$8 80GB NVIDIA A100 GPU nodes configured for multi-GPU, multi-node training using distributed data-parallel (DDP). All other computation for downstream experiments were conducted on single 24GB NVIDIA 3090 GPUs. All WSI processing was supported by OpenSlide (version 4.3.1), openslide-python (version 1.2.0), and CLAM (\href{http://github.com/mahmoodlab/CLAM}{github.com/mahmoodlab/CLAM}). We use Scikit-learn\cite{pedregosa2011scikit} (version 1.2.1) for its implementation of K-Nearest Neighbors, and the logistic regression implementation and SimpleShot implementation provided by the LGSSL codebase (\href{https://github.com/mbanani/lgssl}{github.com/mbanani/lgssl}). Implementations of other visual pretrained encoders benchmarked in the study are found at the following links: ResNet-50 with ImageNet Transfer (\href{https://github.com/mahmoodlab/CLAM}{github.com/mahmoodlab/CLAM}), CTransPath (\href{https://github.com/Xiyue-Wang/TransPath}{github.com/Xiyue-Wang/TransPath}), and REMEDIS (\href{https://github.com/google-research/medical-ai-research-foundations}{github.com/google-research/medical-ai-research-foundations}). We note that REMEDIS requires fulfillment of a data use agreement, which can be accessed and submitted at the PhysioNet website (\href{https://physionet.org/content/medical-ai-research-foundation/1.0.0/}{physionet.org/content/medical-ai-research-foundation})\cite{goldberger2000physiobank,medai_doi}. For multi-head attention visualization, we used the visualisation tools provided by the HIPT codebase (\href{https://github.com/mahmoodlab/HIPT}{github.com/mahmoodlab/HIPT}). For training weakly-supervised ABMIL models, we adapted the training scaffold code from the CLAM codebase (\href{https://github.com/mahmoodlab/CLAM}{github.com/ mahmoodlab/CLAM}). For training semantic segmentation, we use the original Mask2Former implementation (\href{https://github.com/facebookresearch/Mask2Former}{github.com/facebookresearch/Mask2Former}) which is based on detectron2\cite{wu2019detectron2} (version 0.6), and required the following older packages for compatibility: Python (version 3.8) and PyTorch (version 1.9.0, CUDA 11.1). For adding ViT-Adapter to \ours, we adapt its original implementation (\href{https://github.com/czczup/ViT-Adapter}{github.com/czczup/ViT-Adapter}) in detectron2 to train it using Mask2Former. Pillow (version 9.3.0) and OpenCV-python were used to perform basic image processing tasks. Matplotlib (version 3.7.1) and Seaborn (version 0.12.2) were used to create plots and figures. Usage of other miscellaneous Python libraries is listed in the \textbf{Reporting Summary}.

\heading{Data availability}\\
TCGA and CPTAC data consisting of whole slide images and labels can be accessed through the NIH genomic data commons\footnote{\href{https://portal.gdc.cancer.gov}{portal.gdc.cancer.gov}} and proteomics data commons\footnote{\href{https://proteomic.datacommons.cancer.gov/pdc}{proteomic.datacommons.cancer.gov}} respectively. GTEx data added to the pretraining dataset can be accessed through the GTEx portal\footnote{\href{https://www.gtexportal.org/home/}{gtexportal.org}}. CPTAC data consisting of  All publicly-available datasets analyzed in this work can be can accessed in their respective data portals: CRC-100K\footnote{\href{https://zenodo.org/record/1214456}{zenodo.org/record/1214456}}, HunCRC patches\footnote{\href{https://doi.org/10.6084/m9.figshare.c.5927795.v1 }{doi.org/10.6084/m9  .figshare.c.5927795.v1}}, HunCRC slides\footnote{\href{https://doi.org/10.7937/tcia.9cjf-0127   }{doi.org/10.7937/tcia.9cjf-0127}}, BACH\footnote{\href{https://iciar2018-challenge.grand-challenge.org/Dataset/}{iciar2018-challenge.grand-challenge.org/Dataset/}}, TCGA CRC-MSI \footnote{\href{https://zenodo.org/record/3832231}{zenodo.org/record/3832231}}. CCRCC Tissue Classification\footnote{\href{https://zenodo.org/record/7898308}{zenodo.org/record/7898308}}. TCGA-TILs \footnote{\href{https://zenodo.org/record/6604094}{zenodo.org/record/6604094}}, TCGA Uniform\footnote{\href{https://zenodo.org/record/5889558}{zenodo.org/record/5889558}}, UniToPatho \footnote{\href{https://zenodo.org/record/4643645}{zenodo.org/record/4643645}}, ESCA\footnote{\href{https://zenodo.org/record/7548828}{zenodo.org/record/7548828}}, EBRAINS\footnote{\href{https://doi.org/10.25493/WQ48-ZGX  }{doi.org/10.25493/WQ48-ZGX}}, DHMC Kidney\footnote{\href{https://bmirds.github.io/KidneyCancer/}{bmirds.github.io/KidneyCancer/}}, BRACS\footnote{\href{https://www.bracs.icar.cnr.it/}{bracs.icar.cnr.it/}}, PANDA\footnote{\href{https://panda.grand-challenge.org/data/}{panda.grand-challenge.org/data/}}, SegPath\footnote{\href{https://zenodo.org/record/7412731}{zenodo.org/record/7412731}}, and AGGC\footnote{\href{https://zenodo.org/record/6460100}{zenodo.org/record/6460100}}. TCGA, CPTAC, HunCRC, TCGA-TILS can also be accessed using the The Cancer Imaging Archive\cite{clark2013cancer}. We note that data from AGGC was obtained from a public grand challenge (of the same name \footnote{\href{https://aggc22.grand-challenge.org}{https://aggc22.grand-challenge.org}}) with a pending publication\cite{huo2022comprehensive}, with permission granted to present results on this dataset from the challenge organizers. Works that use AGGC should also obtain permission from the challenge organizers, and acknowledge the pending publication of the grand challenge. Image patches used for pretraining and BWH-EMB WSIs were curated with in-house data. Following institution policies, all requests for data collected or curated in-house will be evaluated accordingly on a case-by-case basis to determine whether the data requested and the use case is compliant with intellectual property or patient privacy obligations.

\heading{Code availability}\\
The authors will release code for performing downstream evaluation using the pretrained encoder of this work upon publication. Since our model utilizes proprietary patient data for pretraining, the model weights may be requested and granted institutional permission for access on a case-by-case basis. All methods and software packages used in the study have been documented and explained in ways accessible to the broader scientific audience.

\heading{Author contributions}\\
R.J.C., F.M., T.D., M.Y.L., D.F.K.W. conceived the study and designed the experiments. R.J.C., L.P.L., D.F.K.W., J.J.W., T.D., M.Y.L, G.J., A.V., A.H.S., B.C., A.Z., D.S., M.S., L.O., S.S. collected the data for self-supervised learning. R.J.C., T.D., M.Y.L. performed model development for self-supervised learning. R.J.C., M.Y.L, T.D., B.C., G.J. organized the datasets and codebases for all downstream tasks regarding ROI classification, ROI segmentation, and slide classification. R.J.C., M.Y.L., G.J. performed experimental analysis regarding ROI classification. T.D., M.Y.L., R.J.C., L.W., W.W. performed experimental analysis regarding ROI segmentation. R.J.C., T.D., B.C., D.S., M.S., M.W., A.Z., L.W. performed experimental analysis regarding slide classification. R.J.C., T.D., M.Y.L., D.F.K.W., G.J., A.H.S., L.P.L., G.G., F.M. interpreted the experimental results and provided feedback on the study. R.J.C., T.D., M.Y.L, D.F.K.W., F.M. prepared the manuscript with input from all co-authors. F.M. supervised the research.

\heading{Acknowledgements}\\
This work was supported in part by the BWH \& MGH Pathology, BWH president’s fund, Massachusetts Life Sciences Center, NIGMS R35GM138216 (F.M.), and the BWH President's Scholar fund (G.G.). M.Y.L. was also supported by the Siebel Scholars program. R.J.C., D.S., and S.S. were also supported by the NSF Graduate Fellowship. L.O. was also supported by the German Academic Exchange (DAAD) Fellowship. We thank Timothy Janicki at BWH, and Richard Kenny and the system administration staff at the MGB Enterprise Research Infrastructure \& Services (ERIS) Research Computing Core, for their dedicated support in providing and maintaining access to NVIDIA A100 computing resources. We thank the Computational Digital Pathology Lab and Bioinformatics Institute at the Agency of Science, Technology and Research (A$\ast$STAR) and the AGGC 2022 challenge committee for making the AGGC dataset available for research prior to the official challenge paper being published. We thank Negin Vatanian, Mathangi Thiagarajan, Brenda Fevrier-Sullivan and Justin Kirby at the National Institutes of Health (NIH) for navigating access to whole-slide imaging data in CPTAC.
\end{spacing}

\setcounter{figure}{0}
\renewcommand{\figurename}{Extended Data Figure}

\begin{figure*}
\centering
\includegraphics[width=\textwidth]{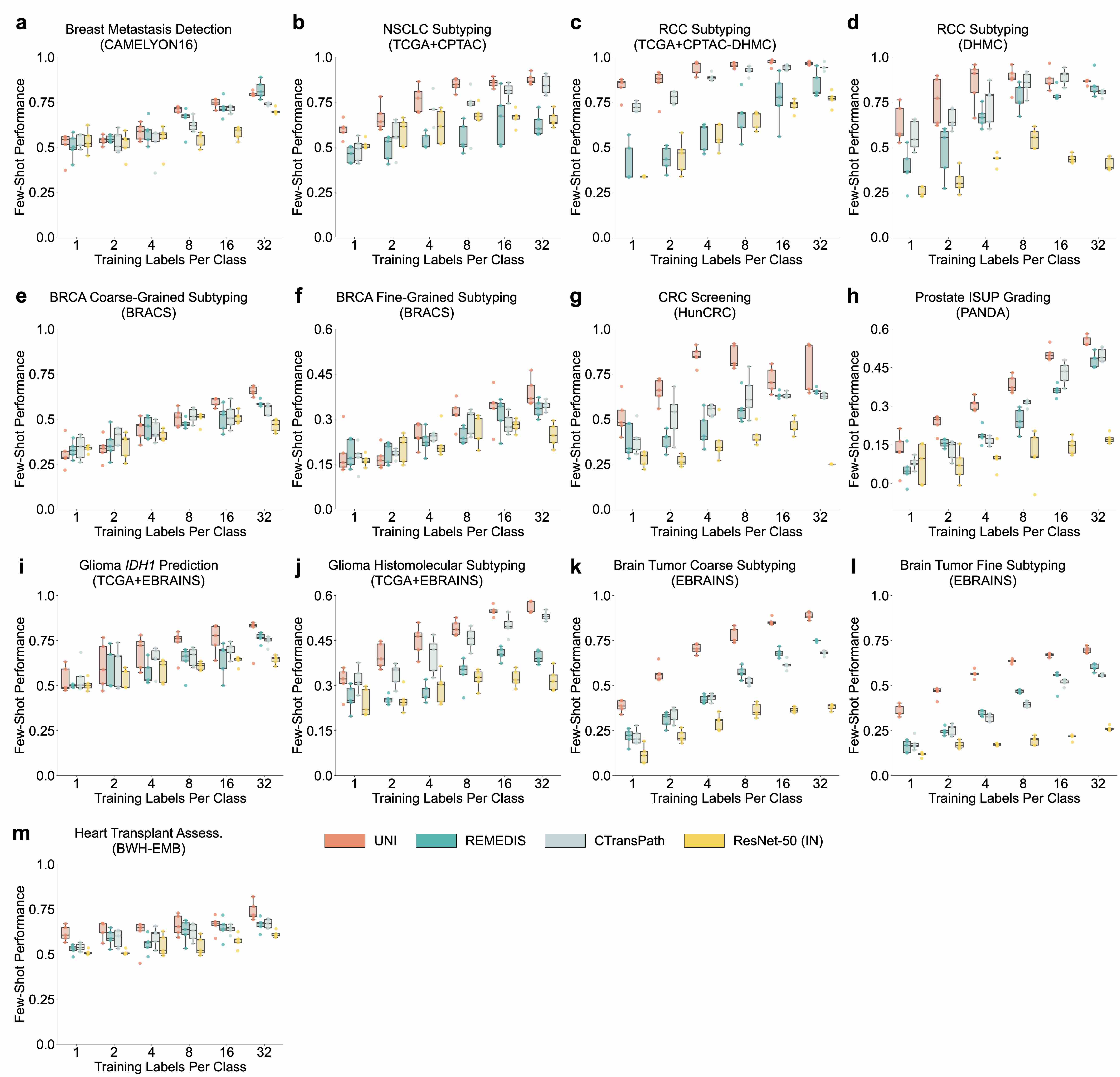}
\caption{\textbf{Few-Shot slide classification.} To study the label efficiency of UNI in slide classification, we compare \ours with other pretrained encoders on: \textbf{a.} breast metastasis detection in CAMELYON16, \textbf{b.} NSCLC subtyping in CPTAC (trained on TCGA) \textbf{c.} RCC subtyping in CPTAC-DHMC (trained on TCGA), \textbf{d.} RCC subtyping in DHMC, \textbf{e.} BRCA coarse-grained subtyping in BRACS, \textbf{f.} BRCA fine-grained subtyping in BRACS, \textbf{g.} CRC screening in HunCRC, \textbf{h.} Prostate ISUP Grading in PANDA, \textbf{i.} glioma IDH1 prediction in EBRAINS (trained on TCGA), \textbf{j.} glioma histomolecular subtyping in EBRAINS (trained on TCGA), \textbf{k.} brain tumor coarse-grained subtyping in EBRAINS, \textbf{l.} brain tumor fine-grained subtyping in EBRAINS, and \textbf{m.} heart transplant assessment in BWH-EMB. The performance is measured across different few-shot settings with $K\in \{1,2,4,8,16,32\}$ training examples used per class. Boxes indicate quartile values of model performance ($n=5$ runs) and whiskers extend to data points within 1.5$\times$ the interquartile range. Overall, we observe that UNI consistently demonstrates superior label efficiency over other baselines.}
\label{fig:slide-level-fs}
\end{figure*}

\begin{figure*}
\centering
\includegraphics[width=0.95\textwidth]{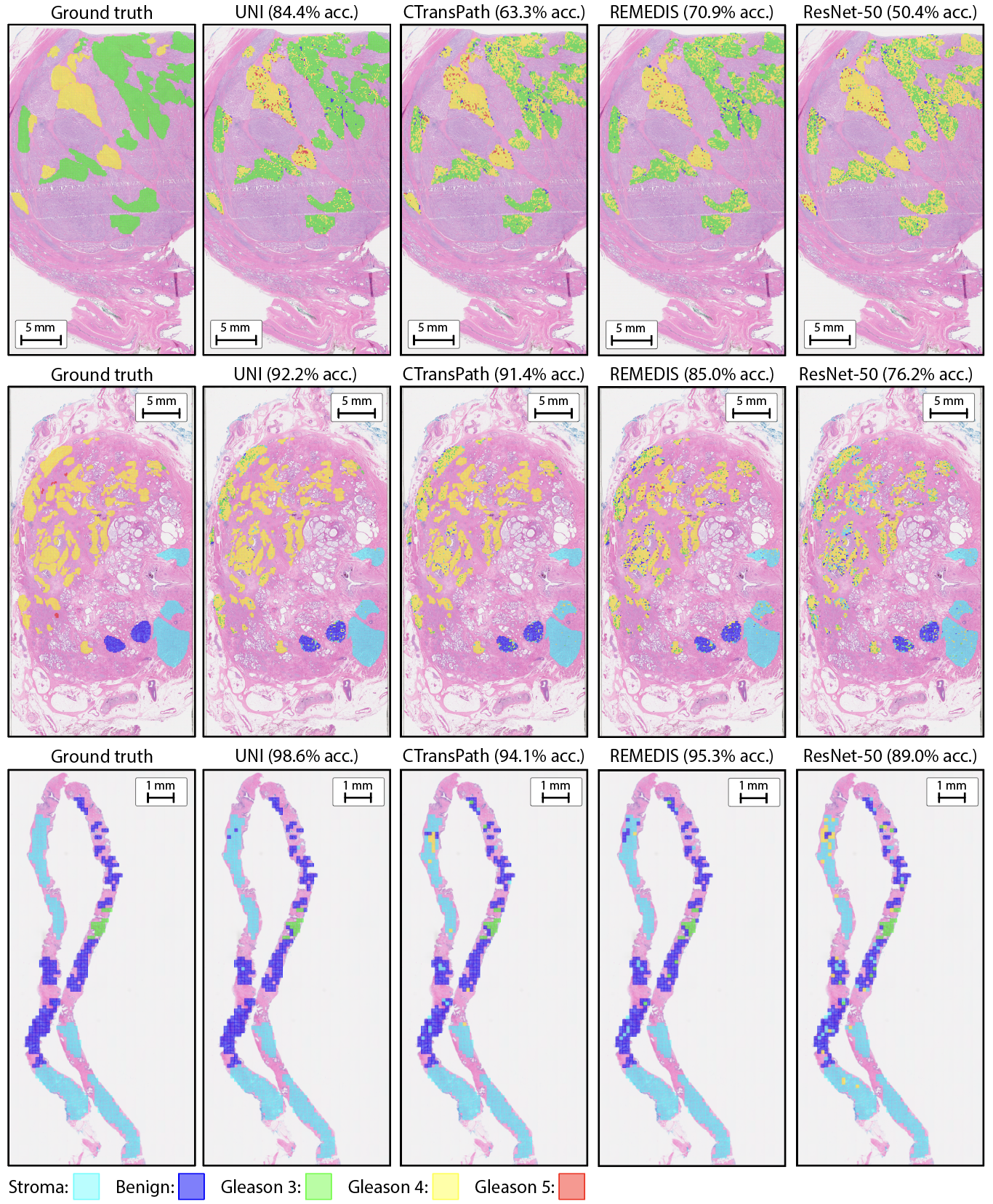}
\caption{\textbf{Comparing supervised performance on PRAD tissue classification in AGCC.} Qualitative illustrations comparing \ours to CTransPath, REMEDIS, and ResNet-50 (IN) via KNN probing on PRAD tissue classification in AGCC. \ours achieves better accuracy (acc.) on all three examples. The reported results are based on partial annotations (left-most panel) provided by pathologists.}
\label{fig:aggc-compare}
\end{figure*}

\begin{figure*}
\centering
\includegraphics[width=\textwidth]{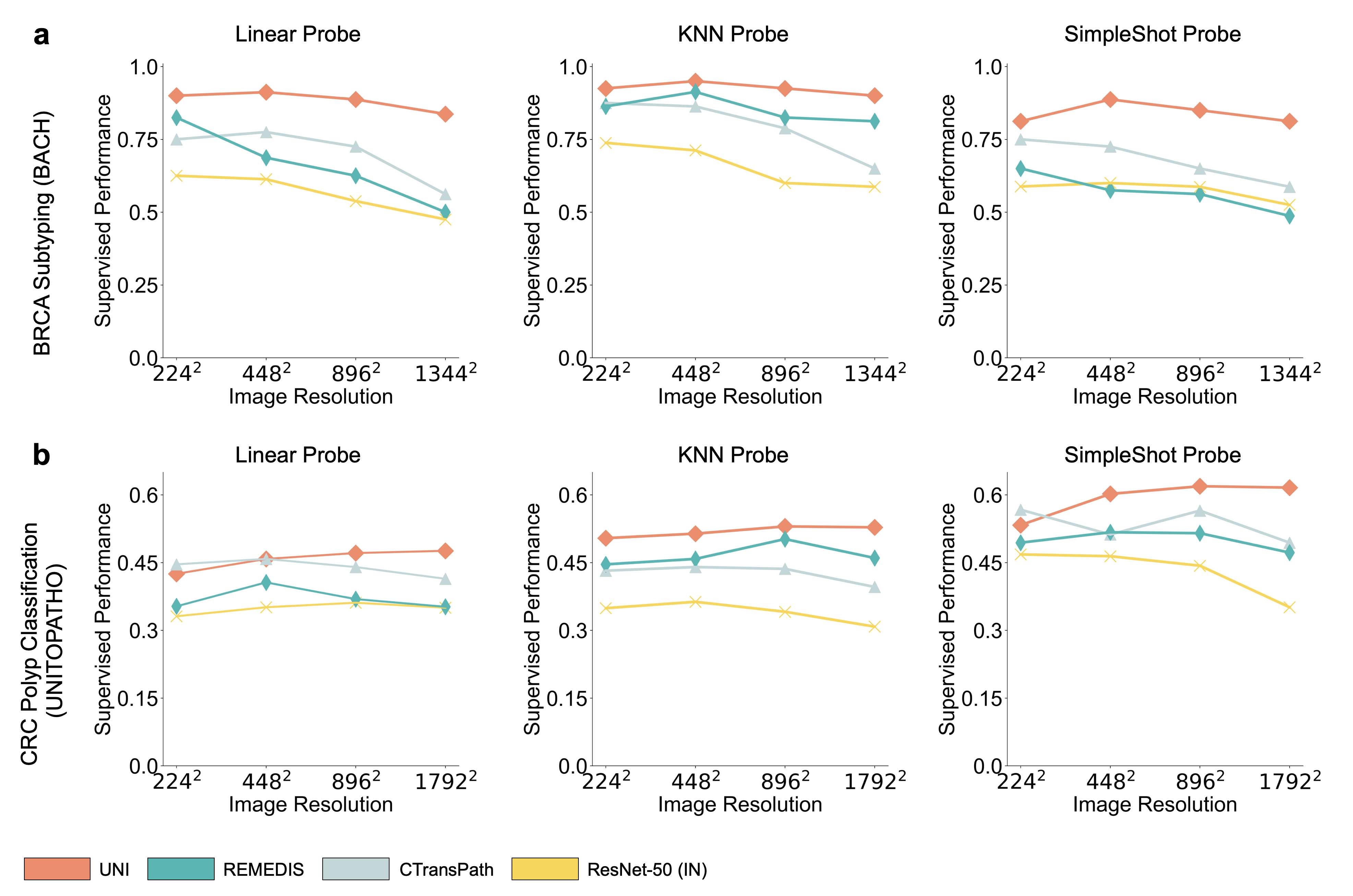}
\caption{\textbf{ROI classification across different image resolutions.} To assess how image resolution affects performance, we compare \ours and other baselines on various resized and center-cropped ROIs for \textbf{a.} BRCA subtyping and \textbf{b.} CRC polyp classification tasks. The original image size is $2048\times 1536$ and $1812 \times 1812$ pixels, respectively. All models are evaluated on linear, SimpleShot (1-NN), and KNN (20-NN) probe settings. \ours consistently outperforms all baselines across all resolutions. The performance metrics are further provided in \textbf{Extended Data Table \ref{tab:patch-bach-lin}, \ref{tab:patch-bach-knn}, \ref{tab:patch-unitopatho-lin}, \ref{tab:patch-unitopatho-knn}}.}
\label{fig:patch-level-res}
\end{figure*}

\begin{figure*}
\centering
\includegraphics[width=\textwidth]{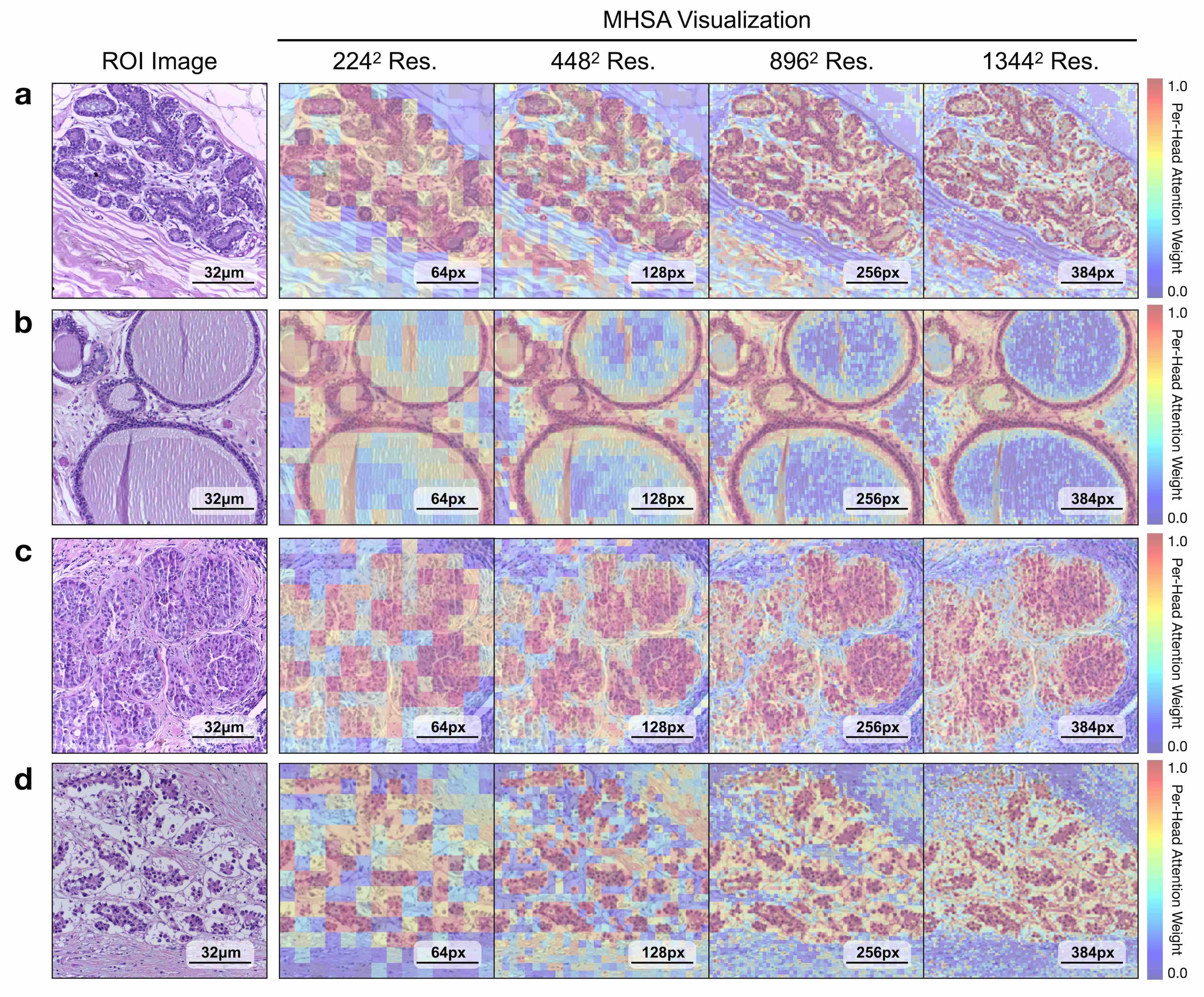}
\caption{\textbf{Multi-head self-attention (MHSA) heatmap visualization of \ours across different image resolutions in BRCA Subtyping in BACH.} Each colored square represents a $16 \times 16$ patch token encoded by \ours, with heatmap color corresponding to the attention weight of that patch token to the global $\textsc{[CLS]}$ token of the penultimate layer in \ours. We show MHSA visualizations for resized and center-cropped ROIs at $224^2, 448^2, 896^2, 1344^2$ resolutions for the normal (\textbf{a}), benign (\textbf{b}), in situ (\textbf{c}), and invasive (\textbf{d}) classes in BACH. In each, the leftmost image is the original H\&E ROI and the right four images are the MHSA visualizations. For comparative purposes, we resize all images within the figure to have the same dimension, but note that at higher resolutions, each colored square has an original image resolution of $16 \times 16$ pixels at 0.42 mpp. As the resolution increases, the heatmaps demonstrate increasing and increasingly fine-grained attention focused on epithelial structures, with relatively lower attention on stroma or other background, neither of which are contributory to the diagnoses in these ROIs.
}
\label{fig:bach-res}
\end{figure*}

\begin{figure*}
\centering
\includegraphics[width=\textwidth]{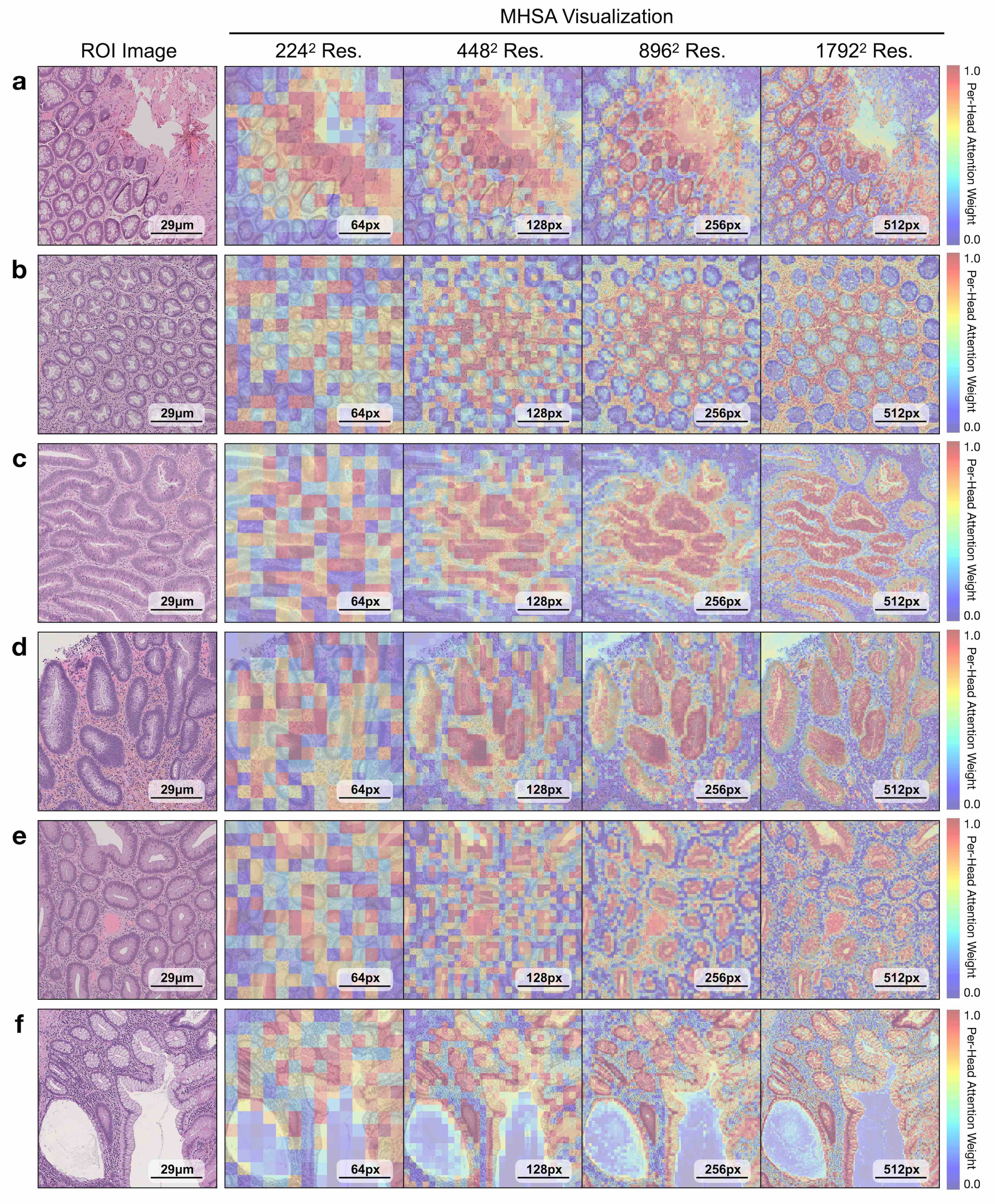}
\label{fig:crc1-res}
\end{figure*}

\begin{figure*}
\centering
\caption{\textbf{Multi-head self-attention (MHSA) heatmap visualization of \ours across different image resolutions for CRC polyp classification in UniToPatho.} Each colored square represents a $16 \times 16$ patch token encoded by \ours, with heatmap color corresponding to the attention weight of that patch token to the global $\textsc{[CLS]}$ token of the penultimate layer in \ours. We show MHSA visualizations for resized and center-cropped ROIs at $224^2, 448^2, 896^2, 1792^2$ resolutions for normal tissue (\textbf{a}), a hyperplastic polyp (\textbf{b}), a tubular adenoma with low-grade dyplasia (\textbf{c}), a tubular adenoma with high-grade dyplasia (\textbf{d}), a tubuluo-villous adenoma with high-grade dyplasia (\textbf{e}), and a tubuluo-villous adenoma with low-grade dyplasia (\textbf{f}). In each, the leftmost image is the original H\&E ROI and the right four images are the MHSA visualizations. For comparative purposes, we resize all images within the figure to have the same dimension, but note that at higher resolutions, each colored square has an original image resolution of $16 \times 16$ pixels at 0.48 mpp. As resolution increases, the heatmaps demonstrate increasing and increasingly fine-grained attention focused on the crypts, in all cases except the hyperplastic polyp in \textbf{b}, focusing on areas a pathologist would use to make the diagnosis. The high attention on the stroma in the case of the hyperplastic polyp in \textbf{b} is interesting and may be evidence that hyperplastic polyps can have a different character of stroma than other colorectal polyps.
}
\label{fig:crc2-res}
\end{figure*}

\begin{figure*}
\centering
\includegraphics[width=\textwidth]{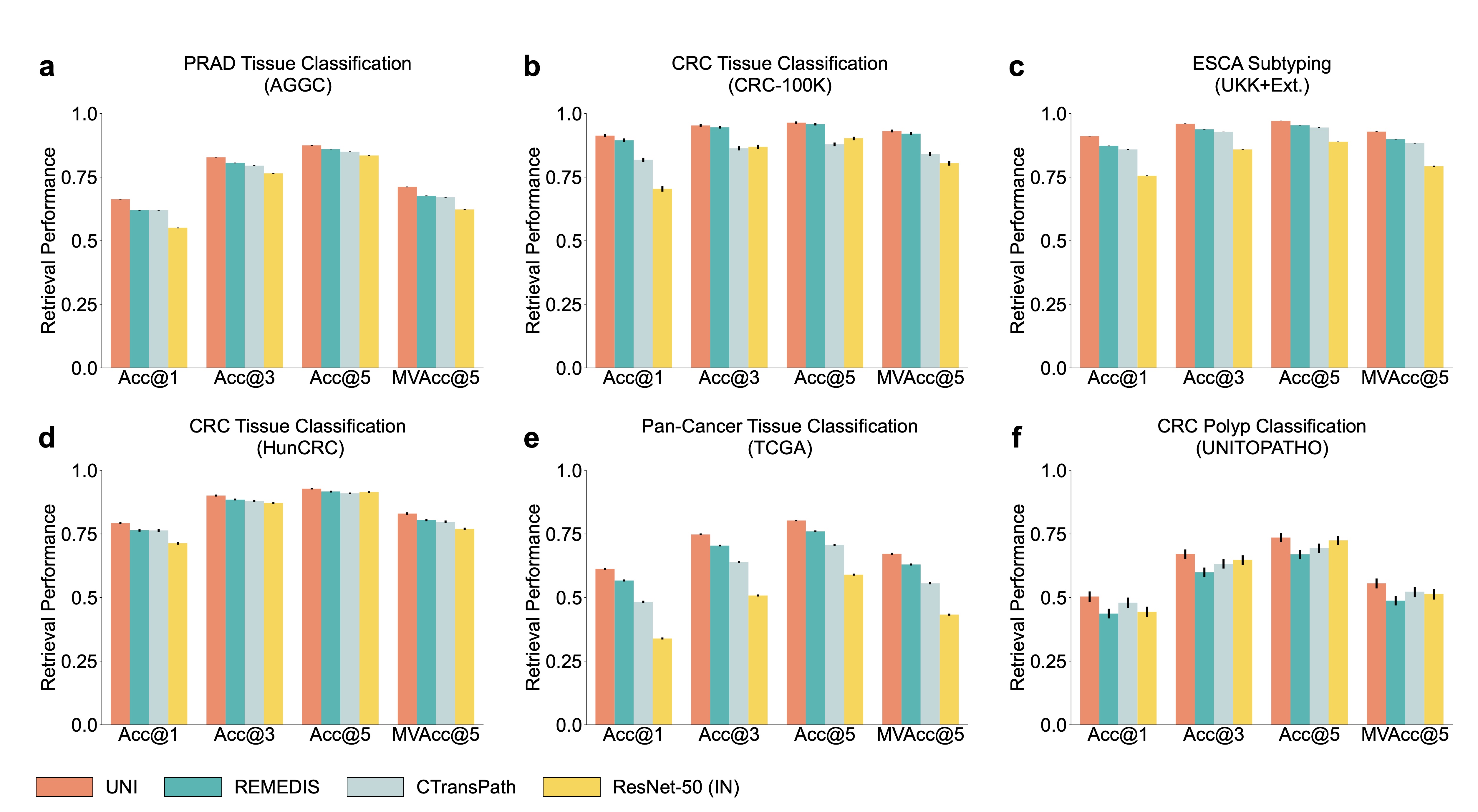}
\caption{\textbf{ROI retrieval.} We evaluate content-based image retrieval for ROI-level classes with at least 5 classes, for \textbf{a.} PRAD tissue classification in AGGC, \textbf{b.} CRC tissue classification in CRC-100K, \textbf{c.} ESCA subtyping on CHA (trained on UKK, WNS and TCGA), \textbf{d.} CRC tissue classification in HunCRC, \textbf{e.} pan-cancer tissue classification in TCGA, and CRC polyp classification in UniToPatho. \ours consistently outperforming all pretrained encoders. Detailed performance metrics are further provided in \textbf{Extended Data Table \ref{tab:aggc-retrieval}-\ref{tab:unitopatho-retrieval}}.}
\label{fig:patch-level-retrieval}
\end{figure*}

\begin{figure*}
\centering
\includegraphics[width=\textwidth]{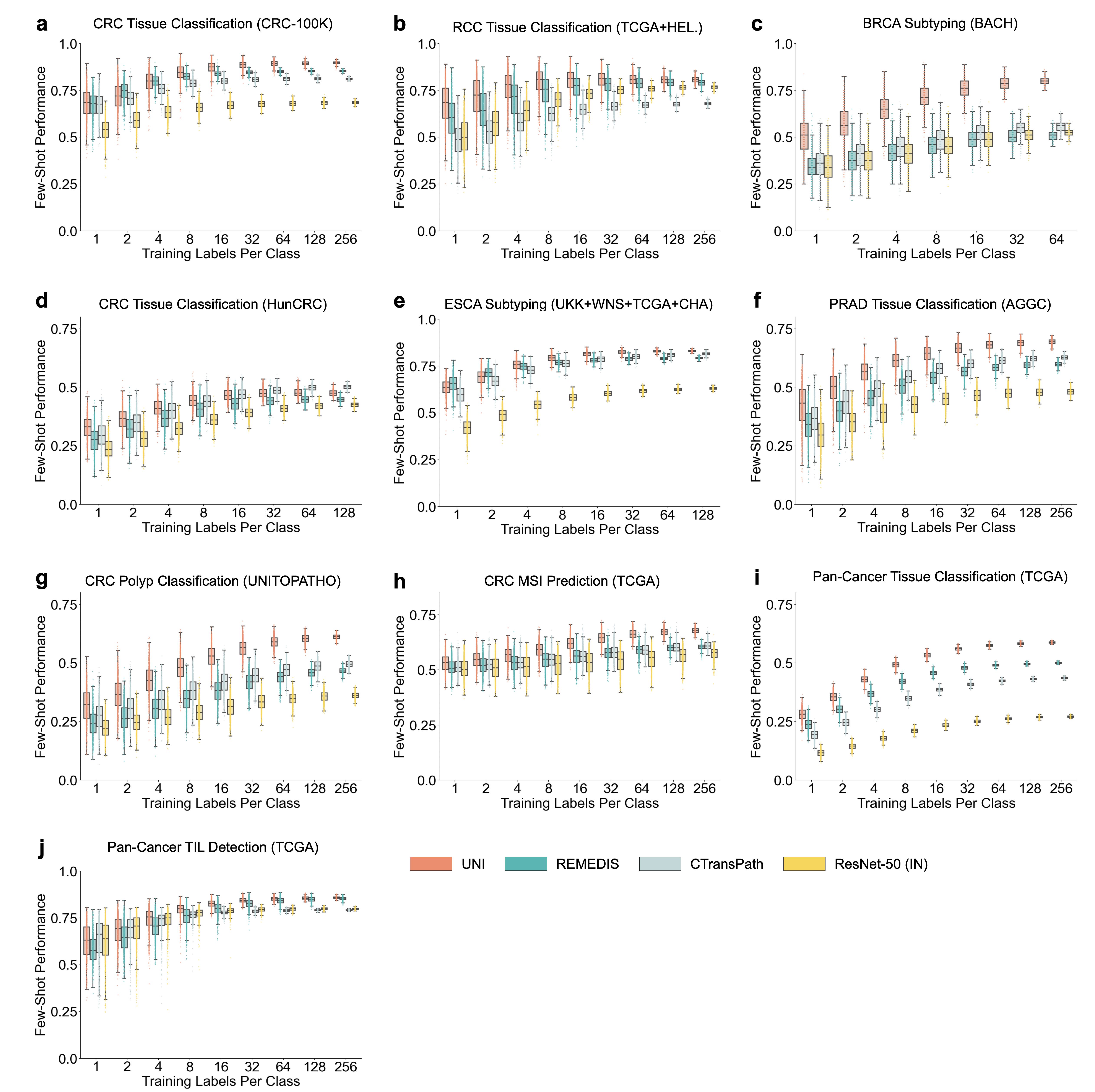}
\caption{\textbf{Few-shot ROI classification using class prototypes.} Similar to slide-level classification, we also assess the label-efficiency of \ours on ROI-level tasks, and observe superior label efficiency of \ours on most tasks except CRC tissue classification on HunCRC. We evaluate all pretrained encoders using the non-parametric SimpleShot framework for \textbf{a.} CRC tissue classification in CRC-100K, \textbf{b.} RCC tissue classification on HEL (trained on TCGA), \textbf{c.} BRCA subtyping in BACH, \textbf{d.} CRC tissue classification in HunCRC, \textbf{e.} ESCA subtyping on CHA (UKK+WNS+TCGA), \textbf{f.} PRAD tissue classification in AGGC, \textbf{g.} CRC polyp classification in UniToPatho, \textbf{h.} CRC MSI prediction in TCGA, \textbf{i.} pan-cancer tissue classification in TCGA, and \textbf{j.} pan-cancer TIL detection in TCGA.}
\label{fig:patch-level-fs}
\end{figure*}

\begin{figure*}
\centering
\includegraphics[width=\textwidth]{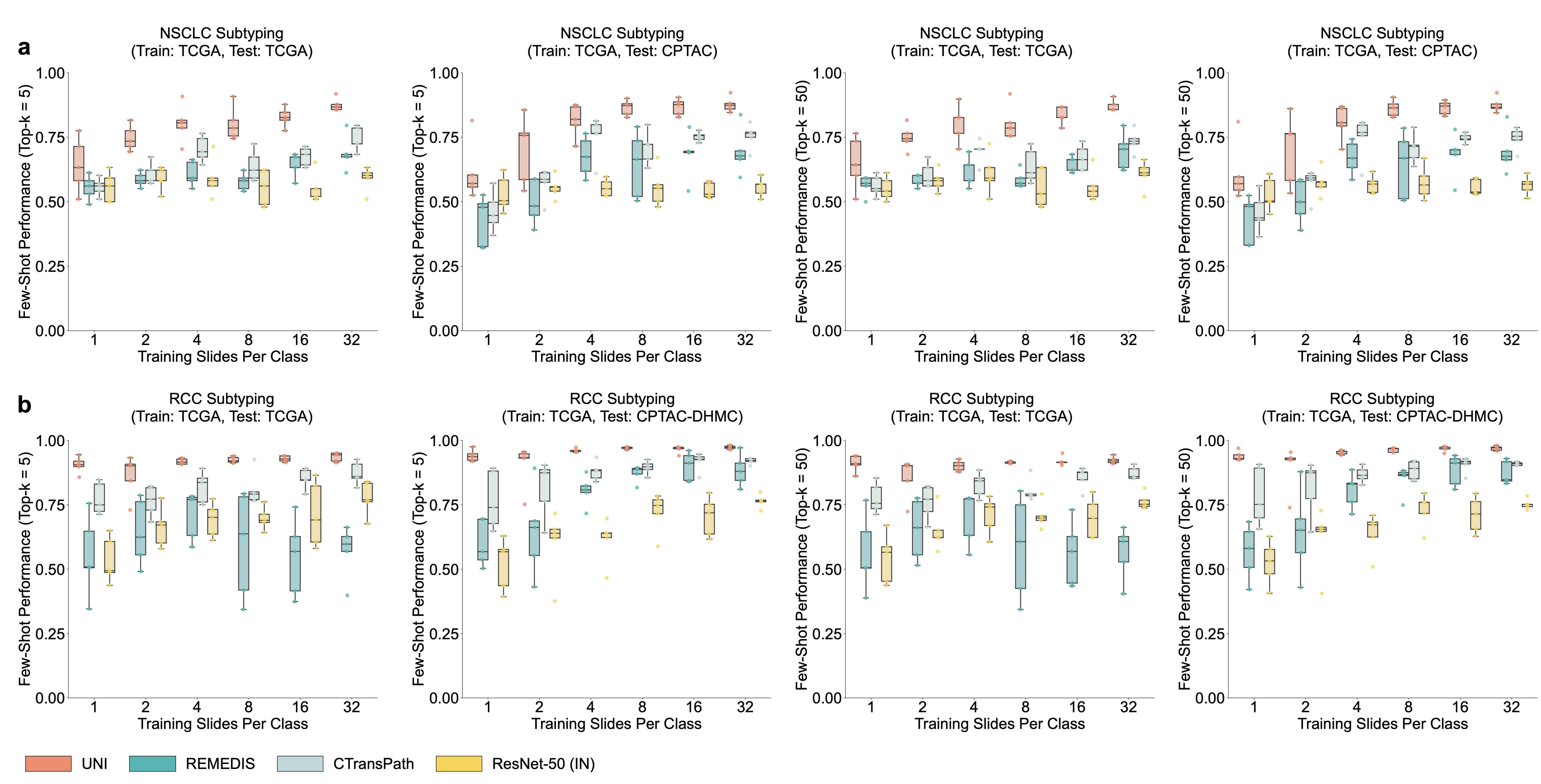}
\caption{\textbf{Few-shot slide classification using class prototypes.} We adapt the SimpleShot framework for slide-level classification, called ``MI-SimpleShot". ROI class prototypes are constructed by averaging the pre-extracted ROI features for each class using the ``TCGA Uniform Tumor" dataset, which we use as ``prompts" for assigning the slide-level label. We assess and compare the few-shot performance of all pretrained encoders on NSCLC subtyping (\textbf{a}) and RCC subtyping task (\textbf{b}), using the same runs ($n$=5) in the few-shot setting for ABMIL for $K\in \{1,2,4,8,16,32\}$ training examples used per class. We compare performance of top-5 and top-50 pooling of nearest patches in the test set, as well as show performance on both the internal test fold in TCGA and external cohort. Boxes indicate quartile values of model performance ($n=5$ runs) and whiskers extend to data points within 1.5$\times$ the interquartile range. We also compare. Overall, we observe that the formed prototypes by \ours can be used to classify slides based on the MI-SimpleShot framework. \textbf{a.} On NSCLC subtyping, we observe that 2-shot and 4-shot performance from \ours outperforms the 32-shot performance of all other models. \textbf{b.} On RCC subtyping, 1-shot performance of \ours also outperforms the 32-shot performance of other models. We also observe that MI-SimpleShot can be combined with other pretrained encoders as well, but generally require more annotated ROIs for creating prototypes.}
\label{fig:slide-level-proto-fs}
\end{figure*}

\begin{figure*}
\centering
\includegraphics[width=\textwidth]{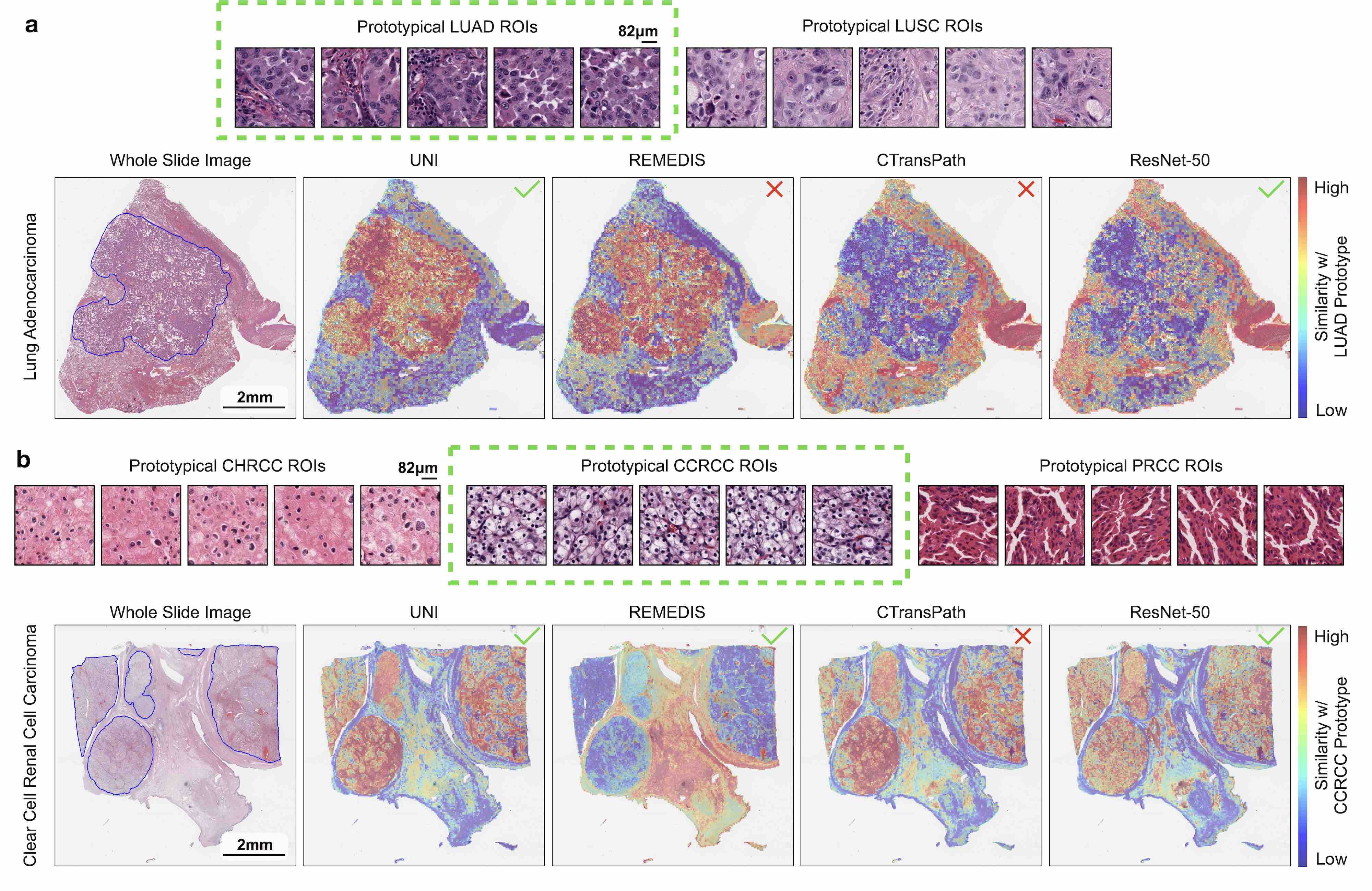}
\caption{\textbf{Comparing 1-shot similarity heatmaps of pretrained encoders with class prototype.} We compare the similarity heatmaps of all pretrained encoders using annotated ROIs from a single slide per class for forming class prototypes in MI-SimpleShot (with top-5 pooling) on NSCLC subtyping (\textbf{a}) and RCC subtyping task (\textbf{b}), with \textbf{top} visualizing example ROIs used for each class, and \textbf{bottom} showing similarity heatmaps. Outlined in blue are pathologist annotations of ROIs that match the label of the histology slide. Similarity heatmaps are created with respect to the class prototype of the correct slide label (indicated in green), with a \checkmark indicating a correct prediction and \xmark\enspace indicating incorrect prediction. Note that since the visualizations are created with respect to the ground truth label, the model may retrieve correct patches that match pathologist annotations but still misclassify the slide. \textbf{a.} On a LUAD slide, we observe strong agreement of the pathologist's annotations with retrieved LUAD patches by \ours. Though retrieved patches by REMEDIS also have strong agreement with the pathologist's annotations, we note that slide was misclassified as LUSC, indicating that the top-5 retrieved patches of the LUSC prototype was higher than that of the LUAD prototype. Vice versa, ResNet-50$_{\text{IN}}$ classifies the slide correctly but incorrectly retrieves the correct patches that agree with the pathologist's annotations, indicating that non-LUAD patches in the slide were more LUAD-like than the pathologist-annotated LUAD patches with respect to the LUAD prototype. The similarity heatmap for CTransPath both misclassified the slide and retried incorrect patches. \textbf{b.} On a example CCRCC slide, we observe strong agreement of the pathologist's annotations with retrieved CCRCC patches by \ours. We observe similar mismatch in predicted class label and retrieved patches, in which REMEDIS classifies the slide correctly but retrieves the incorrect patches, and CTransPath misclassifies the slide but retrieves the correct patches.}
\label{fig:slide-level-proto-heatmaps}
\end{figure*}
\clearpage

\setcounter{table}{0}
\renewcommand{\tablename}{Extended Data Table}

\begin{table}
\centering
\begin{tabular}{lr}
\toprule
Major Tissue Type &  Num. Slides \\
\midrule
Heart                &             10427 \\
Lung                 &              9846 \\
Kidney               &              8388 \\
Bowel / Lower GI                &              8303 \\
Soft Tissue          &              7863 \\
Brain                &              7412 \\
Esophagogastric      &              6705 \\
Endocrine            &              6138 \\
Female Genital Tract &              5796 \\
Lymphatic System     &              4957 \\
Liver Biliary Tract  &              4677 \\
Male Genital Tract   &              4017 \\
Skin                 &              3653 \\
Breast               &              3364 \\
Bone                 &              2667 \\
Pancreas             &              2328 \\
Head \& Neck            &              1555 \\
Peritoneum           &              1210 \\
Bladder              &              1059 \\
Eye                  &                61 \\
\midrule
Total & 100426 \\
\bottomrule
\end{tabular}
\caption{\textbf{Major Tissue Type Distribution of Mass-100K.} Mass-100K is a pretraining dataset that consists of 100,130,900 tissue patches from 100,426 diagnostic whole-slide images (WSIs) across 20 major tissue types collected from the Massachusetts General Hospital (MGH), Brigham \& Women's Hospital (BWH) and the Genotype-Tissue Expression (GTEx) consortium.}
\label{tab:op100k}
\end{table}

\begin{table}
\centering
\begin{tabular}{lr}
\toprule
Major Tissue Type &  Num. Slides \\
\midrule
Heart                &             4 \\
Lung                 &              3044 \\
Kidney               &              435 \\
Bowel / Lower GI                &              2002 \\
Soft Tissue          &              1275 \\
Brain                &              1607 \\
Esophagogastric      &              882 \\
Endocrine            &              549 \\
Female Genital Tract &              1834 \\
Lymphatic System     &              1342 \\
Liver Biliary Tract  &              1403 \\
Male Genital Tract   &              587 \\
Skin                 &              627 \\
Breast               &              1366 \\
Bone                 &              301 \\
Pancreas             &              543 \\
Head \& Neck            &              532 \\
Peritoneum           &              389 \\
Bladder              &              568 \\
Eye                  &                21 \\
\midrule
Total & 21444 \\
\bottomrule
\end{tabular}
\caption{\textbf{Major Tissue Type Distribution of Mass-22K.} Mass-22K is a subset of Mass-100K, which contains 16,059,454 histology image patches sampled from 21,444 WSIs across 20 major tissue types with cancerous tissue collected from Massachusetts General Hospital (MGH) and Brigham \& Women's Hospital (BWH).}
\label{tab:op22k}
\end{table}

\begin{table}
\centering
\begin{tabular}{lr}
\toprule
Major Tissue Type &  Num. Slides \\
\midrule
Heart                &             3 \\
Lung                 &              136 \\
Kidney               &              43 \\
Bowel / Lower GI                &              55 \\
Soft Tissue          &              176 \\
Brain                &              219 \\
Esophagogastric      &              32 \\
Endocrine            &              34 \\
Female Genital Tract &              187 \\
Lymphatic System     &              134 \\
Liver Biliary Tract  &              75 \\
Male Genital Tract   &              33 \\
Skin                 &              43 \\
Breast               &              42 \\
Bone                 &              32 \\
Pancreas             &              28 \\
Head \& Neck            &              61 \\
Peritoneum           &              34 \\
Bladder              &              30 \\
Eye                  &                7 \\
\midrule
Total & 1404 \\
\bottomrule
\end{tabular}
\caption{\textbf{Major Tissue Type Distribution of Mass-1K.} Mass-1K is a subset of Mass-22K, which contains 1,064,615 histology image patches sampled from 1,404 WSIs across 20 major tissue types with cancerous tissue collected from Massachusetts General Hospital (MGH) and Brigham \& Women's Hospital (BWH).}
\label{tab:op1k}
\end{table}

\begin{table}
\centering
\begin{tabular}{p{5cm}l|l}
\toprule
Cancer Type & OncoTree Code & Num. Slides \\
\midrule\midrule
Adrenocortical Carcinoma & ACC &  51 (31:20) \\
\midrule
Ampullary Cancer & AMPCA &  42 (16:26) \\
\midrule
Anal Cancer & ANSC &  65 (51:14) \\
\midrule
Appendiceal Cancer & MAAP &  40 (23:17) \\
\midrule
\multirow{3}{*}{Bladder Cancer} & BLAD &   28 (8:20) \\
            & BLCA &  65 (20:45) \\
            & UTUC &  65 (22:43) \\
\midrule
\multirow{3}{*}{Bone Cancer} & CHS &  32 (10:22) \\
            & ES &  48 (17:31) \\
            & OS &  54 (26:28) \\
\midrule
\multirow{2}{*}{Breast Cancer} & IDC &   65 (56:9) \\
            & ILC &   65 (65:0) \\
            & MDLC &   65 (62:3) \\
\midrule
\multirow{2}{*}{CNS Cancer} & ATM &   20 (7:13) \\
            & MNG &  65 (47:18) \\
\midrule
\multirow{2}{*}{Cervical Cancer} & CESC &   60 (60:0) \\
            & ECAD &   42 (42:0) \\
\midrule
\multirow{3}{*}{Colorectal Cancer} & COAD &  65 (35:30) \\
            & MACR &  35 (14:21) \\
            & READ &  65 (31:34) \\
\midrule
Embryonal Tumor & MBL &  62 (23:39) \\
\midrule
\multirow{5}{*}{Endometrial Cancer} & UCCC &   23 (23:0) \\
            & UCS &   65 (65:0) \\
            & UEC &   65 (65:0) \\
            & UMEC &   25 (25:0) \\
            & USC &   65 (65:0) \\
\midrule
\multirow{5}{*}{Esophagogastric Cancer} & ESCA &  64 (19:45) \\
            & ESCC &  65 (31:34) \\
            & GEJ &  65 (17:48) \\
            & SSRCC &  28 (15:13) \\
            & STAD &  65 (31:34) \\
\midrule
Gastroint. Neuroend. Tumor & GINET &  65 (33:32) \\
\midrule
Gastroint. Stromal Tumor & GIST &  65 (33:32) \\
\midrule
\multirow{2}{*}{Germ Cell Tumor} & MGCT &   36 (0:36) \\
            & SEM &   63 (0:63) \\
\end{tabular}
\end{table}

\begin{table}
\centering
\begin{tabular}{p{5cm}l|l}
\toprule
Cancer Type (Cont.) & OncoTree Code & Num. Slides \\
\midrule
\multirow{9}{*}{Glioma} & AASTR &  54 (20:34) \\
            & AODG &  31 (17:14) \\
            & ASTR &  65 (30:35) \\
            & EPM &  38 (13:25) \\
            & GBM &  65 (33:32) \\
            & HGGNOS &  65 (26:39) \\
            & LGGNOS &  63 (24:39) \\
            & ODG &  58 (28:30) \\
            & PAST &  35 (18:17) \\
\midrule
Head and Neck Cancer & HNSC &  65 (19:46) \\
\midrule
\multirow{3}{*}{Hepatobiliary Cancer} & CHOL &  65 (26:39) \\
            & GBC &  65 (47:18) \\
            & HCC &  65 (25:40) \\
\midrule
\multirow{3}{*}{Mature B-Cell Neoplasms} & CLLSLL &  26 (13:13) \\
            & DLBCLNOS &  65 (23:42) \\
            & FL &  54 (28:26) \\
\midrule
Mature T and NK Neoplasms & MYCF &  37 (15:22) \\
\midrule
Melanoma & MEL &  65 (28:37) \\
\midrule
\multirow{2}{*}{Mesothelioma} & PLBMESO &   63 (6:57) \\
            & PLEMESO &  65 (20:45) \\
\midrule
Misc. Neuroepithelial Tumor & PGNG &  24 (13:11) \\
\midrule
\multirow{2}{*}{Nerve Sheath Tumor} & MPNST &  40 (17:23) \\
            & SCHW &  38 (23:15) \\
\midrule
\multirow{7}{*}{Non-Small Cell Lung Cancer} & ALUCA &  30 (20:10) \\
            & LUAD &  64 (37:27) \\
            & LUAS &  32 (19:13) \\
            & LUCA &  64 (52:12) \\
            & LUNE &  33 (20:13) \\
            & LUSC &  65 (30:35) \\
            & NSCLCPD &  65 (30:35) \\
\midrule
\multirow{8}{*}{Ovarian Cancer} & CCOV &   65 (65:0) \\
            & EOV &   65 (65:0) \\
            & HGSOC &   65 (65:0) \\
            & LGSOC &   62 (62:0) \\
            & MOV &   33 (33:0) \\
            & MXOV &   49 (49:0) \\
            & OCS &   42 (42:0) \\
            & SBOV &   28 (28:0) \\
\midrule
\multirow{2}{*}{Pancreatic Cancer} & PAAD &  64 (26:38) \\
            & PANET &  65 (31:34) \\
\midrule
Peripheral Nervous System & NBL &  65 (27:38) \\
\midrule
Prostate Cancer & PRAD &   65 (0:65) \\
\end{tabular}
\end{table}

\begin{table}
\centering
\begin{tabular}{p{5cm}l|l}
\toprule
Cancer Type (Cont.) & OncoTree Code & Num. Slides \\
\midrule
\midrule
\multirow{4}{*}{Renal Cell Carcinoma} & CCRCC &  65 (26:39) \\
            & CHRCC &  65 (28:37) \\
            & PRCC &  65 (22:43) \\
            & ROCY &   20 (6:14) \\
\midrule
Salivary Gland Cancer & ACYC &  33 (21:12) \\
\midrule
Sellar Tumor & PTAD &  65 (41:24) \\
\midrule
Sex Cord Stromal Tumor & GRCT &   33 (33:0) \\
\midrule
\multirow{2}{*}{Skin Cancer, Non-Melanoma} & CSCC &  65 (15:50) \\
            & MCC &  65 (23:42) \\
\midrule
Small Bowel Cancer & SBC &  53 (18:35) \\
\midrule
Small Cell Lung Cancer & SCLC &  65 (36:29) \\
\midrule
\multirow{12}{*}{Soft Tissue Sarcoma} & ANGS &  50 (28:22) \\
            & DDLS &  65 (28:37) \\
            & DES &  41 (27:14) \\
            & ERMS &   20 (9:11) \\
            & HEMA &  65 (31:34) \\
            & LMS &  65 (50:15) \\
            & MFH &  60 (24:36) \\
            & MFS &  30 (15:15) \\
            & PECOMA &   21 (18:3) \\
            & SFT &  45 (21:24) \\
            & SYNS &  43 (18:25) \\
            & WDLS &  23 (12:11) \\
\midrule
Thymic Tumor & THYM &  24 (14:10) \\
\midrule
\multirow{4}{*}{Thyroid Cancer} & THAP &  30 (19:11) \\
            & THFO &  28 (15:13) \\
            & THME &  34 (15:19) \\
            & THPA &  65 (39:26) \\
\midrule
Uterine Sarcoma & ULMS &   51 (51:0) \\
\midrule
Vaginal Cancer & VSC &   47 (47:0) \\
\midrule
Wilms Tumor & WT &  63 (29:34) \\
\bottomrule
\end{tabular}
\caption{\textbf{Hierarchical label distribution of the coarse and fine-grained tasks in OncoTree (OT) code cancer classification.} The OT-43 and OT-108 tasks is developed from a dataset comprising 5,564 WSIs from 43 cancer types collected from in-house BWH slides, which are further subdivided into 108 OncoTree codes, with at least 20 WSIs per OncoTree code. The OT-43 task is developed using the first level of the hierarchy (``Cancer Type" column), and the OT-108 task is developed using the second level of the hierarchy (``OncoTree Code" column), with both tasks using the same train-test folds. Gender ratio (female:male) is reported for each OncoTree code.}
\label{tab:op43/108}
\end{table}

\begin{table}
\centering
\begin{tabular}{ll|l}
\toprule
Cancer Subtype & Cancer Diagnosis  & Num. Slides           \\
\midrule\midrule
Benign & Normal &        44 \\
   & Pathological Benign &       147 \\
   & Usual Ductal Hyperplasia &        74 \\
\midrule
Atypical & Atypical Ductal Hyperplasia &        48 \\
   & Flat Epithelial Atypia &        41 \\
\midrule
Malignant & Ductal Carcinoma in Situ &        61 \\
   & Invasive Carcinoma &       132 \\
\bottomrule
\end{tabular}
\caption{\textbf{Hierarchical label distribution of BRCA Coarse- and Fine-Grained Subtyping based on BRACS.} The coarse-grained BRCA subtyping task is developed using the first level of the hierarchy (``Cancer Subtype" column), and the fine-grained BRCA subtyping task is developed using the second level of the hierarchy (``Cancer Diagnosis" column), with both tasks using the official train-validation-test folds of the BRACS dataset.}
\label{tab:brca-label}
\end{table}

\begin{table}
\centering
\begin{tabular}{ll|l}
\toprule
\textit{IDH1} status & Histomolecular Subtype & Num. Slides \\
\midrule\midrule
\multirow{3}{*}{\textit{IDH1}-mutant} & Astrocytoma, \textit{IDH1}-mutant &       257 \\
      & Glioblastoma, \textit{IDH1}-mutant &        93 \\
      & Oligodendroglioma, \textit{IDH1}-mutant \& 1p/19q codeleted &       408 \\
\midrule
\multirow{2}{*}{\textit{IDH1}-wildtype} & Astrocytoma, \textit{IDH1}-wildtype &       144 \\
      & Glioblastoma, \textit{IDH1}-wildtype &      1094 \\
\bottomrule
\end{tabular}
\caption{\textbf{Hierarchical label distribution of Glioma \textit{IDH1} Mutation Prediction and Histomolecular Subtyping based on TCGA and EBRAINS.} The glioma \textit{IDH1} mutation prediction task is developed using the first level of the hierarchy (``\textit{IDH1} status" column), and the glioma histomolecular subtyping task is developed using the second level of the hierarchy (``Histomolecular Subtype" column), with both tasks using same train-validation-test folds created using the TCGA, with external evaluation on the EBRAINS Digital Tumor Atlas.}
\label{tab:ebrains-mut-label}
\end{table}

\begin{table}
\centering
\begin{tabular}{>{\raggedright}p{5cm} >{\raggedright}p{8cm}|l}
\toprule
Cancer Subtype & Cancer Diagnosis & Num. Slides           \\
\midrule\midrule
\multirow{8}{*}{Adult-type diffuse gliomas} & Anaplastic astrocytoma, \textit{IDH1}-mutant &        47 \\
                             & Anaplastic astrocytoma, \textit{IDH1}-wildtype &        47\\
                             & Anaplastic oligodendroglioma, \textit{IDH1}-mutant \& 1p/19q codeleted &        91 \\
                             & Glioblastoma, \textit{IDH1}-mutant &        34 \\
                             & Glioblastoma, \textit{IDH1}-wildtype &       474 \\
                             & Gliosarcoma &        59 \\
                             & Oligodendroglioma, \textit{IDH1}-mutant \& 1p/19q codeleted &        85 \\
\midrule
Circumscribed astrocytic gliomas  & Pilocytic astrocytoma &       173 \\
\midrule
Cranial and paraspinal nerve tumours & Schwannoma &        81 \\
\midrule
Embryonal Tumors & Medulloblastoma, non-WNT/non-SHH &        32 \\
\midrule
\multirow{2}{*}{Ependymal Tumours} & Anaplastic ependymoma &        50 \\
                             & Ependymoma &        46 \\
\midrule
Glioneuronal and neuronal tumours & Ganglioglioma &        88 \\
\midrule
\multirow{2}{*}{\parbox{\hsize}{\raggedright Haematolymphoid tumours involving the CNS}} & Diffuse large B-cell lymphoma of the CNS &        59 \\
                             & Langerhans cell histiocytosis &        32 \\
\midrule
\multirow{7}{*}{Meningiomas} & Anaplastic meningioma &        46 \\
                             & Angiomatous meningioma &        31 \\
                             & Atypical meningioma &        83 \\
                             & Fibrous meningioma &        57 \\
                             & Meningothelial meningioma &       104 \\
                             & Secretory meningioma &        41 \\
                             & Transitional meningioma &        68 \\
\midrule
\multirow{4}{*}{\parbox{\hsize}{\raggedright Mesenchymal, non-meningothelial tumours involving the CNS}} & Haemangioblastoma &        88 \\
                             & Haemangioma &        30 \\
                             & Haemangiopericytoma &        34 \\
                             & Lipoma &        38 \\
\midrule
Metastatic tumours & Metastatic tumours &        47 \\
\midrule
Paediatric-type diffuse low-grade gliomas  & Diffuse astrocytoma, \textit{IDH1}-mutant &        70 \\
\midrule
Tumours of the sellar region & Adamantinomatous craniopharyngioma &        85 \\
                             & Pituitary adenoma &        99 \\
\bottomrule
\end{tabular}
\caption{\textbf{Hierarchical label distribution of brain tumor coarse- and fine-grained subtyping based on EBRAINS.} The coarse-grained brain tumor subtyping task is developed using the first level of the hierarchy (``Cancer Subtype" column), and the fine-grained brain tumor subtyping task is developed using the second level of the hierarchy (``Cancer Diagnosis" column), with both tasks using same train-validation-test folds created using the EBRAINS Digital Tumor Atlas.}
\label{tab:ebrains-diagnosis-label}
\end{table}

\begin{table}[h]
  \centering
  \begin{tabular}{p{7.5cm}|p{3cm}}
    \toprule
    Hyper-parameter & Value \\
    \midrule
    Layers & 24 \\
    Heads & 16 \\
    Patch size & 16 \\
    FFN layer & MLP \\
    Head activation & GELU \\
    Embedding dimension & 1024 \\
    Stochastic dropout rate & 0.1 \\
    \midrule
    Global crop scale & 0.48, 1.0 \\
    Global crop number \& size & 2, 224 \\
    Local crop scale & 0.16, 0.48 \\
    Local crop number \& size & 8, 96 \\
    Max masking ratio & 0.5 \\
    Min masking ratio & 0.1 \\
    Gradient clipping max norm & 3.0 \\
    Normalize last layer & \checkmark \\
    Shared head & \xmark \\
    \midrule
    AdamW $\beta$ & (0.9, 0.999) \\
    Batch size & 3072 \\
    Freeze last layer iterations & 1250 \\
    Warmup iterations & 12500 \\
    Warmup teacher temperature iterations & 37500 \\
    High-resolution finetuning iterations & 12500 \\
    Max Iterations & 125000 \\
    Learning rate schedule & Cosine \\
    Learning rate (start) & 0 \\
    Learning rate (post warmup) & 2e-3 \\
    Learning rate (final) & 1e-6 \\
    Teacher temperature (start) & 0.04 \\
    Teacher temperature (final) & 0.4 \\
    Teacher momentum (start) & 0.992 \\
    Teacher momentum (final) & 1.000 \\
    Weight decay (start) & 0.04 \\
    Weight decay (end) & 0.4 \\
    Automatic mixed precision & FP16 \\
    \bottomrule
  \end{tabular}
  \caption{\textbf{DINOv2 hyperparameters used in \ours pretraining}. 4 $\times$ 80GB NVIDIA A100 GPUs were used for training. Batch size refers to the total batch size across GPUs.}
  \label{tab:hparams_dinov2}
\end{table}

\begin{table}[h]
  \centering
  \begin{tabular}{@{}p{4cm}|rrr|r@{}}
    \toprule
    Hyperparameter & Value \\
    \midrule
    Batch size & 1 \\
    Weight decay & 1e-5 \\
    AdamW $\beta$ & (0.9, 0.999) \\
    Peak learning rate & 1e-4 \\
    Learning rate schedule & Cosine \\
    Epochs & 20 \\
    \bottomrule
  \end{tabular}
  \caption{\textbf{Hyperparameters used in slide-level supervised classification}. Single 24GB NVIDIA GeForce RTX 3090 GPUs were used for each ABMIL model using weakly-supervised learning and slide-level labels.}
  \label{tab:hparams_slide_sup}
\end{table}

\begin{table}
\footnotesize
\begin{tabular}{ll|lllll}
\toprule
Encoder & Full? & Top-1 ACC & Top-3 ACC & Top-5 ACC & Weighted F1 & AUROC \\
\midrule\midrule
ResNet-50$_{\text{IN}}$  & \checkmark & 0.336 (0.312-0.359) & 0.557 (0.531-0.581) & 0.674 (0.651-0.697) & 0.274 (0.251-0.297) & 0.869 (0.858-0.879) \\
CTransPath               & \checkmark & 0.578 (0.554-0.602) & 0.796 (0.777-0.815) & 0.876 (0.859-0.891) & 0.562 (0.536-0.586) & 0.957 (0.950-0.963) \\
REMEDIS                  & \checkmark & 0.040 (0.030-0.049) & 0.157 (0.138-0.175) & 0.204 (0.185-0.224) & 0.012 (0.007-0.019) & 0.727 (0.716-0.736) \\
UNI                      & \checkmark & \bfseries 0.720 (0.698-0.741) & \bfseries 0.890 (0.874-0.906) & \bfseries 0.935 (0.924-0.947) & \bfseries 0.719 (0.695-0.741) & \bfseries 0.976 (0.972-0.981) \\
\midrule
ResNet-50$_{\text{IN}}$  & \xmark & 0.327 (0.304-0.349) & 0.552 (0.527-0.576) & 0.657 (0.633-0.680) & 0.258 (0.236-0.280) & 0.862 (0.850-0.873) \\
CTransPath               & \xmark & 0.569 (0.546-0.592) & 0.793 (0.773-0.812) & 0.874 (0.858-0.890) & 0.555 (0.530-0.578) & 0.956 (0.949-0.962) \\
REMEDIS                  & \xmark & 0.593 (0.570-0.617) & 0.802 (0.783-0.822) & 0.875 (0.859-0.890) & 0.592 (0.567-0.615) & 0.954 (0.946-0.961) \\
UNI                      & \xmark & \bfseries 0.731 (0.709-0.752) & \bfseries 0.894 (0.879-0.910) & \bfseries 0.938 (0.926-0.949) & \bfseries 0.729 (0.706-0.750) & \bfseries 0.976 (0.971-0.981) \\
\bottomrule
\end{tabular}
\caption{\textbf{Weakly-supervised 43-class cancer type classification (OT-43) based on in-house BWH data (43 classes)}. Pre-extracted patch features of each encoder with ABMIL were trained and evaluated on curated train-test folds (3944:1620), with test performance ($n=1620$ slides) reported using top-1, top-3, and top-5 accuracy, weighted F1 score, and AUROC. We report results for both extracted features using all tissue patches per WSI, as well as extracted features from 200 representative tissue patches per WSI (described in \textbf{Online Methods}). Best performing model for each metric is bolded. 95\% CI is included in parentheses.}
\label{tab:ot-43-compare}
\end{table}

\begin{table}
\footnotesize
\begin{tabular}{rr|lllll}
\toprule
Data & Iter. & Top-1 ACC & Top-3 ACC & Top-5 ACC & Weighted F1 & AUROC \\
\midrule
\midrule
Mass-1K & 50K      & 0.640 (0.616-0.662) & 0.843 (0.825-0.860) & 0.908 (0.894-0.922) & 0.638 (0.612-0.659) & 0.968 (0.962-0.973) \\
Mass-1K & 75K      & 0.646 (0.622-0.669) & 0.838 (0.820-0.856) & 0.898 (0.883-0.912) & 0.644 (0.619-0.667) & 0.965 (0.959-0.971) \\
Mass-1K & 100K     & 0.649 (0.625-0.672) & 0.847 (0.829-0.863) & 0.905 (0.890-0.919) & 0.649 (0.624-0.671) & 0.964 (0.957-0.970) \\
Mass-1K & 125K     & 0.652 (0.628-0.676) & 0.835 (0.816-0.852) & 0.895 (0.880-0.910) & 0.653 (0.627-0.676) & 0.962 (0.955-0.968) \\
\midrule
Mass-22K & 50K     & 0.707 (0.685-0.730) & 0.887 (0.870-0.902) & 0.930 (0.918-0.943) & 0.704 (0.680-0.728) & 0.974 (0.968-0.979) \\
Mass-22K & 75K     & 0.708 (0.684-0.730) & 0.881 (0.865-0.896) & 0.930 (0.917-0.941) & 0.707 (0.682-0.728) & 0.972 (0.967-0.977) \\
Mass-22K & 100K    & 0.708 (0.686-0.730) & 0.888 (0.873-0.904) & 0.938 (0.925-0.949) & 0.705 (0.682-0.726) & 0.973 (0.967-0.978) \\
Mass-22K & 125K    & 0.694 (0.673-0.716) & 0.891 (0.875-0.905) & 0.931 (0.918-0.943) & 0.692 (0.670-0.714) & 0.969 (0.963-0.975) \\
\midrule
Mass-100K & 50K    & 0.694 (0.672-0.716) & 0.871 (0.855-0.888) & 0.927 (0.913-0.939) & 0.694 (0.669-0.715) & 0.974 (0.968-0.978) \\
Mass-100K & 75K    & 0.704 (0.682-0.725) & 0.889 (0.873-0.904) & 0.935 (0.923-0.948) & 0.703 (0.679-0.724) & 0.977 (0.972-0.981) \\
Mass-100K & 100K   & 0.720 (0.699-0.742) & 0.886 (0.871-0.901) & 0.931 (0.919-0.943) & 0.719 (0.697-0.741) & \bfseries 0.977 (0.972-0.981) \\
Mass-100K & 125K   & \bfseries 0.731 (0.709-0.752) & \bfseries 0.894 (0.879-0.910) & \bfseries 0.938 (0.926-0.949) & \bfseries 0.729 (0.706-0.750) & \bfseries 0.976 (0.972-0.981) \\
\bottomrule
\end{tabular}
\caption{\textbf{Assessing pretraining length of \ours (pretrained on Mass-1K, Mass-22K, and Mass-100K) on OT-43 performance.} Comparisons of weakly-supervised performance on OT-43 of \ours pretrained on different data sizes (Mass-1K, Mass-22K, Mass-100K), across different training iterations ranging from 50K to 125K. Pre-extracted patch features of each encoder with ABMIL were trained and evaluated on curated train-test folds (3944:1620), with test performance ($n=1620$ slides) reported using top-1, top-3, and top-5 accuracy, weighted F1 score, and AUROC. Best performing model for each metric is bolded. 95\% CI is included in parentheses.}
\label{tab:ot-43-length}
\end{table}

\begin{table}
\footnotesize\centering
\begin{tabular}{ll|lllll}
\toprule
Encoder & Full? & Top-1 ACC & Top-3 ACC & Top-5 ACC & Weighted F1 & AUROC \\
\midrule\midrule
ResNet-50$_{\text{IN}}$ & \checkmark & 0.156 (0.138-0.173) & 0.309 (0.287-0.333) & 0.391 (0.368-0.416) & 0.115 (0.101-0.132) & 0.874 (0.866-0.882) \\
CTransPath & \checkmark & 0.391 (0.366-0.415) & 0.620 (0.597-0.645) & 0.715 (0.694-0.736) & 0.358 (0.333-0.383) & 0.958 (0.954-0.963) \\
REMEDIS & \checkmark & 0.118 (0.102-0.133) & 0.225 (0.205-0.244) & 0.293 (0.270-0.314) & 0.074 (0.060-0.087) & 0.857 (0.849-0.867) \\
UNI & \checkmark & \bfseries 0.525 (0.501-0.549) & \bfseries 0.751 (0.728-0.772) & \bfseries 0.829 (0.812-0.847) & \bfseries 0.509 (0.483-0.534) & \bfseries 0.971 (0.967-0.975) \\
\midrule
ResNet-50$_{\text{IN}}$ & \xmark & 0.148 (0.130-0.164) & 0.299 (0.277-0.322) & 0.378 (0.354-0.403) & 0.105 (0.091-0.121) & 0.869 (0.860-0.877) \\
CTransPath & \xmark & 0.399 (0.375-0.423) & 0.625 (0.602-0.649) & 0.723 (0.702-0.746) & 0.365 (0.342-0.389) & 0.959 (0.955-0.963) \\
REMEDIS & \xmark & 0.412 (0.387-0.435) & 0.654 (0.630-0.678) & 0.735 (0.713-0.757) & 0.398 (0.372-0.421) & 0.952 (0.946-0.956) \\
UNI & \xmark & \bfseries 0.538 (0.514-0.562) & \bfseries 0.759 (0.738-0.781) & \bfseries 0.843 (0.826-0.860) & \bfseries 0.522 (0.498-0.548) & \bfseries 0.972 (0.968-0.976) \\
\bottomrule
\end{tabular}
\caption{\textbf{Weakly-supervised 108-class OncoTree code cancer classification (OT-108) based on in-house BWH data (108 classes)}. Pre-extracted patch features of each encoder with ABMIL were trained and evaluated on curated train-test folds (3944:1620), with test performance ($n=1620$ slides) reported using top-1, top-3, and top-5 accuracy, weighted F1 score, and AUROC. We report results for both extracted features using all tissue patches per WSI, as well as extracted features from 200 representative tissue patches per WSI (described in \textbf{Online Methods}). Best performing model for each metric is bolded. 95\% CI is included in parentheses. We }
\label{tab:ot-108-compare}
\end{table}

\begin{table}
\footnotesize\centering
\begin{tabular}{rr|lllll}
\toprule
Data Size & Iter & Top-1 ACC & Top-3 ACC & Top-5 ACC & Weighted F1 & AUROC \\
\midrule\midrule
Mass-1K & 50K       & 0.459 (0.435-0.486) & 0.679 (0.657-0.702) & 0.767 (0.748-0.787) & 0.436 (0.411-0.462) & 0.968 (0.964-0.971) \\
Mass-1K & 75K       & 0.464 (0.440-0.488) & 0.681 (0.660-0.706) & 0.757 (0.736-0.778) & 0.440 (0.416-0.466) & 0.967 (0.963-0.971) \\
Mass-1K & 100K      & 0.458 (0.434-0.482) & 0.675 (0.654-0.699) & 0.760 (0.740-0.781) & 0.436 (0.412-0.461) & 0.964 (0.960-0.968) \\
Mass-1K & 125K      & 0.473 (0.449-0.499) & 0.673 (0.652-0.698) & 0.757 (0.738-0.780) & 0.455 (0.430-0.482) & 0.961 (0.956-0.965) \\
\midrule
Mass-22K & 50K      & 0.511 (0.487-0.536) & 0.752 (0.732-0.775) & 0.830 (0.812-0.848) & 0.493 (0.467-0.519) & \bfseries 0.976 (0.972-0.979) \\
Mass-22K & 75K      & 0.522 (0.498-0.549) & 0.748 (0.727-0.770) & 0.840 (0.821-0.857) & 0.498 (0.474-0.524) & 0.976 (0.972-0.979) \\
Mass-22K & 100K     & 0.512 (0.489-0.538) & 0.747 (0.726-0.770) & 0.830 (0.811-0.848) & 0.491 (0.465-0.517) & 0.974 (0.971-0.977) \\
Mass-22K & 125K     & 0.508 (0.485-0.531) & 0.740 (0.719-0.761) & 0.825 (0.807-0.842) & 0.486 (0.461-0.510) & 0.972 (0.968-0.975) \\
\midrule
Mass-100K & 50K     & 0.511 (0.486-0.536) & 0.741 (0.720-0.764) & 0.827 (0.809-0.846) & 0.491 (0.465-0.516) & 0.974 (0.971-0.977) \\
Mass-100K & 75K     & 0.523 (0.499-0.548) & 0.762 (0.741-0.782) & 0.837 (0.820-0.854) & 0.507 (0.481-0.533) & 0.975 (0.972-0.979) \\
Mass-100K & 100K    & 0.533 (0.508-0.559) & \bfseries 0.768 (0.748-0.788) & 0.841 (0.823-0.859) & 0.516 (0.491-0.542) & 0.975 (0.971-0.978) \\
Mass-100K & 125K    & \bfseries 0.538 (0.514-0.562) & 0.759 (0.738-0.781) & \bfseries 0.843 (0.826-0.860) & \bfseries 0.522 (0.498-0.548) & 0.972 (0.968-0.976) \\
\bottomrule
\end{tabular}
\caption{\textbf{Assessing pretraining length of \ours (pretrained on Mass-1K, Mass-22K, and Mass-100K) on OT-108 performance.}  Comparisons of weakly-supervised performance on OT-108 of \ours pretrained on different data sizes (Mass-1K, Mass-22K, Mass-100K), across different training iterations ranging from 50K to 125K. Pre-extracted patch features of each encoder with ABMIL were trained and evaluated on curated train-test folds (3944:1620), with test performance ($n=1620$ slides) reported using top-1, top-3, and top-5 accuracy, weighted F1 score, and AUROC. Best performing model for each metric is bolded. 95\% CI is included in parentheses.}
\label{tab:ot-108-length}
\end{table}
\begin{table}
\footnotesize\centering
\begin{tabular}{l|lll}
\toprule
Encoder & Balanced ACC & Weighted F1 & AUROC \\
\midrule\midrule
ResNet-50$_{\text{IN}}$ & 0.726 (0.652-0.797) & 0.765 (0.678-0.837) & 0.752 (0.659-0.842) \\
CTransPath & 0.897 (0.836-0.948) & 0.907 (0.851-0.953) & 0.930 (0.864-0.985) \\
REMEDIS & 0.930 (0.884-0.969) & 0.923 (0.877-0.962) & \bfseries 0.981 (0.947-1.000) \\
UNI & \bfseries 0.957 (0.911-0.991) & \bfseries 0.961 (0.922-0.992) & 0.975 (0.937-1.000) \\
\bottomrule
\end{tabular}
\caption{\textbf{Weakly-supervised breast metastasis detection based on CAMELYON16 (2 classes)}. Pre-extracted patch features of each encoder with ABMIL were trained a custom train-validation split (90:10 ratio) of the official train set (270 slides) and tested on the official test set, with test performance ($n=129$) reported using balanced accuracy, weighted F1 score, and AUROC. Best performing model for each metric is bolded. 95\% CI is included in parentheses.}
\label{tab:slide-c16}
\end{table}

\begin{table}
\footnotesize\centering
\begin{tabular}{ll|lll}
\toprule
Encoder & Cohort & Balanced ACC & Weighted F1 & AUROC \\
\midrule\midrule
ResNet-50$_{\text{IN}}$ & TCGA & 0.857 (0.787-0.929) & 0.857 (0.786-0.929) & 0.939 (0.893-0.978) \\
\rowcolor{gray!10} CTransPath & TCGA & 0.949 (0.901-0.990) & 0.949 (0.898-0.990) & 0.993 (0.982-1.000) \\
\rowcolor{gray!10} REMEDIS & TCGA & 0.949 (0.902-0.990) & 0.949 (0.898-0.990) & 0.985 (0.953-1.000) \\
UNI & TCGA & \bfseries 0.969 (0.932-1.000) & \bfseries 0.969 (0.929-1.000) & \bfseries 0.997 (0.989-1.000) \\
\midrule
ResNet-50$_{\text{IN}}$ & CPTAC & 0.852 (0.831-0.871) & 0.849 (0.829-0.869) & 0.924 (0.907-0.939) \\
CTransPath & CPTAC & 0.884 (0.866-0.903) & 0.881 (0.862-0.900) & 0.953 (0.942-0.963) \\
REMEDIS & CPTAC & 0.854 (0.833-0.873) & 0.849 (0.828-0.869) & 0.951 (0.938-0.962) \\
UNI & CPTAC & \bfseries 0.889 (0.870-0.908) & \bfseries 0.891 (0.872-0.909) & \bfseries 0.958 (0.946-0.969) \\
\bottomrule
\end{tabular}
\caption{\textbf{Weakly-supervised NSCLC subtyping based on TCGA and CPTAC (2 classes)}. Pre-extracted patch features of each encoder with ABMIL were trained and evaluated on the TCGA-NSCLC cohort (label-stratified into train-validation-test folds with 80:10:10 ratio, 848:97:98 slides), with external evaluation on slides ($n=1091$) sourced from CPTAC-LUAD ($n=578$) and CPTAC-LUSC ($n=513$). Test performance was reported using balanced accuracy, weighted F1 score, and AUROC. Best performing model for each metric is bolded. 95\% CI is included in parentheses.}
\label{tab:slide-nsclc}
\end{table}

\begin{table}
\footnotesize\centering
\begin{tabular}{ll|lll}
\toprule
Encoder & Cohort & Balanced ACC & Weighted F1 & AUROC \\
\midrule\midrule
ResNet-50$_{\text{IN}}$ & TCGA & 0.849 (0.734-0.940) & 0.877 (0.814-0.938) & 0.977 (0.954-0.993) \\
\rowcolor{gray!10} CTransPath & TCGA & 0.892 (0.800-0.969) & 0.907 (0.847-0.959) & 0.987 (0.970-0.998) \\
\rowcolor{gray!10} REMEDIS & TCGA & \bfseries 0.973 (0.937-1.000) & \bfseries 0.969 (0.927-1.000) & \bfseries 0.997 (0.992-1.000) \\
UNI & TCGA & 0.947 (0.875-0.994) & 0.959 (0.918-0.990) & 0.994 (0.984-1.000) \\
\midrule
ResNet-50$_{\text{IN}}$ & CPTAC-DHMC & 0.824 (0.751-0.894) & 0.903 (0.886-0.921) & 0.972 (0.957-0.984) \\
CTransPath & CPTAC-DHMC & 0.939 (0.888-0.984) & 0.968 (0.957-0.979) & \bfseries 0.996 (0.993-0.998) \\
REMEDIS & CPTAC-DHMC & 0.790 (0.725-0.858) & 0.934 (0.919-0.949) & 0.988 (0.982-0.993) \\
UNI & CPTAC-DHMC & \bfseries 0.963 (0.930-0.987) & \bfseries 0.971 (0.960-0.981) & 0.993 (0.985-0.998) \\
\bottomrule
\end{tabular}
\caption{\textbf{Weakly-supervised RCC subtyping based on TCGA and CPTAC-DHMC (3 classes)}. Pre-extracted patch features of each encoder with ABMIL were trained and evaluated on the TCGA-RCC cohort (label-stratified into train-validation-test folds with 80:10:10 ratio, 736:89:97 slides), with external evaluation on slides $(n=872)$ sourced from CPTAC-CCRCC $(n=404)$ and DHMC-Kidney $(n=468)$ (CCRCC, CHRCC, and PRCC cases only). Test performance was reported using balanced accuracy, weighted F1 score, and AUROC. Best performing model for each metric is bolded. 95\% CI is included in parentheses.}
\label{tab:slide-rcc}
\end{table}

\begin{table}
\footnotesize\centering
\begin{tabular}{l|lll}
\toprule
Encoder & Balanced ACC & Weighted F1 & AUROC \\
\midrule\midrule
ResNet-50$_{\text{IN}}$ & 0.346 (0.306-0.386) & 0.478 (0.383-0.565) & 0.818 (0.785-0.850) \\
CTransPath & 0.804 (0.743-0.879) & 0.883 (0.823-0.937) & 0.987 (0.976-0.995) \\
REMEDIS & 0.865 (0.789-0.929) & 0.877 (0.818-0.929) & 0.973 (0.953-0.988) \\
UNI & \bfseries 0.919 (0.852-0.968) & \bfseries 0.926 (0.882-0.966) & \bfseries 0.993 (0.983-0.999) \\
\bottomrule
\end{tabular}
\caption{\textbf{Weakly-supervised RCC subtyping based on DHMC (5 classes)}. Pre-extracted patch features of each encoder with ABMIL were trained and evaluated on a modified configuration of the official train-validation-test folds (70:4:26 ratio, 393:23:147 slides), with test performance ($n=147$ slides) reported using balanced accuracy, weighted F1 score, and AUROC. Since no CHRCC slides were included in the validation fold, 8 CHRCC slides from the training fold was moved to the valdiation fold. The test fold was unmodified. Best performing model for each metric is bolded. 95\% CI is included in parentheses.}
\label{tab:slide-dhmc}
\end{table}

\begin{table}
\footnotesize\centering
\begin{tabular}{l|lll}
\toprule
Encoder & Balanced ACC & Weighted F1 & AUROC \\
\midrule\midrule
ResNet-50$_{\text{IN}}$ & 0.250 (0.250-0.250) & 0.356 (0.209-0.525) & 0.671 (0.594-0.754) \\
CTransPath & 0.556 (0.454-0.651) & 0.728 (0.603-0.841) & 0.845 (0.779-0.898) \\
REMEDIS & 0.604 (0.499-0.696) & 0.787 (0.670-0.889) & 0.888 (0.814-0.960) \\
UNI & \bfseries 0.643 (0.549-0.725) & \bfseries 0.824 (0.706-0.931) & \bfseries 0.957 (0.908-0.987) \\
\bottomrule
\end{tabular}
\caption{\textbf{Weakly-supervised CRC screening based on HunCRC (4 classes)}. Pre-extracted patch features of each encoder with ABMIL were trained and evaluated on the HunCRC cohort (label-stratified into train-validation-test folds with 50:25:25 ratio, 100:50:50 slides). Test performance was reported using balanced accuracy, weighted F1 score, and AUROC. Best performing model for each metric is bolded. 95\% CI is included in parentheses.}
\label{tab:slide-huncrc}
\end{table}

\begin{table}
\footnotesize\centering
\begin{tabular}{l|lll}
\toprule
Encoder & Balanced ACC & Weighted F1 & AUROC \\
\midrule\midrule
ResNet-50$_{\text{IN}}$ & 0.552 (0.490-0.607) & 0.516 (0.396-0.629) & 0.748 (0.675-0.816) \\
CTransPath & 0.639 (0.554-0.728) & 0.648 (0.536-0.763) & 0.840 (0.776-0.903) \\
REMEDIS & 0.676 (0.578-0.770) & \bfseries 0.696 (0.591-0.796) & 0.864 (0.805-0.915) \\
UNI & \bfseries 0.687 (0.617-0.760) & 0.691 (0.562-0.800) & \bfseries 0.887 (0.833-0.936) \\
\bottomrule
\end{tabular}
\caption{\textbf{Weakly-supervised BRCA coarse-grained subtyping based on BRACS (3 classes)}. Pre-extracted patch features of each encoder with ABMIL were trained and evaluated on the official train-validation-test folds (72:12:16 ratio, 395:65:87 slides), with test performance ($n=87$ slides) reported using balanced accuracy, weighted F1 score, and AUROC. Best performing model for each metric is bolded. 95\% CI is included in parentheses.}
\label{tab:slide-bracs-c}
\end{table}

\begin{table}
\footnotesize\centering
\begin{tabular}{l|lll}
\toprule
Encoder & Balanced ACC & Weighted F1 & AUROC \\
\midrule\midrule
ResNet-50$_{\text{IN}}$ & 0.248 (0.214-0.276) & 0.216 (0.132-0.311) & 0.701 (0.645-0.754) \\
CTransPath & 0.360 (0.284-0.440) & 0.377 (0.280-0.485) & 0.761 (0.706-0.809) \\
REMEDIS & 0.398 (0.307-0.483) & 0.428 (0.320-0.530) & 0.749 (0.689-0.804) \\
UNI & \bfseries 0.468 (0.393-0.550) & \bfseries 0.486 (0.366-0.592) & \bfseries 0.837 (0.794-0.879) \\
\bottomrule
\end{tabular}
\caption{\textbf{Weakly-supervised BRCA fine-grained subtyping based on BRACS (7 classes)}. Pre-extracted patch features of each encoder with ABMIL were trained and evaluated on the official train-validation-test (72:12:16 ratio, 395:65:87 slides), with test performance ($n=87$ slides) reported using balanced accuracy, weighted F1 score, and AUROC. Best performing model for each metric is bolded. 95\% CI is included in parentheses.}
\label{tab:slide-bracs-f}
\end{table}

\begin{table}
\footnotesize\centering
\begin{tabular}{ll|lll}
\toprule
Encoder & Cohort & Balanced ACC & Weighted F1 & AUROC \\
\midrule\midrule
ResNet-50$_{\text{IN}}$ & TCGA & 0.768 (0.710-0.821) & 0.810 (0.767-0.850) & 0.864 (0.825-0.901) \\
\rowcolor{gray!10} CTransPath & TCGA & \bfseries 0.891 (0.859-0.920) & \bfseries 0.862 (0.828-0.894) & \bfseries 0.958 (0.938-0.975) \\
\rowcolor{gray!10} REMEDIS & TCGA & 0.819 (0.770-0.869) & 0.860 (0.824-0.897) & 0.933 (0.907-0.956) \\
UNI & TCGA & 0.808 (0.760-0.855) & 0.831 (0.794-0.867) & 0.925 (0.898-0.949) \\
\midrule
ResNet-50$_{\text{IN}}$ & EBRAINS & 0.759 (0.731-0.786) & 0.777 (0.749-0.804) & 0.822 (0.794-0.849) \\
CTransPath & EBRAINS & 0.836 (0.814-0.861) & 0.819 (0.795-0.846) & 0.924 (0.905-0.941) \\
REMEDIS & EBRAINS & 0.792 (0.765-0.818) & 0.822 (0.795-0.850) & 0.895 (0.873-0.916) \\
UNI & EBRAINS & \bfseries 0.856 (0.836-0.878) & \bfseries 0.841 (0.820-0.864) & \bfseries 0.941 (0.925-0.955) \\
\bottomrule
\end{tabular}
\caption{\textbf{Weakly-supervised GBMLGG \textit{IDH1} mutation prediction based on TCGA and EBRAINS (2 classes)}. Pre-extracted patch features of each encoder with ABMIL were trained and evaluated on the TCGA-GBMLGG cohort (site-stratified into train-validation-test folds with 47:22:33 ratio, 525:243:355 slides), with external evaluation on slides ($n=873$) sourced from the EBRAINS Digital Tumor Atlas (using slides with available \textit{IDH1} status). Test performance was reported using balanced accuracy, weighted F1 score, and AUROC. Best performing model for each metric is bolded. 95\% CI is included in parentheses.}
\label{tab:slide-idh}
\end{table}

\begin{table}
\footnotesize\centering
\begin{tabular}{ll|lll}
\toprule
Encoder & Cohort & Balanced ACC & Weighted F1 & AUROC \\
\midrule\midrule
ResNet-50$_{\text{IN}}$ & TCGA & 0.412 (0.362-0.462) & 0.660 (0.611-0.715) & 0.814 (0.783-0.847) \\
\rowcolor{gray!10} CTransPath & TCGA & 0.487 (0.437-0.543) & 0.718 (0.671-0.769) & 0.866 (0.831-0.905) \\
\rowcolor{gray!10} REMEDIS & TCGA & 0.415 (0.382-0.448) & 0.675 (0.628-0.727) & 0.837 (0.811-0.864) \\
UNI & TCGA & \bfseries 0.673 (0.616-0.732) & \bfseries 0.794 (0.751-0.838) & \bfseries 0.910 (0.880-0.939) \\
\midrule
ResNet-50$_{\text{IN}}$ & EBRAINS & 0.402 (0.379-0.423) & 0.615 (0.581-0.648) & 0.684 (0.654-0.713) \\
CTransPath & EBRAINS & 0.498 (0.472-0.523) & 0.705 (0.674-0.735) & 0.823 (0.798-0.845) \\
REMEDIS & EBRAINS & 0.337 (0.320-0.354) & 0.551 (0.517-0.586) & 0.751 (0.723-0.777) \\
UNI & EBRAINS & \bfseries 0.562 (0.535-0.594) & \bfseries 0.728 (0.702-0.755) & \bfseries 0.868 (0.848-0.887) \\
\bottomrule
\end{tabular}
\caption{\textbf{Weakly-supervised GBMLGG histomolecular subtyping based on TCGA and EBRAINS (5 classes)}. Pre-extracted patch features of each encoder with ABMIL were trained and evaluated on the TCGA-GBMLGG cohort (site-stratified into train-validation-test folds with 47:22:33 ratio, 525:243:355 slides), with external evaluation on slides ($n=873$) sourced from the EBRAINS Digital Tumor Atlas (using slides with available \textit{IDH1} status). Test performance was reported using balanced accuracy, weighted F1 score, and AUROC. Best performing model for each metric is bolded. 95\% CI is included in parentheses.}
\label{tab:slide-molsub}
\end{table}

\begin{table}
\footnotesize\centering
\begin{tabular}{l|lll}
\toprule
Encoder & Balanced ACC & Weighted F1 & AUROC \\
\midrule\midrule
ResNet-50$_{\text{IN}}$ & 0.302 (0.273-0.334) & 0.556 (0.509-0.600) & 0.878 (0.860-0.896) \\
CTransPath & 0.666 (0.608-0.722) & 0.795 (0.758-0.829) & 0.968 (0.957-0.978) \\
REMEDIS & 0.687 (0.638-0.734) & 0.789 (0.753-0.824) & 0.967 (0.952-0.979) \\
UNI & \bfseries 0.883 (0.838-0.924) & \bfseries 0.926 (0.902-0.947) & \bfseries 0.996 (0.994-0.998) \\
\bottomrule
\end{tabular}
\caption{\textbf{Weakly-supervised brain tumor subtyping based on EBRAINS (12 classes)}. Pre-extracted patch features of each encoder with ABMIL were trained and evaluated on the EBRAINS Digital Tumor Atlas (label-stratified into train-validation-test folds with 50:25:25 ratio, 1151:595:573 slides), with test performance reported using balanced accuracy, weighted F1 score, and AUROC. Best performing model for each metric is bolded. 95\% CI is included in parentheses.}
\label{tab:slide-ebrains-c}
\end{table}

\begin{table}
\footnotesize\centering
\begin{tabular}{l|lll}
\toprule
Encoder & Balanced ACC & Weighted F1 & AUROC \\
\midrule\midrule
ResNet-50$_{\text{IN}}$ & 0.219 (0.195-0.241) & 0.300 (0.259-0.339) & 0.893 (0.879-0.907) \\
CTransPath & 0.514 (0.473-0.559) & 0.597 (0.548-0.638) & 0.959 (0.950-0.967) \\
REMEDIS & 0.382 (0.346-0.415) & 0.471 (0.428-0.512) & 0.917 (0.901-0.931) \\
UNI & \bfseries 0.675 (0.633-0.715) & \bfseries 0.746 (0.704-0.783) & \bfseries 0.976 (0.969-0.982) \\
\bottomrule
\end{tabular}
\caption{\textbf{Weakly-supervised brain tumor subtyping based on EBRAINS (30 classes)}. Pre-extracted patch features of each encoder with ABMIL were trained and evaluated on the EBRAINS Digital Tumor Atlas (label-stratified into train-validation-test folds with 50:25:25 ratio, 1151:595:573 slides), with test performance reported using balanced accuracy, weighted F1 score, and AUROC. Best performing model for each metric is bolded. 95\% CI is included in parentheses.}
\label{tab:slide-ebrains-f}
\end{table}

\begin{table}
\footnotesize\centering
\begin{tabular}{l|llll}
\toprule
Encoder & Balanced ACC & Quad. Weighted $\kappa$ & Weighted F1 & AUROC \\
\midrule\midrule
ResNet-50$_{\text{IN}}$ & 0.574 (0.544-0.603) & 0.831 (0.799-0.859) & 0.631 (0.599-0.661) & 0.885 (0.873-0.897) \\
CTransPath & 0.691 (0.658-0.723) & 0.927 (0.912-0.940) & 0.752 (0.723-0.779) & 0.938 (0.929-0.947) \\
REMEDIS & 0.711 (0.679-0.742) & 0.932 (0.918-0.945) & 0.766 (0.737-0.794) & 0.941 (0.931-0.949) \\
UNI & \bfseries 0.757 (0.726-0.785) & \bfseries 0.946 (0.933-0.957) & \bfseries 0.809 (0.783-0.834) & \bfseries 0.956 (0.947-0.963) \\
\bottomrule
\end{tabular}
\caption{\textbf{Weakly-supervised ISUP grading based on PANDA (6 classes)}. Pre-extracted patch features of each encoder with ABMIL were trained and evaluated on label-stratified train-validation-test folds (80:10:10 ratio, 7647:954:954 slides), with test performance (954 slides) reported using balanced accuracy, Cohen's quadratic weighted $\kappa$, weighted F1 core, and ROC AUC. Best performing model for each metric is bolded. 95\% CI is included in parentheses.}
\label{tab:slide-panda}
\end{table}

\begin{table}
\footnotesize\centering
\begin{tabular}{l|lll}
\toprule
Encoder & Balanced ACC & Weighted F1 & AUROC \\
\midrule\midrule
ResNet-50$_{\text{IN}}$ & 0.861 (0.822-0.898) & 0.861 (0.822-0.898) & 0.930 (0.903-0.955) \\
CTransPath & 0.882 (0.847-0.916) & 0.882 (0.846-0.916) & 0.937 (0.909-0.961) \\
REMEDIS & 0.861 (0.823-0.895) & 0.861 (0.822-0.895) & 0.933 (0.908-0.958) \\
UNI & \bfseries 0.896 (0.864-0.927) & \bfseries 0.894 (0.862-0.925) & \bfseries 0.962 (0.944-0.979) \\
\bottomrule
\end{tabular}
\caption{\textbf{Weakly-supervised cellular-mediated allograft rejection of endomyocardial biopsies based on in-house BWH data (2 classes)}. Pre-extracted patch features of each encoder with ABMIL were trained and evaluated on case- and label-stratified train-validation-test folds (70:10:20 ratio, 3547:484:900 slides), with test performance (900 slides) reported using balanced accuracy, weighted F1 core, and ROC AUC. Best performing model for each metric is bolded. 95\% CI is included in parentheses.}
\label{tab:slide-emb}
\end{table}

\begin{table}
\footnotesize\centering
\begin{tabular}{l|lll}
\toprule
Encoder & Balanced ACC & Weighted F1 & AUROC \\
\midrule\midrule
ResNet-50$_{\text{IN}}$ & 0.715 (0.705-0.724) & 0.771 (0.761-0.781) & 0.958 (0.954-0.961) \\
CTransPath & 0.845 (0.836-0.853) & 0.867 (0.859-0.875) & \bfseries 0.991 (0.990-0.992) \\
REMEDIS & 0.787 (0.776-0.796) & 0.802 (0.793-0.811) & 0.980 (0.979-0.982) \\
UNI & \bfseries 0.874 (0.866-0.881) & \bfseries 0.875 (0.868-0.882) & 0.990 (0.988-0.991) \\
\bottomrule
\end{tabular}
\caption{\textbf{Linear probe evaluation for CRC tissue classification based on CRC-100K (9 classes)}. Pre-extracted patch features of each encoder with logistic regression were evaluated on the official train-test folds (100,000:7,180), with test performance (n=7,180 ROIs) reported using balanced accuracy, weighted F1 score, and AUROC. Best performing model for each metric is bolded. 95\% CI is included in parentheses.}
\label{tab:patch-crc100k-lin}
\end{table}

\begin{table}
\footnotesize\centering
\begin{tabular}{l|llll}
\toprule
Encoder  & 1-NN Balanced ACC & 1-NN Weighted F1 & 20-NN Balanced ACC & 20-NN Weighted F1 \\
\midrule\midrule
ResNet-50$_{\text{IN}}$ & 0.686 (0.676-0.696) & 0.698 (0.686-0.708) & 0.797 (0.787-0.806) & 0.833 (0.825-0.841) \\
CTransPath & 0.815 (0.806-0.824) & 0.845 (0.836-0.854) & 0.836 (0.828-0.843) & 0.848 (0.840-0.857) \\
REMEDIS & 0.855 (0.848-0.863) & 0.881 (0.873-0.888) & 0.908 (0.901-0.915) & 0.924 (0.918-0.931) \\
UNI & \bfseries 0.899 (0.892-0.905) & \bfseries 0.912 (0.906-0.919) & \bfseries 0.924 (0.917-0.931) & \bfseries 0.945 (0.940-0.950) \\
\bottomrule
\end{tabular}
\caption{\textbf{Few-shot and KNN evaluation for CRC tissue classification based on CRC-100K (9 classes)}. Pre-extracted patch features of each encoder with SimpleShot ($K$=1) and Nearest Neighbors ($K$=20) were evaluated on the official train-test folds (100,000:7,180 ROIs), with test performance (n=7,180 ROIs) reported using balanced accuracy and weighted F1 score. Best performing model for each metric is bolded. 95\% CI is included in parentheses.}
\label{tab:patch-crc100k-knn}
\end{table}

\begin{table}
\footnotesize\centering
\begin{tabular}{l|lll}
\toprule
Encoder & Balanced ACC & Weighted F1 & AUROC \\
\midrule\midrule
ResNet-50$_{\text{IN}}$ & 0.770 (0.761-0.779) & 0.733 (0.722-0.745) & 0.937 (0.932-0.941) \\
CTransPath & 0.802 (0.791-0.812) & 0.783 (0.773-0.793) & 0.939 (0.934-0.943) \\
REMEDIS & 0.729 (0.717-0.740) & 0.744 (0.733-0.754) & 0.906 (0.899-0.912) \\
UNI  & \bfseries 0.890 (0.883-0.898) & \bfseries 0.880 (0.871-0.888) & \bfseries 0.979 (0.977-0.981) \\
\bottomrule
\end{tabular}
\caption{\textbf{Linear probe evaluation for CCRCC tissue classification based on TCGA and HEL (3 classes)}. Pre-extracted patch features of each encoder with logistic regression were evaluated on the TCGA cohort as the train fold and HEL cohort as the test fold (89:11 ratio, 24,201:2,968 ROIs), with test performance (n=2,968 ROIs) reported using balanced accuracy, weighted F1 score, and AUROC. Best performing model for each metric is bolded. 95\% CI is included in parentheses.}
\label{tab:patch-ccrcc-lin}
\end{table}

\begin{table}
\footnotesize\centering
\begin{tabular}{l|llll}
\toprule
Encoder & 1-NN Balanced ACC & 1-NN Weighted F1 & 20-NN Balanced ACC & 20-NN Weighted F1 \\
\midrule\midrule
ResNet-50$_{\text{IN}}$ & 0.765 (0.755-0.775) & 0.741 (0.730-0.751) & 0.680 (0.670-0.690) & 0.641 (0.629-0.653) \\
CTransPath & 0.679 (0.667-0.691) & 0.653 (0.639-0.665) & 0.697 (0.686-0.708) & 0.679 (0.668-0.690) \\
REMEDIS & 0.785 (0.775-0.795) & 0.788 (0.778-0.798) & 0.833 (0.823-0.843) & 0.827 (0.816-0.836) \\
UNI  & \bfseries 0.801 (0.790-0.810) & \bfseries 0.801 (0.791-0.811) & \bfseries 0.866 (0.857-0.875) & \bfseries 0.860 (0.851-0.868) \\
\bottomrule
\end{tabular}
\caption{\textbf{Few-shot and KNN evaluation for CCRCC tissue classification based on TCGA and HEL (3 classes)}. Pre-extracted patch features of each encoder with SimpleShot ($K$=1) and Nearest Neighbors ($K$=20) were evaluated on the TCGA cohort as the train fold and HEL cohort as the test fold (89:11 ratio, 24,201:2,968 ROIs), with test performance (n=2,968 ROIs) reported using balanced accuracy and weighted F1 score. Best performing model for each metric is bolded. 95\% CI is included in parentheses.}
\label{tab:patch-ccrcc-knn}
\end{table}

\begin{table}
\footnotesize\centering
\begin{tabular}{ll|lll}
\toprule
Encoder & Img Res. & Balanced ACC & Weighted F1 & AUROC \\
\midrule\midrule
ResNet-50$_{\text{IN}}$ & 224 & 0.738 (0.634-0.824) & 0.733 (0.630-0.828) & 0.911 (0.854-0.954) \\
CTransPath & 224 & 0.875 (0.804-0.939) & 0.872 (0.796-0.937) & 0.982 (0.967-0.994) \\
REMEDIS & 224 & 0.863 (0.782-0.933) & 0.864 (0.786-0.937) & 0.981 (0.959-0.997) \\
UNI & 224 & \bfseries 0.925 (0.856-0.976) & \bfseries 0.926 (0.866-0.975) & \bfseries 0.994 (0.983-1.000) \\
\midrule
ResNet-50$_{\text{IN}}$ & 448 & 0.712 (0.610-0.810) & 0.714 (0.610-0.815) & 0.901 (0.841-0.951) \\
CTransPath & 448 & 0.863 (0.785-0.933) & 0.864 (0.790-0.937) & 0.973 (0.946-0.992) \\
REMEDIS & 448 & 0.913 (0.845-0.970) & 0.913 (0.842-0.964) & 0.997 (0.990-1.000) \\
UNI & 448 & \bfseries 0.950 (0.898-0.989) & \bfseries 0.950 (0.901-0.988) & \bfseries 0.998 (0.995-1.000) \\
\midrule
ResNet-50$_{\text{IN}}$ & 896 & 0.600 (0.484-0.704) & 0.607 (0.496-0.711) & 0.848 (0.777-0.909) \\
CTransPath & 896 & 0.788 (0.688-0.868) & 0.791 (0.695-0.866) & 0.949 (0.909-0.978) \\
REMEDIS & 896 & 0.825 (0.741-0.900) & 0.830 (0.748-0.902) & 0.960 (0.926-0.984) \\
UNI & 896 & \bfseries 0.925 (0.858-0.976) & \bfseries 0.926 (0.862-0.975) & \bfseries 0.996 (0.990-1.000) \\
\midrule
ResNet-50$_{\text{IN}}$ & 1344 & 0.587 (0.464-0.697) & 0.593 (0.478-0.697) & 0.829 (0.754-0.894) \\
CTransPath & 1344 & 0.650 (0.547-0.749) & 0.656 (0.554-0.753) & 0.879 (0.817-0.929) \\
REMEDIS & 1344 & 0.812 (0.721-0.886) & 0.821 (0.728-0.891) & 0.966 (0.936-0.987) \\
UNI & 1344 & \bfseries 0.900 (0.829-0.960) & \bfseries 0.901 (0.829-0.962) & \bfseries 0.987 (0.970-0.997) \\
\bottomrule
\end{tabular}
\caption{\textbf{Linear probe evaluation for BRCA subtyping based on BACH (4 classes)}. Pre-extracted patch features of each encoder with logistic regression were evaluated on label-stratified train-test folds (80:20 ratio, 320:80 ROIs) across multiple image resolutions (without stain normalization), with test performance (n=80 ROIs) reported using balanced accuracy, weighted F1 score, and AUROC. Best performing model for each metric is bolded. 95\% CI is included in parentheses.}
\label{tab:patch-bach-lin}
\end{table}

\begin{table}
\footnotesize\centering
\begin{tabular}{ll|llll}
\toprule
Encoder & Img Res. & 1-NN Balanced ACC & 1-NN Weighted F1 & 20-NN Balanced ACC & 20-NN Weighted F1 \\
\midrule\midrule
ResNet-50$_{\text{IN}}$ & 224 & 0.588 (0.484-0.688) & 0.579 (0.455-0.685) & 0.625 (0.516-0.723) & 0.618 (0.504-0.723) \\
CTransPath & 224 & 0.750 (0.662-0.834) & 0.739 (0.635-0.834) & 0.750 (0.656-0.837) & 0.745 (0.640-0.838) \\
REMEDIS & 224 & 0.650 (0.550-0.742) & 0.628 (0.506-0.741) & 0.825 (0.741-0.904) & 0.823 (0.735-0.901) \\
UNI & 224 & \bfseries 0.812 (0.725-0.896) & \bfseries 0.813 (0.723-0.897) & \bfseries 0.900 (0.825-0.962) & \bfseries 0.902 (0.830-0.962) \\
\midrule
ResNet-50$_{\text{IN}}$ & 448 & 0.600 (0.497-0.697) & 0.585 (0.471-0.703) & 0.613 (0.506-0.717) & 0.601 (0.483-0.716) \\
CTransPath & 448 & 0.725 (0.630-0.814) & 0.713 (0.599-0.819) & 0.775 (0.682-0.863) & 0.771 (0.671-0.862) \\
REMEDIS & 448 & 0.575 (0.460-0.680) & 0.569 (0.452-0.680) & 0.687 (0.578-0.784) & 0.688 (0.583-0.782) \\
UNI & 448 & \bfseries 0.887 (0.812-0.948) & \bfseries 0.888 (0.813-0.950) & \bfseries 0.912 (0.841-0.965) & \bfseries 0.912 (0.838-0.963) \\
\midrule
ResNet-50$_{\text{IN}}$ & 896 & 0.587 (0.486-0.688) & 0.565 (0.448-0.691) & 0.538 (0.429-0.636) & 0.520 (0.399-0.638) \\
CTransPath & 896 & 0.650 (0.545-0.742) & 0.641 (0.531-0.745) & 0.725 (0.622-0.811) & 0.728 (0.624-0.817) \\
REMEDIS & 896 & 0.562 (0.451-0.667) & 0.568 (0.452-0.673) & 0.625 (0.514-0.726) & 0.638 (0.530-0.732) \\
UNI & 896 & \bfseries 0.850 (0.768-0.925) & \bfseries 0.849 (0.766-0.925) & \bfseries 0.887 (0.813-0.950) & \bfseries 0.885 (0.806-0.950) \\
\midrule
ResNet-50$_{\text{IN}}$ & 1344 & 0.525 (0.422-0.619) & 0.507 (0.390-0.621) & 0.475 (0.365-0.574) & 0.474 (0.363-0.577) \\
CTransPath & 1344 & 0.587 (0.478-0.691) & 0.591 (0.478-0.701) & 0.562 (0.444-0.663) & 0.571 (0.460-0.674) \\
REMEDIS & 1344 & 0.487 (0.371-0.594) & 0.503 (0.386-0.613) & 0.500 (0.389-0.610) & 0.512 (0.398-0.616) \\
UNI & 1344 & \bfseries 0.812 (0.722-0.892) & \bfseries 0.808 (0.710-0.889) & \bfseries 0.837 (0.749-0.914) & \bfseries 0.836 (0.750-0.913) \\
\bottomrule
\end{tabular}
\caption{\textbf{Few-shot and KNN evaluation for BRCA subtyping based on BACH (4 classes)}. Pre-extracted patch features of each encoder with SimpleShot ($K$=1) and Nearest Neighbors ($K$=20) were evaluated on label-stratified train-test folds (80:20 ratio, 320:80 ROIs) across multiple image resolutions (without stain normalization), with test performance (n=80 ROIs) reported using balanced accuracy and weighted F1 score. Best performing model for each metric is bolded. 95\% CI is included in parentheses.}
\label{tab:patch-bach-knn}
\end{table}

\begin{table}
\footnotesize\centering
\begin{tabular}{l|lll}
\toprule
Encoder & Balanced ACC & Weighted F1 & AUROC \\
\midrule
ResNet-50$_{\text{IN}}$ & 0.265 (0.259-0.271) & 0.787 (0.781-0.793) & 0.830 (0.822-0.839) \\
CTransPath & 0.422 (0.409-0.435) & 0.828 (0.823-0.833) & 0.881 (0.874-0.889) \\
REMEDIS & 0.410 (0.395-0.424) & 0.816 (0.811-0.821) & 0.836 (0.827-0.845) \\
UNI & \bfseries 0.465 (0.449-0.481) & \bfseries 0.844 (0.838-0.848) & \bfseries 0.888 (0.879-0.896) \\
\bottomrule
\end{tabular}
\caption{\textbf{Linear probe evaluation for CRC tissue classification based on HunCRC (9 classes)}. Pre-extracted patch features of each encoder with logistic regression were evaluated on case-stratified train-test folds (77:23 ratio, 76,753:22,655 ROIs), with test performance (n=22,655 ROIs) reported using balanced accuracy, weighted F1 score, and AUROC. Best performing model for each metric is bolded. 95\% CI is included in parentheses.}
\label{tab:patch-hun-lin}
\end{table}

\begin{table}
\footnotesize\centering
\begin{tabular}{l|lllll}
\toprule
Encoder & 1-NN Balanced ACC & 1-NN Weighted F1 & 20-NN Balanced ACC & 20-NN Weighted F1 \\
\midrule
ResNet-50$_{\text{IN}}$ & 0.421 (0.398-0.446) & 0.570 (0.563-0.576) & 0.379 (0.363-0.393) & 0.775 (0.769-0.781) \\
CTransPath & \bfseries 0.496 (0.475-0.517) & 0.695 (0.690-0.701) & 0.437 (0.425-0.453) & 0.806 (0.800-0.811) \\
REMEDIS & 0.445 (0.426-0.467) & 0.681 (0.675-0.687) & 0.400 (0.385-0.414) & 0.805 (0.800-0.810) \\
UNI & 0.470 (0.451-0.492) & \bfseries 0.747 (0.741-0.752) & \bfseries 0.448 (0.434-0.461) & \bfseries 0.831 (0.826-0.836) \\
\bottomrule
\end{tabular}
\caption{\textbf{Few-shot and KNN evaluation for CRC tissue classification based on HunCRC (9 classes)}. Pre-extracted patch features of each encoder with SimpleShot ($K$=1) and Nearest Neighbors ($K$=20) were evaluated on case-stratified train-test folds (77:23 ratio, 76,753:22,655 ROIs), with test performance (n=22,655 ROIs) reported using balanced accuracy and weighted F1 score. Best performing model for each metric is bolded. 95\% CI is included in parentheses.}
\label{tab:patch-huncrc-knn}
\end{table}

\begin{table}
\footnotesize\centering
\begin{tabular}{l|lll}
\toprule
Encoder & Balanced ACC & Weighted F1 & AUROC \\
\midrule\midrule
ResNet-50$_{\text{IN}}$ & 0.576 (0.573-0.579) & 0.853 (0.852-0.855) & 0.938 (0.936-0.940) \\
CTransPath & 0.728 (0.723-0.734) & 0.925 (0.924-0.926) & 0.964 (0.963-0.966) \\
REMEDIS & 0.774 (0.768-0.779) & 0.879 (0.878-0.881) & 0.969 (0.967-0.970) \\
UNI & \bfseries 0.829 (0.823-0.834) & \bfseries 0.942 (0.941-0.943) & \bfseries 0.971 (0.970-0.973) \\
\bottomrule
\end{tabular}
\caption{\textbf{Linear probe evaluation for ESCA subtyping based on UKK, WNS, TCGA and CHA (11 classes)}. Pre-extracted patch features of each encoder with logistic regression were evaluated using the UKK, WNS and TCGA cohorts as the train fold and the CHA cohort as the test fold (189,142:178,187), with test performance (n=178,187 ROIs) reported using balanced accuracy, weighted F1 score, and AUROC. Best performing model for each metric is bolded. 95\% CI is included in parentheses.}
\label{tab:patch-esca-lin}
\end{table}

\begin{table}
\footnotesize\centering
\begin{tabular}{l|llll}
\toprule
Encoder & 1-NN Balanced ACC & 1-NN Weighted F1 & 20-NN Balanced ACC & 20-NN Weighted F1 \\
\midrule\midrule
ResNet-50$_{\text{IN}}$ & 0.641 (0.635-0.646) & 0.720 (0.718-0.722) & 0.631 (0.626-0.636) & 0.837 (0.836-0.839) \\
CTransPath4 & 0.821 (0.817-0.826) & 0.856 (0.855-0.858) & 0.785 (0.780-0.789) & 0.918 (0.916-0.919) \\
REMEDIS & 0.793 (0.789-0.798) & 0.867 (0.865-0.868) & 0.777 (0.773-0.781) & 0.918 (0.917-0.919) \\
UNI & \bfseries 0.835 (0.830-0.840) & \bfseries 0.904 (0.903-0.905) & \bfseries 0.836 (0.832-0.840) & \bfseries 0.947 (0.946-0.948) \\
\bottomrule
\end{tabular}
\caption{\textbf{Few-shot and KNN evaluation for ESCA subtyping based on UKK, WNS, TCGA and CHA (11 classes)}. Pre-extracted patch features of each encoder with SimpleShot ($K$=1) and Nearest Neighbors ($K$=20) were evaluated using the UKK, WNS and TCGA cohorts as the train fold and the CHA cohort as the test fold (189,142:178,187 ROIs), with test performance (n=189,142 ROIs) reported using balanced accuracy and weighted F1 score. Best performing model for each metric is bolded. 95\% CI is included in parentheses.}
\label{tab:patch-esca-knn}
\end{table}

\begin{table}
\footnotesize\centering
\begin{tabular}{ll|lll}
\toprule
Encoder & Img Res. & Balanced ACC & Weighted F1 & AUROC \\
\midrule\midrule
ResNet-50$_{\text{IN}}$ & 224 & 0.349 (0.327-0.370) & 0.436 (0.412-0.457) & 0.830 (0.819-0.840) \\
CTransPath & 224 & 0.432 (0.410-0.453) & 0.481 (0.459-0.505) & \bfseries 0.843 (0.833-0.851) \\
REMEDIS & 224 & 0.446 (0.422-0.472) & 0.473 (0.452-0.494) & 0.801 (0.786-0.813) \\
UNI & 224 & \bfseries 0.504 (0.483-0.526) & \bfseries 0.533 (0.511-0.554) & 0.825 (0.814-0.835) \\
\midrule
ResNet-50$_{\text{IN}}$ & 448 & 0.363 (0.343-0.383) & 0.438 (0.416-0.460) & 0.832 (0.821-0.841) \\
CTransPath & 448 & 0.440 (0.419-0.460) & 0.497 (0.475-0.519) & \bfseries 0.840 (0.829-0.849) \\
REMEDIS & 448 & 0.458 (0.435-0.481) & 0.534 (0.513-0.555) & 0.816 (0.805-0.828) \\
UNI & 448 & \bfseries 0.514 (0.491-0.535) & \bfseries 0.565 (0.543-0.588) & 0.836 (0.824-0.847) \\
\midrule
ResNet-50$_{\text{IN}}$ & 896 & 0.341 (0.322-0.361) & 0.438 (0.415-0.461) & 0.816 (0.806-0.826) \\
CTransPath & 896 & 0.436 (0.418-0.456) & 0.507 (0.486-0.530) & 0.831 (0.820-0.841) \\
REMEDIS & 896 & 0.502 (0.478-0.525) & 0.553 (0.532-0.573) & 0.830 (0.819-0.841) \\
UNI & 896 & \bfseries 0.530 (0.507-0.552) & \bfseries 0.589 (0.568-0.609) & \bfseries 0.853 (0.842-0.863) \\
\midrule
ResNet-50$_{\text{IN}}$ & 1792 & 0.308 (0.288-0.328) & 0.435 (0.414-0.458) & 0.780 (0.770-0.790) \\
CTransPath & 1792 & 0.396 (0.376-0.414) & 0.474 (0.453-0.496) & 0.805 (0.795-0.816) \\
REMEDIS & 1792 & 0.460 (0.436-0.485) & 0.561 (0.542-0.583) & 0.819 (0.808-0.830) \\
UNI & 1792 & \bfseries 0.528 (0.507-0.549) & \bfseries 0.572 (0.550-0.591) & \bfseries 0.856 (0.845-0.866) \\
\bottomrule
\end{tabular}
\caption{\textbf{Linear probe evaluation for CRC polyp classification based on UniToPatho (6 classes)}. Pre-extracted patch features of each encoder with logistic regression were evaluated on the official train-test folds (72:28 ratio, 6,270:2,399 ROIs) across multiple image resolutions, with test performance (n=2,399 ROIs) reported using balanced accuracy, weighted F1 score, and AUROC. Best performing model for each metric is bolded. 95\% CI is included in parentheses.}
\label{tab:patch-unitopatho-lin}
\end{table}

\begin{table}
\footnotesize\centering
\begin{tabular}{ll|llll}
\toprule
Encoder & Img Res. & 1-NN Balanced ACC & 1-NN Weighted F1 & 20-NN Balanced ACC & 20-NN Weighted F1 \\
\midrule\midrule
ResNet-50$_{\text{IN}}$ & 224 & 0.468 (0.440-0.493) & 0.397 (0.375-0.417) & 0.331 (0.309-0.353) & 0.432 (0.408-0.453) \\
CTransPath & 224 & \bfseries 0.567 (0.544-0.589) & \bfseries 0.488 (0.467-0.510) & \bfseries 0.446 (0.425-0.468) & \bfseries 0.502 (0.481-0.525) \\
REMEDIS & 224 & 0.494 (0.470-0.517) & 0.420 (0.400-0.441) & 0.353 (0.330-0.377) & 0.466 (0.444-0.488) \\
UNI & 224 & 0.533 (0.508-0.554) & 0.429 (0.408-0.451) & 0.425 (0.405-0.445) & 0.476 (0.455-0.497) \\
\midrule
ResNet-50$_{\text{IN}}$ & 448 & 0.464 (0.437-0.489) & 0.387 (0.364-0.408) & 0.351 (0.330-0.373) & 0.443 (0.421-0.465) \\
CTransPath & 448 & 0.592 (0.569-0.614) & 0.507 (0.486-0.527) & \bfseries 0.458 (0.436-0.480) & \bfseries 0.514 (0.494-0.535) \\
REMEDIS & 448 & 0.517 (0.495-0.540) & 0.398 (0.378-0.418) & 0.406 (0.383-0.430) & 0.464 (0.443-0.487) \\
UNI & 448 & \bfseries 0.602 (0.577-0.627) & \bfseries 0.550 (0.530-0.571) & 0.458 (0.438-0.480) & 0.503 (0.482-0.523) \\
\midrule
ResNet-50$_{\text{IN}}$ & 896 & 0.443 (0.420-0.469) & 0.393 (0.372-0.413) & 0.361 (0.341-0.384) & 0.456 (0.436-0.479) \\
CTransPath & 896 & 0.565 (0.540-0.587) & 0.497 (0.477-0.517) & 0.440 (0.417-0.462) & 0.502 (0.482-0.524) \\
REMEDIS & 896 & 0.515 (0.490-0.538) & 0.426 (0.405-0.446) & 0.369 (0.348-0.389) & 0.442 (0.420-0.463) \\
UNI & 896 & \bfseries 0.619 (0.597-0.641) & \bfseries 0.543 (0.523-0.565) & \bfseries 0.471 (0.451-0.491) & \bfseries 0.508 (0.486-0.529) \\
\midrule
ResNet-50$_{\text{IN}}$ & 1792 & 0.351 (0.326-0.376) & 0.323 (0.303-0.343) & 0.350 (0.328-0.371) & 0.469 (0.448-0.491) \\
CTransPath & 1792 & 0.494 (0.471-0.516) & 0.474 (0.456-0.494) & 0.414 (0.394-0.436) & 0.494 (0.473-0.517) \\
REMEDIS & 1792 & 0.472 (0.446-0.495) & 0.413 (0.393-0.433) & 0.352 (0.330-0.376) & 0.464 (0.443-0.484) \\
UNI & 1792 & \bfseries 0.616 (0.594-0.639) & \bfseries 0.568 (0.549-0.587) & \bfseries 0.476 (0.456-0.498) & \bfseries 0.529 (0.506-0.549) \\
\bottomrule
\end{tabular}
\caption{\textbf{Few-shot and KNN evaluation for CRC polyp classification based on UniToPatho (6 classes)}. Pre-extracted patch features of each encoder with SimpleShot ($K$=1) and Nearest Neighbors ($K$=20) were evaluated on the official train-test folds (72:28 ratio, 6,270:2,399 ROIs) across multiple image resolutions, with test performance (n=2,399 ROIs) reported using balanced accuracy and weighted F1 score. Best performing model for each metric is bolded. 95\% CI is included in parentheses.}
\label{tab:patch-unitopatho-knn}
\end{table}

\begin{table}
\footnotesize\centering
\begin{tabular}{l|llll}
\toprule
Encoder & Balanced ACC & Quad. Weighted $\kappa$ & Weighted F1 & AUROC \\
\midrule\midrule
ResNet-50$_{\text{IN}}$ & 0.506 (0.504-0.507) & 0.592 (0.589-0.594) & 0.612 (0.610-0.614) & 0.862 (0.861-0.862) \\
CTransPath & 0.630 (0.629-0.632) & 0.779 (0.777-0.781) & 0.723 (0.722-0.725) & 0.915 (0.914-0.915) \\
REMEDIS  & 0.630 (0.629-0.632) & 0.761 (0.759-0.764) & 0.716 (0.714-0.717) & 0.906 (0.905-0.906) \\
UNI & \bfseries 0.658 (0.656-0.660) & \bfseries 0.797 (0.796-0.799) & \bfseries 0.743 (0.741-0.744) & \bfseries 0.922 (0.922-0.923) \\
\bottomrule
\end{tabular}
\caption{\textbf{Linear probe evaluation for PRAD tissue classification based on AGGC (5 classes)}. Pre-extracted patch features of each encoder with logistic regression were evaluated on the label stratified train-test folds (69:31 ratio, 780,619:345,021 ROIs) and test performance (n=345,021 ROIs) reported using balanced accuracy, Cohen's quadratic weighted $\kappa$, weighted F1 score, and AUROC. Best performing model for each metric is bolded. 95\% CI is included in parentheses.}
\label{tab:patch-aggc-lin}
\end{table}

\begin{table}
\centering\footnotesize
\begin{tabular}{l|llll}
\toprule
Encoder & 1-NN Balanced ACC & 1-NN Weighted F1 & 20-NN Balanced ACC & 20-NN Weighted F1 \\
\midrule\midrule
ResNet-50$_{\text{IN}}$ & 0.562 (0.560-0.564) & 0.482 (0.481-0.484) & 0.527 (0.526-0.529) & 0.605 (0.603-0.606) \\
CTransPath & 0.691 (0.690-0.693) & 0.630 (0.628-0.631) & 0.599 (0.597-0.600) & 0.666 (0.664-0.668) \\
REMEDIS & 0.655 (0.653-0.657) & 0.600 (0.598-0.602) & 0.617 (0.615-0.619) & 0.674 (0.672-0.675) \\
UNI & \bfseries 0.738 (0.736-0.740) & \bfseries 0.697 (0.696-0.699) & \bfseries 0.655 (0.653-0.656) & \bfseries 0.714 (0.712-0.715) \\
\bottomrule
\end{tabular}
\caption{\textbf{Few-shot and KNN evaluation for PRAD tissue classification based on AGGC (5 classes)}. Pre-extracted patch features of each encoder with SimpleShot ($K$=1) and Nearest Neighbors ($K$=20) were evaluated on label stratified train-test folds (69:31 ratio, 780,619:345,021 ROIs) and test performance (n=345,021 ROIs) reported using balanced accuracy and weighted F1 score. Best performing model for each metric is bolded. 95\% CI is included in parentheses.}
\label{tab:patch-aggc-knn}
\end{table}

\begin{table}
\footnotesize\centering
\begin{tabular}{ll|lll}
\toprule
Encoder & SN & Balanced ACC & Weighted F1 & AUROC \\
\midrule\midrule
ResNet-50$_{\text{IN}}$ & \xmark & 0.606 (0.598-0.614) & 0.686 (0.681-0.691) & 0.623 (0.615-0.633) \\
\rowcolor{gray!10} CTransPath & \xmark & 0.714 (0.707-0.720) & 0.772 (0.768-0.776) & 0.782 (0.776-0.789) \\
\rowcolor{gray!10} REMEDIS & \xmark & 0.649 (0.642-0.656) & 0.792 (0.788-0.797) & 0.740 (0.733-0.747) \\
UNI & \xmark & \bfseries 0.716 (0.708-0.723) & \bfseries 0.813 (0.809-0.817) & \bfseries 0.797 (0.791-0.804) \\
\midrule
ResNet-50$_{\text{IN}}$ & \checkmark & 0.606 (0.599-0.614) & 0.678 (0.673-0.683) & 0.629 (0.620-0.639) \\
\rowcolor{gray!10} CTransPath & \checkmark & 0.710 (0.703-0.717) & 0.763 (0.759-0.767) & 0.777 (0.770-0.784) \\
\rowcolor{gray!10} REMEDIS & \checkmark & 0.631 (0.625-0.639) & 0.777 (0.773-0.781) & 0.696 (0.688-0.703) \\
UNI & \checkmark & \bfseries 0.715 (0.708-0.722) & \bfseries 0.805 (0.801-0.809) & \bfseries 0.794 (0.788-0.801) \\
\bottomrule
\end{tabular}
\caption{\textbf{Linear probe evaluation for CRC MSI prediction based on TCGA (2 classes)}. Pre-extracted patch features of each encoder with logistic regression were evaluated on official train-test folds (38:62 ratio, 19,557:32,361 ROIs), with test performance (n=32,361 ROIs) reported using balanced accuracy, weighted F1 score, and AUROC with and without stain normalization (SN). Best performing model for each metric is bolded. 95\% CI is included in parentheses.}
\label{tab:patch-msi-lin}
\end{table}

\begin{table}
\footnotesize\centering
\begin{tabular}{ll|llll}
\toprule
Encoder & SN & 1-NN Balanced ACC & 1-NN Weighted F1 & 20-NN Balanced ACC & 20-NN Weighted F1 \\
\midrule\midrule
ResNet-50$_{\text{IN}}$ & \xmark & 0.595 (0.587-0.602) & 0.663 (0.658-0.668) & 0.596 (0.590-0.604) & 0.672 (0.667-0.676) \\
\rowcolor{gray!10} CTransPath & \xmark & 0.614 (0.608-0.622) & 0.669 (0.665-0.674) & 0.630 (0.624-0.638) & 0.736 (0.732-0.741) \\
\rowcolor{gray!10} REMEDIS & \xmark & 0.583 (0.576-0.590) & 0.669 (0.664-0.673) & 0.603 (0.595-0.610) & 0.737 (0.733-0.742) \\
UNI & \xmark & \bfseries 0.682 (0.675-0.688) & \bfseries 0.726 (0.722-0.731) & \bfseries 0.694 (0.687-0.700) & \bfseries 0.769 (0.765-0.773) \\
\midrule
ResNet-50$_{\text{IN}}$ & \checkmark & 0.592 (0.584-0.600) & 0.656 (0.651-0.661) & 0.596 (0.590-0.604) & 0.671 (0.666-0.676) \\
\rowcolor{gray!10} CTransPath & \checkmark & 0.616 (0.608-0.623) & 0.658 (0.653-0.663) & 0.634 (0.627-0.641) & 0.730 (0.726-0.735) \\
\rowcolor{gray!10} REMEDIS & \checkmark & 0.610 (0.603-0.617) & 0.658 (0.654-0.663) & 0.617 (0.610-0.624) & 0.724 (0.720-0.728) \\
UNI & \checkmark & \bfseries 0.683 (0.676-0.690) & \bfseries 0.723 (0.718-0.727) & \bfseries 0.683 (0.676-0.690) & \bfseries 0.755 (0.751-0.759) \\
\bottomrule
\end{tabular}
\caption{\textbf{Few-shot and KNN evaluation for CRC MSI prediction based on TCGA (2 classes)}. Pre-extracted patch features of each encoder with SimpleShot ($K$=1) and Nearest Neighbors ($K$=20) were evaluated on official train-test folds (38:62 ratio, 19,557:32,361 ROIs), with test performance (n=32,361 ROIs) reported using balanced accuracy and weighted F1 score with and without stain normalization (SN). Best performing model for each metric is bolded. 95\% CI is included in parentheses.}
\label{tab:patch-msi-knn}
\end{table}

\begin{table}
\footnotesize\centering
\begin{tabular}{ll|lll}
\toprule
Encoder & SN & Balanced ACC & Weighted F1 & AUROC \\
\midrule\midrule
ResNet-50$_{\text{IN}}$ & \xmark & 0.439 (0.434-0.444) & 0.525 (0.521-0.529) & 0.926 (0.924-0.927) \\
\rowcolor{gray!10} CTransPath & \xmark & 0.604 (0.599-0.610) & 0.678 (0.674-0.682) & 0.967 (0.966-0.968) \\
\rowcolor{gray!10} REMEDIS & \xmark & 0.650 (0.646-0.655) & 0.709 (0.705-0.713) & 0.959 (0.958-0.961) \\
UNI & \xmark & \bfseries 0.685 (0.680-0.690) & \bfseries 0.744 (0.741-0.748) & \bfseries 0.978 (0.978-0.979) \\
\midrule
ResNet-50$_{\text{IN}}$ & \checkmark & 0.413 (0.408-0.418) & 0.498 (0.493-0.502) & 0.916 (0.914-0.918) \\
\rowcolor{gray!10} CTransPath & \checkmark & 0.561 (0.555-0.566) & 0.637 (0.633-0.641) & 0.960 (0.959-0.961) \\
\rowcolor{gray!10} REMEDIS & \checkmark & 0.610 (0.605-0.615) & 0.678 (0.674-0.682) & 0.958 (0.956-0.959) \\
UNI & \checkmark & \bfseries 0.657 (0.652-0.662) & \bfseries 0.718 (0.714-0.721) & \bfseries 0.975 (0.974-0.976) \\
\bottomrule
\end{tabular}
\caption{\textbf{Linear probe evaluation for pan-cancer tissue classification based on TCGA (32 classes)}. Pre-extracted patch features of each encoder with logistic regression were evaluated on label-stratified train-test folds (216,350:55,360 ROIs), with test performance (n=55,360 ROIs) reported using balanced accuracy, weighted F1 score, and AUROC. Best performing model for each metric is bolded. 95\% CI is included in parentheses.}
\label{tab:patch-tcga-uniform-lin}
\end{table}

\begin{table}
\footnotesize\centering
\begin{tabular}{ll|llll}
\toprule
Encoder & SN & 1-NN Balanced ACC & 1-NN Weighted F1 & 20-NN Balanced ACC & 20-NN Weighted F1 \\
\midrule\midrule
ResNet-50$_{\text{IN}}$ & \xmark & 0.287 (0.282-0.292) & 0.269 (0.265-0.272) & 0.347 (0.342-0.351) & 0.415 (0.410-0.419) \\
\rowcolor{gray!10} CTransPath & \xmark & 0.460 (0.454-0.465) & 0.468 (0.464-0.472) & 0.495 (0.490-0.499) & 0.564 (0.560-0.568) \\
\rowcolor{gray!10} REMEDIS & \xmark & 0.553 (0.548-0.558) & 0.594 (0.590-0.598) & 0.608 (0.603-0.612) & 0.669 (0.665-0.673) \\
UNI & \xmark & \bfseries 0.641 (0.636-0.646) & \bfseries 0.675 (0.671-0.679) & \bfseries 0.645 (0.640-0.649) & \bfseries 0.707 (0.703-0.711) \\
\midrule
ResNet-50$_{\text{IN}}$ & \checkmark & 0.275 (0.270-0.280) & 0.254 (0.249-0.257) & 0.318 (0.314-0.323) & 0.382 (0.378-0.386) \\
\rowcolor{gray!10} CTransPath & \checkmark & 0.442 (0.436-0.447) & 0.438 (0.434-0.442) & 0.463 (0.459-0.468) & 0.534 (0.529-0.538) \\
\rowcolor{gray!10} REMEDIS & \checkmark & 0.504 (0.499-0.510) & 0.529 (0.525-0.533) & 0.541 (0.537-0.546) & 0.618 (0.614-0.622) \\
UNI & \checkmark & \bfseries 0.591 (0.586-0.597) & \bfseries 0.611 (0.607-0.615) & \bfseries 0.595 (0.590-0.600) & \bfseries 0.663 (0.660-0.667) \\
\bottomrule
\end{tabular}
\caption{\textbf{Few-shot and KNN evaluation for pan-cancer tissue classification based on TCGA (32 classes)}. Pre-extracted patch features of each encoder with SimpleShot ($K$=1) and Nearest Neighbors ($K$=20) were evaluated on label-stratified train-test folds (216,350:55,360 ROIs), with test performance (n=55,360 ROIs) reported using balanced accuracy and weighted F1 score with and without stain normalization (SN). Best performing model for each metric is bolded. 95\% CI is included in parentheses.}
\label{tab:patch-tcga-uniform-knn}
\end{table}

\begin{table}
\footnotesize\centering
\begin{tabular}{ll|lll}
\toprule
Encoder & SN & Balanced ACC & Weighted F1 & AUROC \\
\midrule\midrule
ResNet-50$_{\text{IN}}$ & \xmark & 0.667 (0.662-0.671) & 0.845 (0.842-0.849) & 0.919 (0.916-0.922) \\
\rowcolor{gray!10} CTransPath & \xmark & 0.760 (0.756-0.765) & 0.893 (0.891-0.896) & 0.944 (0.942-0.946) \\
\rowcolor{gray!10} REMEDIS & \xmark & \bfseries 0.891 (0.886-0.894) & 0.937 (0.935-0.939) & 0.975 (0.973-0.976) \\
UNI & \xmark & 0.887 (0.883-0.891) & \bfseries 0.944 (0.942-0.946) & \bfseries 0.979 (0.977-0.980) \\
\midrule
ResNet-50$_{\text{IN}}$ & \checkmark & 0.651 (0.646-0.655) & 0.836 (0.833-0.840) & 0.915 (0.911-0.917) \\
\rowcolor{gray!10} CTransPath & \checkmark & 0.758 (0.753-0.762) & 0.891 (0.889-0.894) & 0.930 (0.928-0.933) \\
\rowcolor{gray!10} REMEDIS & \checkmark & \bfseries 0.894 (0.890-0.898) & \bfseries 0.944 (0.942-0.946) & \bfseries 0.978 (0.977-0.979) \\
UNI & \checkmark & 0.870 (0.865-0.874) & 0.938 (0.936-0.940) & 0.975 (0.974-0.977) \\
\bottomrule
\end{tabular}
\caption{\textbf{Linear probe evaluation for pan-cancer TIL detection based on TCGA (2 classes)}. Pre-extracted patch features of each encoder with logistic regression were evaluated on official train-validation-test folds (69:13:19 ratio, 209,221:38,601:56,275 ROIs, train-validation combined during training) and test performance (n=56,275 ROIs) reported using balanced accuracy, weighted F1 score, and AUROC with and without stain normalization (SN). Best performing model for each metric is bolded. 95\% CI is included in parentheses.}
\label{tab:patch-tcga-tils-lin}
\end{table}

\begin{table}
\footnotesize\centering
\begin{tabular}{ll|llll}
\toprule
Encoder & SN & 1-NN Balanced ACC & 1-NN Weighted F1 & 20-NN Balanced ACC & 20-NN Weighted F1 \\
\midrule\midrule
ResNet-50$_{\text{IN}}$ & \xmark & 0.811 (0.806-0.815) & 0.842 (0.839-0.845) & 0.821 (0.816-0.826) & 0.908 (0.905-0.910) \\
\rowcolor{gray!10} CTransPath & \xmark & 0.806 (0.802-0.811) & 0.827 (0.824-0.829) & 0.840 (0.836-0.845) & 0.919 (0.916-0.921) \\
\rowcolor{gray!10} REMEDIS & \xmark & 0.804 (0.800-0.808) & 0.847 (0.844-0.849) & 0.836 (0.832-0.841) & 0.905 (0.903-0.908) \\
UNI & \xmark & \bfseries 0.886 (0.883-0.890) & \bfseries 0.931 (0.929-0.933) & \bfseries 0.865 (0.861-0.870) & \bfseries 0.935 (0.933-0.937) \\
\midrule
ResNet-50$_{\text{IN}}$ & \checkmark & 0.812 (0.807-0.816) & 0.848 (0.845-0.850) & 0.806 (0.801-0.811) & 0.900 (0.898-0.903) \\
\rowcolor{gray!10} CTransPath & \checkmark & 0.795 (0.790-0.799) & 0.813 (0.810-0.816) & 0.831 (0.827-0.836) & 0.914 (0.911-0.916) \\
\rowcolor{gray!10} REMEDIS & \checkmark & 0.856 (0.852-0.860) & 0.885 (0.883-0.888) & \bfseries 0.861 (0.857-0.865) & 0.930 (0.928-0.932) \\
UNI & \checkmark & \bfseries 0.865 (0.861-0.868) & \bfseries 0.907 (0.904-0.909) & 0.855 (0.851-0.859) & \bfseries 0.931 (0.928-0.933) \\
\bottomrule
\end{tabular}
\caption{\textbf{Few-shot and KNN evaluation for pan-cancer TIL detection based on TCGA (2 classes)}. Pre-extracted patch features of each encoder with SimpleShot ($K$=1) and Nearest Neighbors ($K$=20) were evaluated on official train-validation-test folds (69:13:19 ratio, 209,221:38,601:56,275 ROIs, train-validation combined during training) with test performance (n=56,275 ROIs) reported using balanced accuracy and weighted F1 score with and without stain normalization (SN). Best performing model for each metric is bolded. 95\% CI is included in parentheses.}
\label{tab:patch-tcga-tils-knn}
\end{table}

\begin{table}
\footnotesize\centering
\begin{tabular}{l|llll}
\toprule
Encoder & Acc@1 & Acc@3 & Acc@5 & MVAcc@5 \\
\midrule \midrule
ResNet-50$_{\text{IN}}$ & 0.551 (0.549, 0.552) & 0.765 (0.763, 0.766) & 0.835 (0.834, 0.836) & 0.623 (0.621, 0.624) \\
CTransPath & 0.620 (0.618, 0.621) & 0.795 (0.794, 0.797) & 0.850 (0.849, 0.851) & 0.671 (0.669, 0.672) \\
REMEDIS & 0.620 (0.618, 0.621) & 0.806 (0.804, 0.807) & 0.860 (0.859, 0.861) & 0.676 (0.675, 0.678) \\
UNI & \bfseries 0.663 (0.662, 0.665) & \bfseries 0.828 (0.827, 0.829) & \bfseries 0.875 (0.873, 0.876) & \bfseries 0.712 (0.710, 0.713) \\
\bottomrule
\end{tabular}
\caption{\textbf{Image retrieval for PRAD tissue classification based on AGGC (5 classes).} Pre-extracted patch features of each encoder with retrieval were evaluated on case-stratified train-test folds (780,619:345,021), with test performance (n=345,021 patches) reported using Acc@K for K $\in {1,3,5}$ and MVAcc@5. Best performing model for each model is bolded. 95\% CI is included in parentheses.}
\label{tab:aggc-retrieval}
\end{table}

\begin{table}
\footnotesize\centering
\begin{tabular}{l|llll}
\toprule
Encoder & Acc@1 & Acc@3 & Acc@5 & MVAcc@5 \\
\midrule \midrule
ResNet-50$_{\text{IN}}$ & 0.704 (0.693, 0.714) & 0.869 (0.861, 0.877) & 0.903 (0.895, 0.909) & 0.805 (0.795, 0.814) \\
CTransPath & 0.818 (0.809, 0.826) & 0.863 (0.855, 0.871) & 0.879 (0.871, 0.886) & 0.840 (0.832, 0.849) \\
REMEDIS & 0.895 (0.888, 0.902) & 0.946 (0.941, 0.951) & 0.958 (0.953, 0.962) & 0.921 (0.915, 0.927) \\
UNI & \bfseries 0.913 (0.907, 0.919) & \bfseries 0.953 (0.948, 0.958) & \bfseries 0.964 (0.960, 0.969) & \bfseries 0.931 (0.926, 0.937) \\
\bottomrule
\end{tabular}
\caption{\textbf{Image retrieval for CRC tissue classification based on CRC-100K (9 classes).} Pre-extracted patch features of each encoder with retrieval were evaluated on the official train-test folds (100,000:7,180), with test performance (n=7,180 patches) reported using Acc@K for K $\in {1,3,5}$ and MVAcc@5. Best performing model for each model is bolded. 95\% CI is included in parentheses.}
\label{tab:crc-retrieval}
\end{table}

\begin{table}
\footnotesize\centering
\begin{tabular}{l|llll}
\toprule
Encoder & Acc@1 & Acc@3 & Acc@5 & MVAcc@5 \\
\midrule \midrule
ResNet-50$_{\text{IN}}$ & 0.755 (0.753, 0.757) & 0.859 (0.858, 0.861) & 0.889 (0.888, 0.890) & 0.793 (0.791, 0.795) \\
CTransPath & 0.859 (0.857, 0.861) & 0.928 (0.927, 0.929) & 0.945 (0.944, 0.947) & 0.884 (0.882, 0.885) \\
REMEDIS & 0.873 (0.871, 0.874) & 0.938 (0.937, 0.939) & 0.954 (0.953, 0.955) & 0.899 (0.898, 0.901) \\
UNI & \bfseries 0.911 (0.910, 0.912) & \bfseries 0.960 (0.959, 0.961) & \bfseries 0.971 (0.971, 0.972) & \bfseries 0.929 (0.928, 0.930) \\
\bottomrule
\end{tabular}
\caption{\textbf{Image retrieval for ESCA subtyping based on UKK, WNS, TCGA and CHA (11 classes).} Pre-extracted patch features of each encoder with retrieval were evaluated using the UKK, WNS and TCGA cohorts as the train fold and the CHA cohort as the test fold (189,142:178,187), with test performance (n=178,187 patches) reported using Acc@K for K $\in {1,3,5}$ and MVAcc@5. Best performing model for each model is bolded. 95\% CI is included in parentheses.}
\label{tab:esca-retrieval}
\end{table}

\begin{table}
\footnotesize\centering
\begin{tabular}{l|llll}
\toprule
Encoder & Acc@1 & Acc@3 & Acc@5 & MVAcc@5 \\
\midrule \midrule
ResNet-50$_{\text{IN}}$ & 0.714 (0.708, 0.719) & 0.872 (0.867, 0.876) & 0.915 (0.911, 0.918) & 0.770 (0.765, 0.775) \\
CTransPath & 0.764 (0.758, 0.769) & 0.880 (0.876, 0.884) & 0.910 (0.906, 0.913) & 0.798 (0.793, 0.803) \\
REMEDIS & 0.765 (0.759, 0.770) & 0.885 (0.882, 0.889) & 0.917 (0.913, 0.920) & 0.805 (0.800, 0.809) \\
UNI & \bfseries 0.793 (0.788, 0.798) & \bfseries 0.901 (0.897, 0.905) & \bfseries 0.928 (0.925, 0.931) & \bfseries 0.830 (0.825, 0.835) \\
\bottomrule
\end{tabular}
\caption{\textbf{Image retrieval for CRC tissue classification based on HunCRC (9 classes).} Pre-extracted patch features of each encoder with retrieval were evaluated on case-stratified train-test folds (76,753:22,655), with test performance (n=22,655 patches) reported using Acc@K for K $\in {1,3,5}$ and MVAcc@5. Best performing model for each model is bolded. 95\% CI is included in parentheses.}
\label{tab:huncrc-retrieval}
\end{table}

\clearpage

\begin{table}
\footnotesize\centering
\begin{tabular}{ll|llll}
\toprule
Encoder & SN & Acc@1 & Acc@3 & Acc@5 & MVAcc@5 \\
\midrule \midrule
ResNet-50$_{\text{IN}}$ & \xmark & 0.367 (0.363, 0.371) & 0.532 (0.528, 0.537) & 0.610 (0.606, 0.614) & 0.453 (0.449, 0.458) \\
\rowcolor{gray!10} CTransPath & \xmark & 0.521 (0.517, 0.525) & 0.660 (0.657, 0.664) & 0.722 (0.718, 0.725) & 0.578 (0.575, 0.583) \\
\rowcolor{gray!10} REMEDIS & \xmark & 0.627 (0.623, 0.631) & 0.739 (0.735, 0.743) & 0.786 (0.782, 0.789) & 0.674 (0.670, 0.678) \\
UNI & \xmark & \bfseries 0.666 (0.663, 0.671) & \bfseries 0.780 (0.777, 0.784) & \bfseries 0.825 (0.822, 0.828) & \bfseries 0.712 (0.708, 0.716) \\
\midrule
ResNet-50$_{\text{IN}}$ & \checkmark & 0.339 (0.335, 0.343) & 0.508 (0.504, 0.512) & 0.590 (0.586, 0.594) & 0.433 (0.429, 0.437) \\
\rowcolor{gray!10} CTransPath & \checkmark & 0.483 (0.479, 0.488) & 0.639 (0.635, 0.643) & 0.707 (0.703, 0.711) & 0.556 (0.552, 0.560) \\
\rowcolor{gray!10} REMEDIS & \checkmark & 0.567 (0.563, 0.571) & 0.704 (0.701, 0.708) & 0.760 (0.757, 0.764) & 0.630 (0.626, 0.634) \\
UNI & \checkmark & \bfseries 0.613 (0.609, 0.617) & \bfseries 0.748 (0.745, 0.752) & \bfseries 0.803 (0.800, 0.806) & \bfseries 0.672 (0.668, 0.676) \\
\bottomrule
\end{tabular}
\caption{\textbf{Image retrieval for pan-cancer tissue classification based on TCGA (32 classes).} Pre-extracted patch features of each encoder with retrieval were evaluated on case-stratified train-test folds (216,350:55,360), with test performance (n=55,360 patches) reported using Acc@K for K $\in {1,3,5}$ and MVAcc@5 with and without stain normalization (SN). Best performing model for each model is bolded. 95\% CI is included in parentheses.}
\label{tab:tcga-uniform-retrieval}
\end{table}

\begin{table}
\footnotesize\centering
\begin{tabular}{l|llll}
\toprule
Encoder & Acc@1 & Acc@3 & Acc@5 & MVAcc@5 \\
\midrule \midrule
ResNet-50$_{\text{IN}}$ & 0.444 (0.424, 0.464) & 0.648 (0.628, 0.666) & 0.725 (0.707, 0.742) & 0.514 (0.492, 0.534) \\
CTransPath & 0.480 (0.460, 0.500) & 0.632 (0.614, 0.651) & 0.694 (0.675, 0.712) & 0.523 (0.501, 0.541) \\
REMEDIS & 0.437 (0.418, 0.456) & 0.599 (0.580, 0.618) & 0.670 (0.651, 0.688) & 0.488 (0.469, 0.506) \\
UNI & \bfseries 0.504 (0.483, 0.524) & \bfseries 0.671 (0.652, 0.689) & \bfseries 0.736 (0.718, 0.753) & \bfseries 0.556 (0.536, 0.575) \\
\bottomrule
\end{tabular}
\caption{\textbf{Image retrieval for CRC polyp classification based on UniToPatho.} Pre-extracted patch features of each encoder with retrieval were evaluated on the official case-stratified train-test folds (6,270:2,399) with resized $1792^2$ image resolutions, with test performance (n=345,021 patches) reported using Acc@K for K $\in {1,3,5}$ and MVAcc@5. Best performing model for each model is bolded. 95\% CI is included in parentheses.}
\label{tab:unitopatho-retrieval}
\end{table}
\begin{table}
\footnotesize\centering
\begin{tabular}{ll|lll}
\toprule
Encoder & Cell type & Dice & Precision & Recall \\
\midrule\midrule
ResNet-50$_{\text{IN}}$ & Endothelium       & 0.665 (0.658-0.674) & 0.709 (0.697-0.722) & 0.658 (0.654-0.663) \\
CTransPath & Endothelium & 0.658 (0.644-0.663) & 0.695 (0.682-0.702) & 0.656 (0.641-0.663) \\
REMEDIS & Endothelium & 0.684 (0.677-0.694) & 0.679 (0.672-0.688) & \bfseries 0.729 (0.724-0.736) \\
UNI & Endothelium & \bfseries 0.696 (0.686-0.703) & \bfseries 0.718 (0.707-0.726) &  0.709 (0.700-0.716) \\
\midrule
ResNet-50$_{\text{IN}}$ & Epithelium        & 0.812 (0.808-0.814) & 0.840 (0.837-0.841) & 0.816 (0.813-0.819) \\
CTransPath & Epithelium & 0.815 (0.811-0.817) & 0.842 (0.839-0.845) & 0.816 (0.812-0.818) \\
REMEDIS & Epithelium & 0.824 (0.819-0.829)& 0.843 (0.840-0.846) & 0.834 (0.829-0.838) \\
UNI & Epithelium                            & \bfseries 0.827 (0.823-0.830) & \bfseries 0.849 (0.847-0.851) & \bfseries 0.834 (0.831-0.838) \\
\midrule
ResNet-50$_{\text{IN}}$ & Leukocyte         & 0.691 (0.689-0.693) & 0.726 (0.724-0.727) & 0.692 (0.690-0.695) \\
CTransPath & Leukocyte                      & 0.686 (0.684-0.688) & 0.710 (0.707-0.711) & 0.699 (0.696-0.702) \\
REMEDIS & Leukocyte & 0.706 (0.704-0.707) & 0.725 (0.722-0.727) & \bfseries 0.719 (0.717-0.721) \\
UNI & Leukocyte & \bfseries 0.706 (0.705-0.709) & \bfseries 0.736 (0.734-0.737) & 0.713 (0.712-0.716) \\
\midrule
ResNet-50$_{\text{IN}}$ & Lymphocyte        & 0.631 (0.625-0.638) & 0.693 (0.684-0.701) & 0.626 (0.618-0.634) \\
CTransPath & Lymphocyte                     & 0.629 (0.623-0.635) & 0.677 (0.671-0.682) & 0.627 (0.620-0.632) \\
REMEDIS & Lymphocyte                        & \bfseries 0.653 (0.650-0.660) & \bfseries 0.686 (0.680-0.696) & 0.665 (0.659-0.669) \\
UNI & Lymphocyte                            & 0.651 (0.647-0.658) & 0.686 (0.680-0.693) & \bfseries 0.665 (0.657-0.673) \\
\midrule
ResNet-50$_{\text{IN}}$ & Smooth Muscle     & 0.650 (0.646-0.652) & 0.699 (0.696-0.703) & 0.668 (0.664-0.672) \\
CTransPath & Smooth Muscle                  & 0.655 (0.648-0.659) & 0.703 (0.698-0.707) & 0.670 (0.665-0.675) \\
REMEDIS & Smooth Muscle                     & 0.674 (0.668-0.678) & 0.724 (0.719-0.728) & 0.687 (0.684-0.693) \\
UNI & Smooth Muscle                         & \bfseries 0.690 (0.685-0.694) & \bfseries 0.736 (0.731-0.740) & \bfseries 0.704 (0.700-0.708) \\
\midrule
ResNet-50$_{\text{IN}}$                     & Myeloid Cell & 0.615 (0.611-0.620) & 0.698 (0.690-0.703) & 0.596 (0.592-0.602) \\
CTransPath                                  & Myeloid Cell & 0.621 (0.618-0.627) & 0.697 (0.692-0.700) & 0.604 (0.599-0.610) \\
REMEDIS & Myeloid Cell                      & 0.652 (0.647-0.659) & \bfseries 0.728 (0.719-0.733) & 0.630 (0.625-0.638) \\
UNI & Myeloid Cell                          & \bfseries 0.656 (0.655-0.662) & 0.726 (0.724-0.730) & \bfseries 0.637 (0.634-0.643) \\
\midrule
ResNet-50$_{\text{IN}}$ & Plasma Cell       & 0.703 (0.692-0.706) & 0.761 (0.745-0.769) & 0.692 (0.683-0.696) \\
CTransPath & Plasma Cell                    & 0.713 (0.703-0.720) & 0.754 (0.743-0.760) & 0.709 (0.700-0.716) \\
REMEDIS & Plasma Cell                       & \bfseries 0.742 (0.734-0.747) & 0.784 (0.775-0.786) & \bfseries 0.736 (0.725-0.742) \\
UNI & Plasma Cell                           & 0.737 (0.730-0.743) & \bfseries 0.788 (0.778-0.794) & 0.728 (0.722-0.732) \\
\midrule
ResNet-50$_{\text{IN}}$ & RBC               & 0.797 (0.793-0.802) & 0.827 (0.825-0.831) & 0.808 (0.800-0.812) \\
CTransPath & RBC                            & 0.786 (0.784-0.792) & 0.818 (0.815-0.821) & 0.794 (0.789-0.800) \\
REMEDIS & RBC                               & 0.795 (0.791-0.800) & 0.824 (0.821-0.827) & 0.807 (0.802-0.815) \\
UNI & RBC                                   & \bfseries 0.803 (0.800-0.808) & \bfseries 0.839 (0.837-0.843) & \bfseries 0.810 (0.804-0.817) \\
\midrule\midrule
ResNet-50$_{\text{IN}}$ & Average & 0.696 & 0.744 & 0.695 \\
CTransPath & Average & 0.695 & 0.737 & 0.700 \\
REMEDIS & Average & 0.716 & 0.749 & \bfseries 0.726 \\
UNI & Average & \bfseries 0.721 & \bfseries 0.760 & 0.725 \\
\bottomrule
\end{tabular}
\caption{\textbf{Pan-cancer cell type segmentation based on SegPath (8 cell types treated as individual tasks}. ROI-level cell segmentation of eight major cell types using their official train-test folds in the SegPath dataset. We finetune each pretrained encoder using the Mask2Former framework\cite{cheng2021mask2former}, a flexible framework used commonly for adapting self-supervised models as segmentation backbones for dense prediction tasks. All encoders were trained and evaluated on the official train-validation-test folds, with test performance reported using dice score, precision, and recall. Best performing model for each metric is bolded. 95\% CI is included in parentheses.}
\label{tab:patch-level-seg}
\end{table}
\begin{table}
\footnotesize\centering
\begin{tabular}{lcc|lll}
\toprule
Encoder & Cohort & Top K & Balanced ACC & Weighted F1 \\
\midrule\midrule
ResNet-50$_{\text{IN}}$ & TCGA & 5 & 0.531 (0.500-0.566) & 0.398 (0.280-0.522) \\
\rowcolor{gray!10} CTransPath & TCGA & 5 & 0.796 (0.705-0.869) & 0.796 (0.712-0.868)  \\
\rowcolor{gray!10} REMEDIS & TCGA & 5 & 0.796 (0.707-0.874) & 0.795 (0.709-0.877) \\
UNI & TCGA & 5 & \bfseries 0.908 (0.847-0.960) & \bfseries 0.908 (0.847-0.959) \\
\midrule
ResNet-50$_{\text{IN}}$ & TCGA & 50 & 0.531 (0.488-0.574) & 0.412 (0.295-0.526) \\
\rowcolor{gray!10} CTransPath & TCGA & 50 & 0.806 (0.724-0.878) & 0.806 (0.724-0.878) \\
\rowcolor{gray!10} REMEDIS & TCGA & 50 & 0.786 (0.701-0.868) & 0.785 (0.701-0.867) \\
UNI & TCGA & 50 & \bfseries 0.918 (0.863-0.970) & \bfseries 0.918 (0.867-0.969) \\
\midrule\midrule
ResNet-50$_{\text{IN}}$ & CPTAC & 5 & 0.530 (0.517-0.544) & 0.444 (0.407-0.481) \\
CTransPath & CPTAC & 5 & 0.845 (0.823-0.865) & 0.847 (0.825-0.867) \\
REMEDIS & CPTAC & 5 & 0.771 (0.748-0.793) & 0.752 (0.723-0.778) \\
UNI & CPTAC & 5 & \bfseries 0.902 (0.882-0.918) & \bfseries 0.902 (0.883-0.919) \\
\midrule
ResNet-50$_{\text{IN}}$ & CPTAC & 50 & 0.535 (0.520-0.551) & 0.461 (0.426-0.497) \\
CTransPath & CPTAC & 50 & 0.827 (0.805-0.848) & 0.830 (0.808-0.852)  \\
REMEDIS & CPTAC & 50 & 0.806 (0.783-0.825) & 0.793 (0.767-0.816) \\
UNI & CPTAC & 50 & \bfseries 0.901 (0.882-0.918) & \bfseries 0.901 (0.883-0.918) \\
\bottomrule
\end{tabular}
\caption{\textbf{Prototypical NSCLC subtyping based on TCGA and CPTAC (2 classes)}. Class prototypes for MI-SimpleShot were developed using annotated ROIs from the TCGA Uniform Tumor Dataset, which we evaluate on the NSCLC subtyping task using the same folds in weakly-supervised ABMIL evaluation (train-test folds with 86:14, 602:98 slides), with external evaluation on slides ($n=1091$) sourced from CPTAC-LUAD ($n=578$) and CPTAC-LUSC ($n=513$). Test performance was reported using balanced accuracy, and weighted F1 score. Best performing model for each metric is bolded. 95\% CI is included in parentheses.}
\label{tab:proto-nsclc}
\end{table}

\begin{table}
\footnotesize\centering
\begin{tabular}{lcc|lll}
\toprule
Encoder & Cohort & Top K & Balanced ACC & Weighted F1 \\
\midrule\midrule
ResNet-50$_{\text{IN}}$ & TCGA & 5 & 0.636 (0.530-0.725) & 0.501 (0.394-0.601)  \\
\rowcolor{gray!10} CTransPath & TCGA & 5 & 0.896 (0.814-0.950) & 0.867 (0.803-0.928) \\
\rowcolor{gray!10} REMEDIS & TCGA & 5 & 0.769 (0.712-0.828) & 0.728 (0.625-0.826) \\
UNI & TCGA & 5 & \bfseries 0.928 (0.886-0.964) & \bfseries 0.888 (0.827-0.940) \\
\midrule
ResNet-50$_{\text{IN}}$ & TCGA & 50 & 0.700 (0.597-0.791) & 0.607 (0.502-0.707) \\
\rowcolor{gray!10} CTransPath & TCGA & 50 & 0.902 (0.824-0.956) & 0.878 (0.813-0.930) \\
\rowcolor{gray!10} REMEDIS & TCGA & 50 & 0.757 (0.696-0.820) & 0.710 (0.609-0.809) \\
UNI & TCGA & 50 & \bfseries 0.938 (0.903-0.969) & \bfseries 0.897 (0.837-0.949) \\
\midrule\midrule
ResNet-50$_{\text{IN}}$ & CPTAC & 5 & 0.701 (0.654-0.741) & 0.439 (0.405-0.473) \\
CTransPath & CPTAC & 5 & 0.919 (0.905-0.930) & 0.828 (0.804-0.849) \\
REMEDIS & CPTAC & 5 & \bfseries 0.957 (0.940-0.973) & \bfseries 0.939 (0.924-0.954) \\
UNI & CPTAC & 5 & 0.952 (0.940-0.961) & 0.896 (0.878-0.914) \\
\midrule
ResNet-50$_{\text{IN}}$ & CPTAC & 50 & 0.744 (0.694-0.791) & 0.575 (0.543-0.606) \\
CTransPath & CPTAC & 50 & 0.891 (0.846-0.925) & 0.829 (0.807-0.849) \\
REMEDIS & CPTAC & 50 & \bfseries 0.957 (0.937-0.975) & \bfseries 0.951 (0.937-0.965) \\
UNI & CPTAC & 50 & 0.957 (0.946-0.967) & 0.909 (0.892-0.925) \\
\bottomrule
\end{tabular}
\caption{\textbf{Prototypical RCC subtyping based on TCGA and CPTAC-DHMC (3 classes)}. Class prototypes for MI-SimpleShot were developed using annotated ROIs from the TCGA Uniform Tumor Dataset, which we evaluate on the RCC subtyping task using the same folds in weakly-supervised ABMIL evaluation (train-test folds with 82:18, 431:97 slides), with external evaluation on slides $(n=872)$ sourced from CPTAC-CCRCC $(n=404)$ and DHMC-Kidney $(n=468)$ (CCRCC, CHRCC, and PRCC cases only). Test performance was reported using balanced accuracy, and weighted F1 score. Best performing model for each metric is bolded. 95\% CI is included in parentheses.}
\label{tab:proto-rcc}
\end{table}

\clearpage
\begin{nolinenumbers}
\Heading{References}

\vspace{2mm}

\begin{spacing}{0.9}
\bibliographystyle{naturemag}
\bibliography{main}
\end{spacing}
\end{nolinenumbers}
\clearpage

\end{document}